\newif\ifetna
\etnafalse

\ifetna
\documentclass[final,leqno,letterpaper]{etna}
\setbibdata{1}{xx}{46}{2017} 

\hypersetup{%
    pdftitle={Graph Lineages and Skeletal Products},
    pdfauthor={Eric Mjolsness and Cory B. Scott},
    pdfkeywords={graph theory, numerical linear algebra, graph algorithms}
    }

\title{Graph lineages and skeletal graph products\thanks{%
Received... Accepted... Published online on... Recommended by....
}}

\author{Eric Mjolsness\footnotemark[2]
        \and Cory B. Scott\footnotemark[3]}

\shorttitle{SKELETAL PRODUCTS} 
\shortauthor{E.~MJOLSNESS, AND C.~SCOTT}

\usepackage{lineno}
\linenumbers

\else
\documentclass{article}

\title{Graph lineages and skeletal graph products}

\author{Eric Mjolsness\footnotemark[2]
        \and Cory B. Scott\footnotemark[3]}

\usepackage{url}
\usepackage{hyperref}
\usepackage{fullpage}

\fi

\usepackage{algorithm2e}
\usepackage{amsfonts}
\usepackage{amssymb}

\usepackage{times}

\usepackage[square,sort,comma,numbers]{natbib}

\usepackage{amsmath}

\usepackage{caption}

\usepackage{multirow}

\usepackage{float}

\newfloat{diagfig}{hbtp}{lop}
\floatname{diagfig}{Diagram}

\usepackage{tikz}
\usepackage{tikz-cd}
\tikzset{pullback/.style={path picture={
\draw[opacity=1,black,-,#1] (-0.5,-0.5) -- (-0.0,0.5) -- (0.5,0.5);%
}}}

\usetikzlibrary{positioning}

\begin{document}

\maketitle

\renewcommand{\thefootnote}{\fnsymbol{footnote}}

\footnotetext[2]{Department of Computer Science, 
University of California Irvine, Irvine, CA, 92612}
\footnotetext[3]{Department of Mathematics and Computer Science, Colorado College, 14 E. Cache La Poudre, Colorado Springs, CO 80903}

\begin{abstract}
Graphs, and sequences of successively larger interrelated graphs, can be used to specify
the architecture of mathematical models in many fields including machine learning
and computational science. 
Here we define structured graph ``lineages'' 
(ordered by level number)
that grow in a hierarchical fashion, so that:
(1) the number of graph vertices and edges increases roughly exponentially in 
level number; 
(2) 
bipartite graphs
connect successive levels within a graph lineage and, 
as in algebraic multigrid methods, 
can constrain sparsity-compatible orthogonal
prolongation and restriction
matrices relating successive levels;
(3) 
using prolongation 
maps within a graph lineage,
process-derived distance measures between graphs
at successive levels
can be defined;
(4)
a category of ``graded graphs'' can be defined,
and using it low-cost ``skeletal'' variants of standard algebraic 
graph operations
and type constructors
(cross product, box product, disjoint sum, and function types) can be derived for graded graphs
and hence hierarchical graph lineages;
(5) these skeletal binary operators have similar but not identical
algebraic and category-theoretic
properties to their standard counterparts;
(6) graph lineages and their skeletal product constructors
can approach continuum limit objects.
Additional space-efficient unary operators 
on graded graphs are also derived:
thickening, which creates a graph lineage of multiscale graphs,
and escalation to a graph lineage of search frontiers
(useful as a generalization of adaptive grids
and in defining ``skeletal'' functions).
The result is an algebraic type theory for graded graphs and
(hierarchical) graph lineages.
The approach is expected to be particularly well suited to defining
hierarchical model architectures - ``hierarchitectures''
- and local sampling, search, or optimization algorithms on them.
We demonstrate such application to deep neural networks
(including visual and feature scale spaces)
and to multigrid numerical methods.
\end{abstract}

\ifetna

\fi

\section{Introduction}
\label{sec:intro}
Graphs, and especially growing families of related graphs, are a fundamental object in computational science. Hierarchies of interrelated graphs have applications in pattern recognition \cite{fukushima1988neocognitron, felzenszwalb2007hierarchical,liu2018learning, jin2006context}, 
partial differential equation (PDE) solving \cite{stuben2001review,memin1998multigrid,bank1988hierarchical,hartmann2008adaptive,wesseling1995introduction,trottenberg2001multigrid,brandt2006guide}, and in the construction of machine learning architectures \cite{ke2017multigrid,you2020structure,gong2020geometrically,lei2017deriving,luz2020learning, liu2018progressive, elsken2019neural,scott2020graph}. In this work we 
feature 
the last application. 
%
%
%
In a pattern recognition context, many tasks necessitate finding an optimal assignment over the set of $nm$ combinations between $n$ input elements and $m$ patterns. A specific example is the Hough transform \cite{duda1972use,ballard1981generalizing,illingworth1987adaptive,illingworth1988survey}, the typical formulation of which requires a $r \times t$ accumulator array to estimate the intercepts and slopes of lines in an image at parameter resolutions $(1/r)$, $(1/t)$ respectively. Finer and finer discretizations of the range of possible parameters are related in a type of hierarchy called a \emph{scale space} \cite{lindeberg2013scale}.  Adding additional scale spaces (for example, if we must also search over locations in an image or over object joint angles) results in the problem size growing as the product of the individual spaces. This ``curse of dimensionality'' is also present in machine learning architecture design, leading to very large search spaces of possible models \cite{pham2018efficient}. 

We propose that this kind of scale space product can be modelled by a product of graphs which represent discretized versions of the spaces involved. However, this does not directly resolve the growth rate issue. Thus, we define ``skeletonized'' versions of standard graph products. These products (strictly speaking, products between growing lineages of graphs and not products of graphs themselves) generalize the idea of adaptive resolution, allowing us to model a trade-off in representation complexity between multiple scale spaces given a fixed computational budget. We justify these products via category-theoretic constructions in the category of graphs and graph homomorphisms\footnote{More precisely, we give a construction of graph lineages as a slice category of the category of graphs and graph homomorphisms, and we construct pullback diagrams that demonstrate how graph categorical products can be extended to graph \emph{lineage} categorical products.}, 
as well as by demonstrating their similar algebraic propoerties; 
furthermore, we validate their use in a machine learning context by using skeletal products to reproduce and train a common machine learning architecture. Additionally, we demonstrate that for PDE solving on domains that can be decomposed as products of graph lineages, skeletal products can be used to develop accelerated versions of multigrid-like smoothing algorithms. 

\subsection{Prior work}
We first discuss some prior work in the intersection of category theory, graph theory, and machine learning. Machine learning (ML), while certainly a transformational science, has attracted criticism \cite{mcgreivy2024weak} for being an \emph{ad hoc} discipline. While ML practitioners have assembled a large library of effective techniques, there is thus far no unifying theory that describes why some ML models are more effective than others at a given task. Small changes, such as reordering the order of BatchNorm layers and activation functions in a model architecture, can have a disproportionate impact on that model's performance. Furthermore, the dependence of model performance on these factors is not well-understood, and can seriously harm the reproducibility of results \cite{olah2017research, raff2019step, pineau2021improving, d2022underspecification, madan2022and}. Thus, a major goal of modern machine learning research is the formalization of design principles and construction procedures for machine learning architectures. Extensive work in recent years \cite{baydin2018automated, parekh2000constructive, feurer2015automated, kang2020model, srivastava2020empirical, kirsch2018modular, ritter2017deriving, zweig2021functional} has attempted to a) formalize the ML architecture construction process by mathematically expressing design principles, and b) use those principles in the discovery of new model architectures. Often these techniques can be used to enhance Neural Architecture Search (NAS; \cite{mellor2021neural, zoph2016neural, deng2017peephole, liu2018progressive, istrate2019tapas, elsken2019neural}) and related methods for programmatic discovery of performant models given a space of possible hypotheses. 

Attempts to formalize modern machine learning practice make use of a wide variety of mathematical techniques. Most relevant to this work are works which use category-theoretic constructions to build machine learning architectures satisfying certain properties, as we do in Section \ref{sec:skel_prods}. Many of these works \cite{cruttwell2022categorical, xu2022neural, fabregat2023exploring, gavranovic2024position} formalize machine learning architectures by analyzing the category-theoretic properties of small neural network components, as well as rules for connecting those components together \cite{badias2024neural}. This is well-motivated by the way machine learning models are assembled in practice, and leads to models whose learning capabilities are well-understood (in fact guaranteed) by category-theoretic proofs \cite{bradley2023structure}. Parallel approaches use other formalisms such as topology \cite{carlsson2020topological}, string diagrams \cite{alvarezpicallo2023functorial, xu2022neural}, or algebraic representations \cite{watanabe2001algebraic, jackson2017algebraic, joyce2024algebraic, flinkow2024towards, daniely2016deeper}. 

As motivation for the model we discuss in Section \ref{subsec:cnn_model_desc}, we briefly discuss previous applications of scale spaces in computer vision. We refer the interested reader to \cite{salehin2024automl} for a general review of automated architecture search and \cite{voulodimos2018deep} for a review of 
machine learning in
computer vision. As mentioned above, 
a prototypical pattern recognition algorithm is the Hough transform \cite{duda1972use} and its variants \cite{ballard1981generalizing,illingworth1988survey}. The variant due to Ballard et al. 
\cite{ballard1981generalizing}
is notable for including a scale space over object description parameters in addition to object pose parameters. Another common technique \cite{illingworth1987adaptive} for computing the Hough transform makes use of the scale space over parameters to gradually refine the accumulation array. Because a hierarchy of Gaussian kernels of varying widths produces a scale space of images, the theory of scale spaces \cite{babaud1986uniqueness, sporring2013gaussian, lindeberg2013scale} is a natural formalism to apply to image feature extraction, as in the well-known ``Gaussian Pyramid'' for efficient visual processing \cite{adelson1984pyramid}. Other works use properties of scale spaces to define ML models which are invariant or equivariant to affine transformations of their input \cite{araujo1997novel, zhu2022scaling, jansson2022scale, bekkersb, lehner2023gauge, worrall2019deep, kalogeropoulos2024scale}.  

While not specifically a scale space, many modern machine learning architectures utilize multiscale ideas: the Neocognitron of Fukushima et al. \cite{fukushima1988neocognitron} is an early example of a model inspired by the multiscale way in which animal brains process visual data,
progressively trading off spatial resolution with 
specificity in a hierarchy of learned feature extractors. 
More recent work such as Convolutional Neural Networks and their variants \cite{liu2023multiresolution, krizhevsky2012imagenet, lecun2015deep, he2016deep, ronneberger2015u} realize this architecture in the context
of the backpropagation learning algorithm with convolutional weight-sharing. 
Other examples of multiscale ML include  applications of neural networks to problems from the multigrid literature \cite{chen2022meta, huang2022learning}. In this context, we refer both to ML accelerations for multigrid solvers \cite{chen2021transfer, liang2024solving, perera2024multiscale, yang2022amgnet, yang2023reinforcement, zhang2024blending} as well as the use of multigrid-like graph hierarchies or prolongation operators to build ML models \cite{reed2017parallel, lim2023graph, grundmann2010efficient, guo2021hierarchical, zhang2022hierarchical, ying2018hierarchical, yehudai2021local, tian2024visual, perret2019higra, li2019semi}.

In this work, by contrast, we adopt 
algebraic, category-theoretic 
descriptions of particular infinitely large graphs as our main formalism. 
Finite portions of these \emph{graph lineages} can be used computationally,
and are chosen to have a strong space cost advantage over
conventional graph constructions while nevertheless having
similar algebraic and category-theoretic properties.
They can also be used to define
the continuum
limit of a growing sequence of graphs \cite{scott2021graph},
hence connecting to geometry. 
We note here that this definition of a continuous object is different than the \emph{graphon} of \cite{lovasz2012large}, in particular because graph lineages permit infinitely large {\it sparse} graphs. 
We will also briefly outline connections to hyperbolic geometry 
as a scale space (Section~\ref{contin_skeletal}).

Graph-related constructions of machine learning architectures \cite{maroninvariant} are increasingly popular, owing to the ability of these models to directly encode important invariances and equivariances in the data of interest. We propose to jointly model distributions of data and model architectures as {graph lineages}.
Their ``skeletal products'' (defined below)
can formalize space-efficient Neocognitron/CNN-style architectures
and could also 
efficiently organize a joint hierarchy of 
classical AI specialization (Is-A) and compositional (Part-Of)
hierarchical relationships between represented objects,
as outlined in the proposed neural architecture of Mjolsness et al. \cite{mjolsness1988neural}.

Graph functions based on graph box (Cartesian) products
have been discussed in \cite{ImrichKlavzar} (Appendix C.4),
but these were (a) restricted to be graph homomorphisms,
which are ours (Section~\ref{section:skeletal_function_space})
are not necessarily,
and (b) were not skeletonized and so would in general
be costly to implement.


    

\section{Preliminaries}

In this section we introduce notation and concepts necessary for later sections. We assume standard definitions of graph $G$, 
graph vertex and edge sets $V(G)$ and $E(G)$,
(edge-) weighted graph,
adjacency matrix
($A(G)$ or sometimes we just write $G$ which can be 0/1 valued or integer-valued), 
graph Laplacian 
$L(G) = G - \text{diag}(1 \cdot G)$ 
(for a graph or a nonnegatively weighted graph)
and graph spectrum $\lambda(L(G))$.
\footnote[2]{
This sign of the graph Laplacian is chosen
for compatibility with physical process models and 
with the original Laplacian operator, 
$\nabla^2 = \sum_i \partial_i^2$,
in the limit of large level number for grid graph refinement sequences.
Graph theorists, and users of exterior calculus, often reverse this sign of the Laplacian. }
Likewise for 
directed graphs (digraphs), 
bipartite graphs, and multigraphs.
Sometimes graphs will have ``self-loops'', i.e. edges from a vertex to itself.

\subsection{Graph sums and products}
\label{subsec:graph_prods}
\subsubsection{Disjoint Union, $\oplus$}

For graphs, the disjoint union $G_1 \oplus G_2$ is a graph with adjacency matrix
\begin{equation}
(G_1 \oplus G_2) = \Big(
\label{def_oplus}
\begin{array}{cc}
G_1 & 0 \\
0 & G_2
\end{array} \Big) .
\end{equation}

This formula is equally applicable to the case of undirected graphs,
for which the adjacency matrix is symmetric ($G=G^T$),
and directed graphs for which it need not be symmetric.
We will generally not consider the two cases separately.

The same relationship holds for the graph Laplacians
of undirected graphs.
Eigenvectors of $L(G_1 \oplus G_2)$ are concatenations of an eigenvector from one
graph and a zero vector for the other, so that the spectra are unioned together as sets:
$\lambda(G_1 \oplus G_2) = \lambda(G_1) \cup \lambda(G_2)$.

\subsubsection{Box product, $\Box$}

The graph box product $G_1 \Box G_2$ 
(also called the Cartesian product)
is a graph with adjacency matrix
\begin{equation}
(G_1 \Box G_2)_{(i a), (j b)} = G_{1;i, j} \delta_{a, b} +\delta_{i, j} G_{2;a, b} .
\label{box_prod_coordinates}
\end{equation}
Here $\delta_{a b}$ is the Kronecker delta (=1 iff $a = b$, otherwise 0) representing
the identity matrix in component notation.
This formula is equally applicable to the case of undirected graphs,
for which the adjacency matrix is symmetric ($G=G^T$),
and directed graphs for which it need not be symmetric.

For undirected graphs 
the combined graph Laplacian is (now using matrix rather than component notation)

\begin{equation}
L(G_1 \Box G_2) = L(G_1) \otimes I_2 + I_1 \otimes L(G_2)
\end{equation}

Eigenvectors of $L(G_1 \Box G_2)$ are vector outer products of an eigenvector from each
graph so that the spectra add in all possible ways:
$\lambda(G_1 \Box G_2) = \lambda(G_1) +  \lambda(G_2)$. See Fiedler \cite{fiedler1973algebraic}, Item 3.4 for further discussion of box product eigenvectors.

\subsubsection{Cross product, $\times$}
The graph cross product 
$G_1 \times G_2$ 
(also called the direct product)
is a graph with adjacency matrix
\begin{equation}
(G_1 \times G_2)_{(i a), (j b)} = G_{1;i, j} G_{2;a, b}.
\label{cross_prod_coordinates}
\end{equation}

The combined graph Laplacian is (using matrix rather than component notation)
\begin{equation}
L(G_1 \times G_2) = L(G_1) \times L(G_2)  + L(G_1) \times \text{diag}(1_1 \cdot G_2) + \text{diag}(1_2 \cdot G_1) \times L(G_2).
\end{equation}

Eigenvectors of $G_1 \times G_2$ (but not in general of their Laplacians!)
are products of an eigenvector from each
graph, so that the spectra multiply:
$$\lambda_{\text{adjacency}}(G_1 \times G_2) = \lambda_{\text{adjacency}}(G_1)  \lambda_{\text{adjacency}}(G_2).$$
{\it If} $G_1$ and $G_2$ each have constant degree $d_1$ and $d_2$, then the eigenvectors
of the Laplacians are outer products of the eigenvectors of the factors,
and $\lambda(G_1 \times G_2) = \lambda(G_1)  \lambda(G_2) + d_1  \lambda(G_2) +  d_2 \lambda(G_1) $.

The box and cross notations are also pictograms of the
respective products of the 2-cliques 
$K_2 \times K_2 = $ ``$\times$'', 
$K_2 \Box K_2 = $ ``$\Box$''.

\subsubsection{Growth rate of graph products}
Both graph products are defined using the Cartesian product of vertex sets,
which grows rapidly in computational resources potentially required
as more products are taken.  To mitigate this problem we ``skeletonize''
these products in Section~\ref{sec:skel_prods} below,
and compute the greatly reduced space cost in
Section~\ref{product_space_cost} .


\subsection{Graph lineages}
\label{subsec:graph_lin}

\subsubsection{Definition}
\label{defn:glineage}
We use the same definitions of \emph{graph sequence}, \emph{graded graph}, and \emph{graph lineage} as given in \cite{scott2021graph},
except that to permit graph sums we now allow for 
``unrooted'' (more accurately, multiply rooted) versions of each. 

A {\it graph sequence} is just a mapping from the 
natural numbers ${\mathbb N}$ into some set of graphs,
eg $l:\mathbb{N} \mapsto G_{l}$.
A {\it rooted graph sequence} obeys (1) $|V(G_0)|=|E(G_0)|=1$, i.e. $G_0$ is a near-minimal graph of one vertex and one self-loop. 
A 
(rooted or not) 
{\it hierarchical graph sequence} $\cal G$ 
is a 
(rooted or not) 
graph sequence which also obeys 
(2) The vertex (node) cardinalities $|V(G_l) |$  and the edge (link) cardinalities $|E(G_l)|$
are each bounded above 
by an exponential-like
growth function: 
They are
$O(b^{l^{1+\epsilon}})$, 
for some  $b:{\mathbb{R}} \geq 1$ 
and for all $\epsilon>0$.
A {\it graded graph} is a graph whose vertices are labeled by nonnegative integers
differing by at most one between vertices directly connected by an edge. 
One kind of example is an abstract cell complex with $k$-cells
labelled by dimension number $k:\mathbb{N}$,
and connected through the boundary relationship
which changes dimension $k$ by $\pm 1$,
but not by zero.
(In Section~\ref{sec:skel_prods} below we will give an alternative
definition of graded graphs in terms of graph homomorphisms,
which streamline our proofs.)
Clearly, the vertices and the $\Delta l =0$ edges in a graded graph form a graph sequence;
this graded graph sequence could be
hierarchical, and rooted or unrooted.
Also the $\Delta l = \pm 1 $ edges of
such 
a graded graph form a sequence of bipartite graphs.
If the number of edges between layers is also bounded above to grow (only) exponentially
as $O(b^{l^{1+\epsilon}})$ (for all $\epsilon > 0$),
then we have a ``graph lineage''.

{\bf Define} a 
(rooted or not) 
{\it graph lineage} as a graded graph whose vertices and
same-level ($\Delta l =0$) edges form a 
(rooted or not) 
hierarchical graph sequence, and whose
vertices and unequal-level ($\Delta l = \pm 1 $) edges form a hierarchical
graph sequence of bipartite graphs (each connecting some level $l$ to $l+1$).
The grade of any root nodes is defined to be zero.

\subsubsection{Examples}
\label{subsubsec:glin_ex}

In this section we illustrate several examples of graph lineages. In each of the figures in this section, solid lines indicate within-level ($\Delta l = 0$) edges, and dashed lines indicate intra-level ($\Delta l = 1$) connections. 

\paragraph{Path Graphs}

\begin{figure}
\centering
\begin{minipage}{.5\linewidth}
\centering
\includegraphics[width=\linewidth]{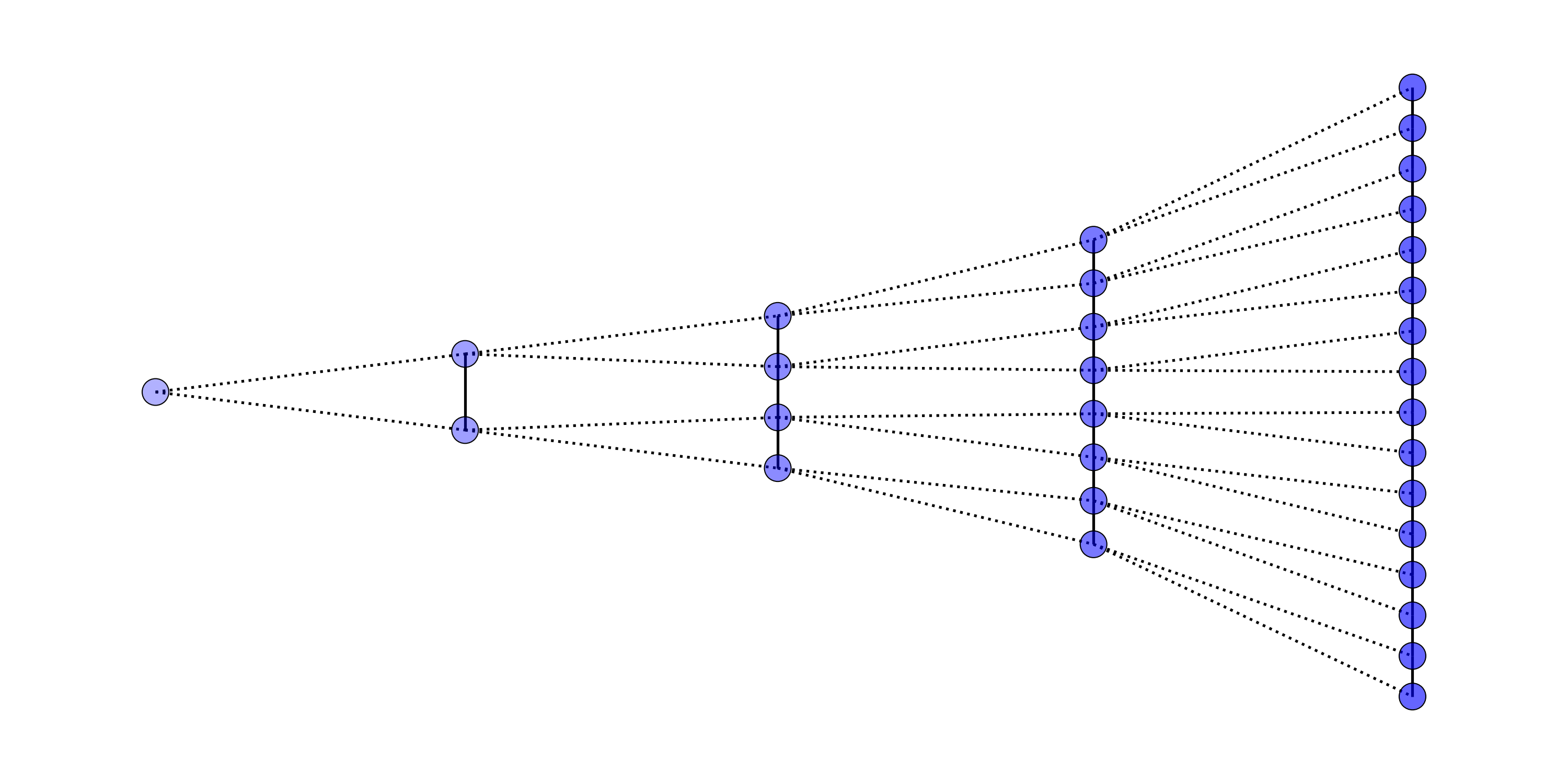}
\caption{A lineage of growing path graphs.}
\label{fig:glin_paths}
\end{minipage}\hfil
\begin{minipage}{.5\linewidth}
\centering
\includegraphics[width=\linewidth]{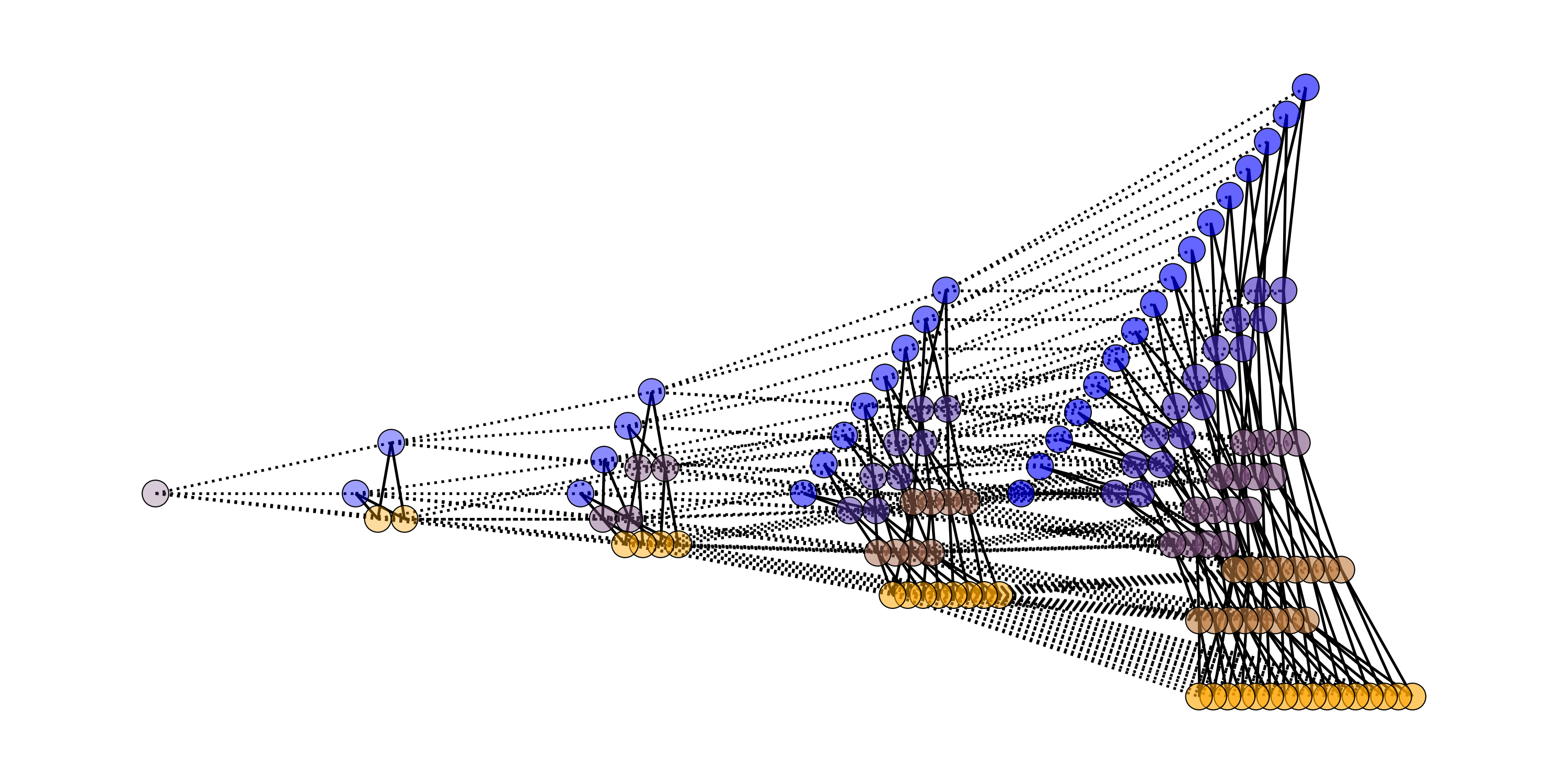}
\caption{A lineage of growing ``butterfly network'' graphs, 
like those used in the Fast Fourier Transform.}
\label{fig:glin_fourier}
\end{minipage}\\
\begin{minipage}{.4\linewidth}
\centering
\includegraphics[width=\linewidth]{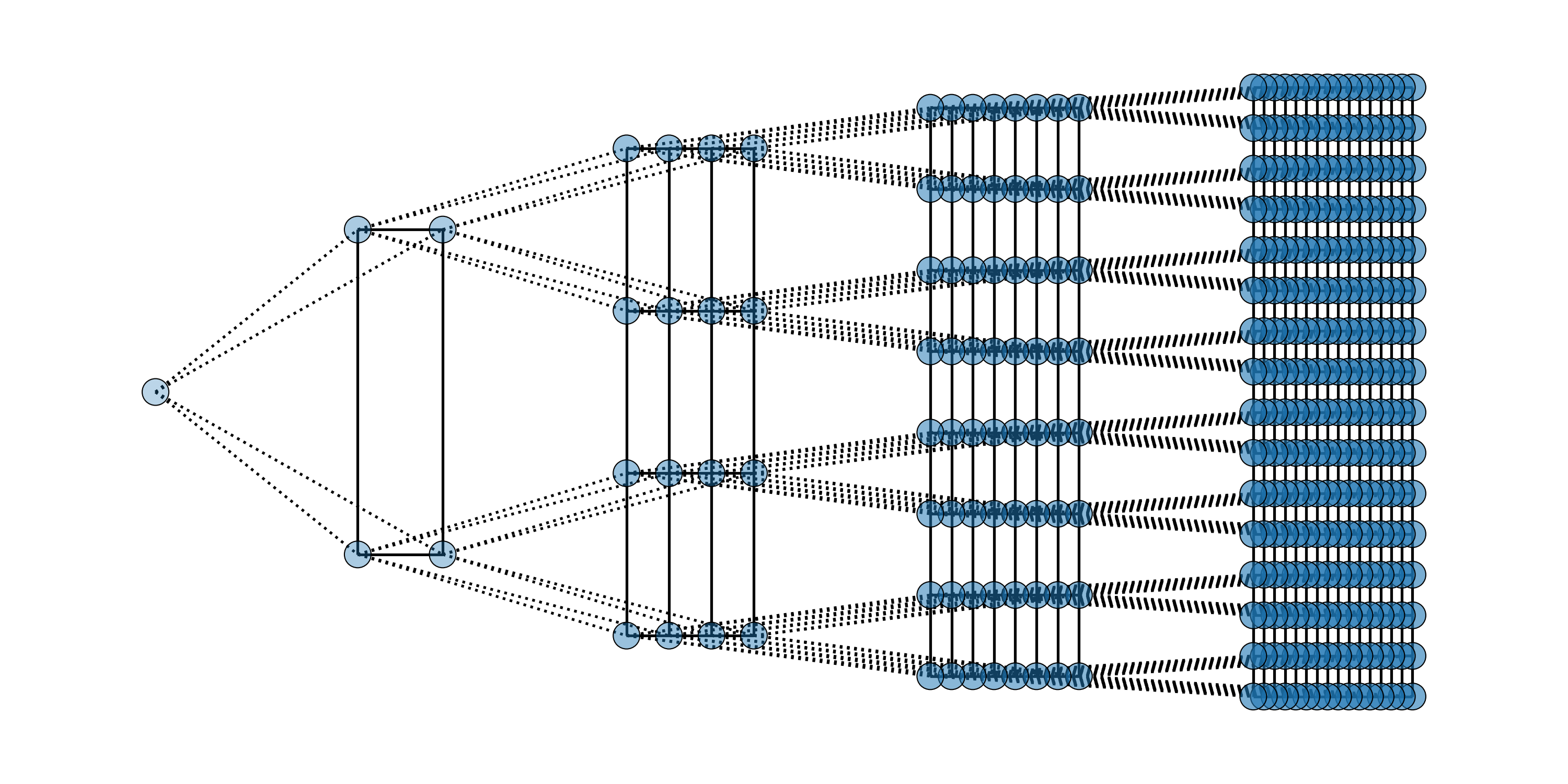}
\caption{A lineage of growing grid graphs.}
\label{fig:glin_grids}
\end{minipage}\hfil
\begin{minipage}{.5\linewidth}
\centering
\includegraphics[width=\linewidth]{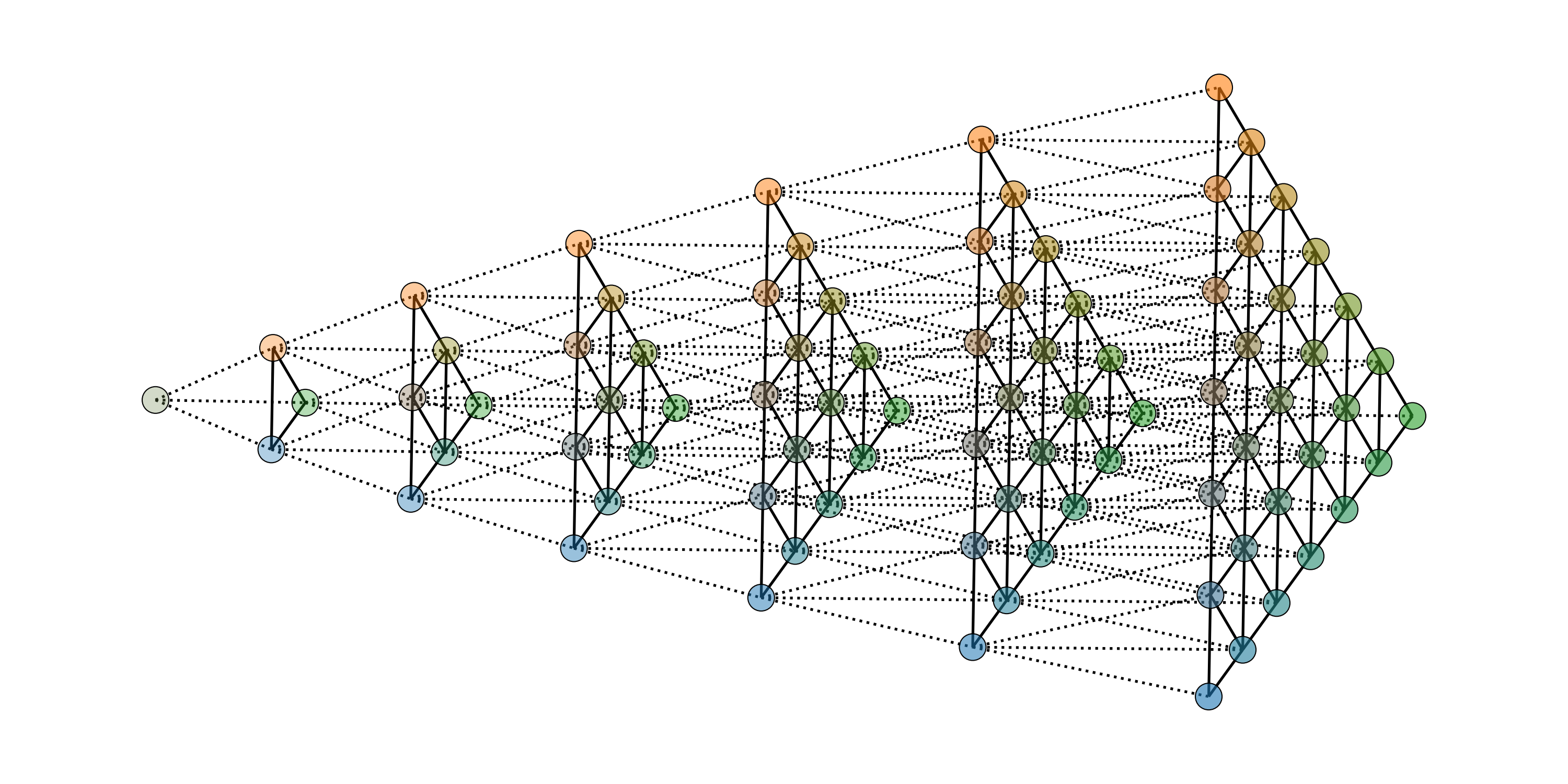}
\caption{A lineage of growing triangular grid graphs.}
\label{fig:glin_tri_grids}
\end{minipage}%
\end{figure}

See Figure \ref{fig:glin_paths}.

\paragraph{Butterfly Network Graphs} 
These are in the Fast Fourier Transform.
See Figure \ref{fig:glin_fourier}.
\paragraph{Grid Graphs}
%
Rectangular and triangular grids in nD. See Figure \ref{fig:glin_grids}, \ref{fig:glin_tri_grids} for examples of 2D grids.
\cite{scott2021graph} .

\section{Category theory of graph lineages}
\label{sec:cat_thy_graphs}
The graph lineages mentioned above can also be justified 
and related to abstract algebra
by category-theoretic constructions in the category of graphs. Here we redefine these objects in category theory terms. This will also motivate our construction (in Section \ref{sec:skel_prods}) of several operations on graph lineages which inherit useful properties of corresponding operations on graphs. We assume some 
elementary
knowledge of category theory in this section; we refer the reader to \cite{awodey2010category} for a discussion of these concepts.  

\subsection{The ``Graph'' category}
\label{subsec:graph_cat}
There is a category ``Graph'' of graphs and edge-preserving graph homomorphisms.
In fact there are two such categories, for directed and undirected graphs,
which are related.  
We will take directed graphs as the default and 
regard undirected graphs as directed graphs satisfying the condition that
every pair of nodes $x$ and $y$ having an edge from $x$ to $y$ also has an edge from $y$ to $x$. 
They form a sub-category, again with edge-preserving graph homomorphisms,
and all our constructions go through in this sub-category as well.
Alternatively following \cite{KnauerAlgGraphs} one can use ``graph'' for both,
only distinguishing when necessary.
Either way we allow self-edges from node $x$ to $x$.
There is a forgetful functor from Graph to Set, the category of sets and functions from sets to sets,
which maps each graph to the set of its vertices and each graph homomorphism 
to the corresponding function from graph vertices to graph vertices.

\subsection{Categorical products of graphs}
\label{subsec:cat_thy_gr_prod}
The graph cross product $G_1 \times G_2$ of graphs 
is the categorical product $G_1 \otimes_{\text{Gr}} G_2$ in the category of
graphs and edge-preserving graph homomorphisms.
Therefore it has the universal diagram property:
There are projection maps $\pi_a$ from $G_1 \otimes_{\text{Gr}} G_2$ onto $G_a$,
and any other graph 
$X$
with such projection maps $\pi^{\prime}_a$ 
has an induced graph homomorphism to $G_1 \otimes_{\text{Gr}} G_2$
which when composed with $\pi_q$ equals  $\pi^{\prime}_a$.
Thus Diagram \ref{diagram_cat_prod} commutes.
\begin{figure}
    \centering
    \begin{minipage}{0.47\textwidth}
    \centering
\begin{tikzcd}
    & X \arrow[ld , "\pi'_1" above] \arrow[rd, "\pi'_2" above] \arrow[d, dotted, "\chi" right]     &     \\
G_1 & G_1 \otimes_\text{Gr} G_2 \arrow[r, "\pi_2" above] \arrow[l, "\pi_1" above] & G_2
\end{tikzcd}
\captionof{diagfig}{Commutative diagram for the categorical or cross product of two graphs. $\pi_1$ and $\pi_2$ are projections to the separate graph factors.
The universality property is given by the induced dotted-line
graph homomorphism $\chi$ from any other graph $X$ that
also has graph homomorphisms to the graph factors.}
\label{diagram_cat_prod}
    \end{minipage}
    \hfill
    \begin{minipage}{0.47\textwidth}
    \centering
\begin{tikzcd}
E_1 \times E_2 \arrow[rd, "\text{bimorphism}" {sloped,below}, "\xi" {sloped,above}] \arrow[rr,"\text{bimorphism}" {below}, "\tau" {above} ] &   & G_1 \Box G_2 \arrow[ld, dotted, "\xi^*" {sloped,above}] \\
                                     & X &                        
\end{tikzcd}
\captionof{diagfig}{Categorical diagram for $\Box$, the graph tensor product,
following [Kauer 01].
Graph homomorphism $\xi^*$ is induced by bimorphisms $\tau$ and $\xi$,
each a graph homomorphism on $G_1$ for any fixed vertex of $G_2$ and vice versa.}
\label{diagram_tensor_prod}

    \end{minipage}
\end{figure}

The disjoint union $\oplus$ is the coproduct $\oplus_{\text{Gr}}$ in the category of graphs.
It has the dual universal diagram to that of $\otimes_{\text{Gr}}$ in the category of graphs: all arrows are reversed.
There are actually several plausible categories of graphs;
we accept vertex loops (self-edges) in our undirected graphs.

The Cartesian or ``box'' product of two graphs (\cite{ImrichKlavzar}), on the other hand, 
is the natural way to combine independent processes acting on different spaces:
$W = \ W_1 \Box W_2$, since $L(G_1 \Box G_2) = L(G_1) \Box L(G_2)$ as previously calculated.

The box product also has a categorical interpretation as a {\it tensor product} in a suitable graph category
(\cite{KnauerAlgGraphs},
Theorem 4.3.5), 
either the category of graphs and graph homomorphisms
which we will use,
or the category of graphs and maps that either preserve or contract edges.
The latter category also has as categorical product the ``strong graph product'' or ``box-cross'' product
consisting of the edges in the union of box and cross products.

The tensor product diagram is shown in Diagram~\ref{diagram_tensor_prod}.
In this diagram,
$E_i$ is the set of edges in graph $G_i$. 
(The ``tensor product''
name can be justified by the fact that the tensor product of vector
spaces has a similar universal diagram, with ``bimorphism''
replaced by ``bilinear map''.)
This diagram
works for any concrete category (a category with a faithful functor to the category of sets)
so that the Cartesian product of sets is defined,
so it works for the category of graphs and graph homomorphisms.
Acting on graph vertices, $\tau$ is the identity mapping from $V_1 \times V_2 \rightarrow G_1 \Box G_2$.
Acting on graph edges $\tau$ is a ``bimorphism'':
holding fixed a vertex $y$ in $G_2$, $\tau$ is a homomorphism from $G_1$ to $G_1 \Box G_2$,
and vice versa. (This is a less common use of the term,
but it is used in \cite{KnauerAlgGraphs}
which we are following for the definition of tensor product.)
The universal diagram means that if $\xi$ is another bimorphism to another graph $X$,
then $\xi^* = \xi \circ \tau^{-1}$ exists and is a graph homomorphism from $G_1 \Box G_2$ to $X$.

An alternate characterization of the graph box product, 
which also uses the universal diagram for tensors, 
is as the \emph{funny tensor product}
\cite{NLab_FTV},
in the category of graphs 
whose morphisms map paths to paths,
as if the graph were itself a category,
rather than edges to edges.
It is one of two categorical graph product
constructions, the other being the cross product
(cf. \cite{NLab_FTV, foltz1980algebraic}).

\section{Skeletal graph products}
\label{sec:skel_prods}

We will define new operations on graph lineages by way
of the definitions of the operations on ``graded graphs''
that include the structure of witness sparse prolongation maps $P$
compatible with a bipartite graph with adjacency matrix $S$.
These maps can be dropped or replaced,
if desired, to get the corresponding
operations on (hierarchical) graph sequences. In each of the following sections, we will motivate an operation on graded graphs by first describing the operation's action on the 
foundational graded graph $\hat{\mathbb N}$ of the natural numbers
and their successor relationship, as we will define and illustrate below. 

\subsection{Graded graphs}
\label{gradedgraphsnewops}

The componentwise box and cross products of two graph lineages
${\cal G}_1$ and ${\cal G}_2$ are graph lineages
whose vertex cardinality grows as $O([\text{base}(G_1) \text{base}(G_2)]^l)$. 
Iteration of this product operation will have a high computational cost since the bases multiply.
We seek slower growth, closer to 
$O(l^p b^l )$
or $\log(|\text{vertices}|) \sim l \log (b)$ where $b=\max(\text{base}(G_1), \text{base}(G_2))$,
possibly at the cost of some form of approximation.
To this end we consider how to build level $l$ of a ``skeletal'' graph product ${\cal G}$ out of
combinations of graphs of generation number pairs ($l_1$ of ${\cal G}_1$  , $l_2$ of ${\cal G}_2$ )
where $l = l_1 + l_2$, or more generally $l = \sum_a \alpha_a l_a$ where $\alpha_a >0$.

For a (hierarchical) graph lineage $\cal G$, {\bf define} the associated (infinite) graded graph $Gd({\cal G})$
that labels each vertex in $\oplus_{l=0}^{+\infty} G_l$ with its level number $l$ and connects
vertices in adjacent levels $(l,l+1)$ iff they have nonzero entries in the prolongation matrix
$P^{(l+1,l)}$ that optimizes
a graph-graph distance measure
$D_R$ \cite{scott2021graph}. The root node is defined to have level number $l=0$.
The ``pure graded graph'' associated to $\cal G$
keeps the intra-level connections but omits all the inter-level connections.

A graded graph $Gd({\cal G})$ can alternatively and more abstractly
be {\bf defined} as a pair $(G, \varphi_G)$
where $G$ is a graph and $\varphi_G$ is a graph homomorphism
(a vertex mapping that preserves edges)
$\varphi_G : G \rightarrow \hat{\mathbb{N}}$
from $G$  to the special graph $\hat{\mathbb{N}}$
whose vertices are nonnegative integers $n:\mathbb{N}$
and whose edges are self-loops at each integer $n$ together
with predecessor/successor pairs $\{n,n+1\}$
in the case of undirected graphs, or ordered pairs $(n,n+1)$ otherwise.


The equivalence to the previous definition is straightforward:
edges are only allowed if they map to the same or successive natural numbers.
With this definition, a {\it homomorphism of graded graphs} is a graph homomorphism
$h: G \rightarrow G^{\prime}$ that makes a commutative diagram $\varphi_G = h \circ \varphi_{G^{\prime}}$ (Diagram \ref{diagram_graded_graph_hom}),
\begin{figure}
\begin{minipage}{.47\textwidth}
\centering
\begin{tikzcd}
G \arrow[rd, "\varphi_G" {sloped,above}] \arrow[rr, "h" {above} ] &   & G' \arrow[ld, "\varphi_{G'}" {sloped,above}] \\
                                     & \hat{\mathbb{N}} &                        
\end{tikzcd}
\captionof{diagfig}{The commutative diagram for a graded graph homomorphism.}
\label{diagram_graded_graph_hom}
\end{minipage}
\hfill
\begin{minipage}{0.47\textwidth}
\centering
\begin{tikzcd}[execute at end picture={\path (\tikzcdmatrixname-1-1) node[below right=.25 and .25]{\Huge $\lrcorner$};}]
\theta(G) \arrow[rr, "p" above] \arrow[dd, "f" left] &  & G \arrow[dd, "\varphi_G" left]     \\
                                             &  &                  \\
\theta(\hat{\mathbb{N}}) \arrow[rr, "\pi_{\theta(\hat{\mathbb{N}})}" below]          &  & \hat{\mathbb{N}}
\end{tikzcd}
\captionof{diagfig}{Pullback diagram, in the category of graphs, for the thickening operator $\theta$}
\label{diagram_thick_operator}
\end{minipage}
\end{figure}
and thus graded graphs
together with level-preserving graph homomorphisms form a slice category. 
Of course, $(\hat{\mathbb{N}}, id_{\hat{\mathbb{N}}})$ is a graded graph.

We can decompose a graded graph into its core (the same-level edges)
and its halo (the unequal-level edges). The core gives a graph sequence.
The halo can be used as the structure for a space of candidate
sparse prolongation matrices between successive generations in
the graph sequence. Thus there is a mapping 
${\cal G} \mapsto G$
from graded graphs to graph sequences.

So long as prolongation maps are optimized over
compact manifolds of matrices, 
such as orthogonal matrices of a given structure,
then ``inf'' is ``min'' and there
is a right-inverse map from graph sequences to graded graphs as well.
This mapping converts each prolongation matrix
$P^{l+1,l}$ between adjacent levels into a 0/1-valued
adjacency
matrix $S^{l+1,l}$ representing the sparsity structure of $P$,
and adds those edges as the halo of ${\cal G}$.
The core of ${\cal G}$ is just the
union of all the edges in
the graph sequences member graphs,
since we assume the vertices of different member graphs are disjoint.
Assuming each vertex in ${\cal G}$ is {\it reachable} from
some
$l=0$ root vertex, then each $S^{l+1,l}$ also records a set-covering
of the vertices $G_{l+1}$ indexed by those of $G_l$.

\subsection{Thickening operator, $\theta$}

In like manner to skeletal products 
(sections~\ref{subsec:skel_cross_prod} and \ref{skeletal_box_product} below),
we define the unary ``thickening'' operator
first by its effect on $\hat{\mathbb N}$ and then for all other graded graphs by
a universal diagram - in this case, a pullback.

The effect of the thickening operator 
$\theta$
on $\hat{\mathbb N}$ is as follows: 
The graded graph $\theta(\hat{\mathbb N})$ has vertices
$\{ (l,i) | l,i: {\mathbb N} \quad \wedge \quad i \leq l \}$. 
It has grading $\varphi(l,i)=l$.
It has edges 
from vertex $(l^\prime, i^\prime)$ to $(l,i)$
wherever $\hat{\mathbb N} \Box \hat{\mathbb N}$
would, i.e. for $l = l^{\prime} $ and $i =  i^{\prime} \pm 1$ (within a grade)
or $l = l^{\prime} \pm 1 $ and $i = i^{\prime} $ (between grades)
in the case of undirected graphs, and
for $l^{\prime} = l $ and $i= i^{\prime} + 1$ (within a grade)
or $l = l^{\prime} + 1 $ and $ i^{\prime} =i $ (between grades)
in the case of directed graphs.
In addition to the grading $\varphi_{\theta(\hat{\mathbb N}) } $, there is another graph homomorphism
$\pi_{\theta(\hat{\mathbb N})}$ from $\theta(\hat{\mathbb N})$ to $\hat{\mathbb N}$,
where $\pi(l,i)=i$, which will be used below.
In component notation, 
$\theta(\hat{\mathbb N})$ is
\begin{equation}
\label{theta_N_components}
\begin{split}
V(\theta(\hat{\mathbb N})) &= \{ (l:{\mathbb N}, i:{\mathbb N}|_{i \leq l} ) \} \\
G[\theta(\hat{\mathbb N})]_{(l, i) \; (l^\prime,i^{\prime} ) }  
&= \delta_{l \; l^{\prime}} \; (\delta_{i \; i^{\prime}}+ \psi_{i \; i^{\prime}}) \\
S[\theta(\hat{\mathbb N})]_{(l, i) \; (l^\prime,i^{\prime} ) } 
&=  \psi_{l\; l^{\prime}} \; \delta_{i \; i^{\prime}}   \quad , \\
\end{split}
\end{equation}
and
$\hat{\mathbb N}$ is
\begin{equation}
\label{N_components}
\begin{split}
V(\hat{\mathbb N}) &= \{ i:{\mathbb N}  \} \\
G[\hat{\mathbb N}]_{i \; i^{\prime} }  
&= \delta_{i \; i^{\prime}} \\
S[\hat{\mathbb N}]_{i \; i^{\prime}}
&=  
\psi_{i \; i^{\prime}}  \quad . \\
\end{split}
\end{equation} 
Here $\psi_{n \; n^\prime} = \delta_{n \; n^\prime +1}$ 
in the category of directed graphs,
and $\psi_{n \; n^\prime} = \delta_{n \; n^\prime +1} + \delta_{n+1 \; n^\prime}$
in the category of undirected graphs
so these adjacency matrices all become symmetric.

\begin{figure}
    \centering
    \includegraphics[width=0.5\linewidth]{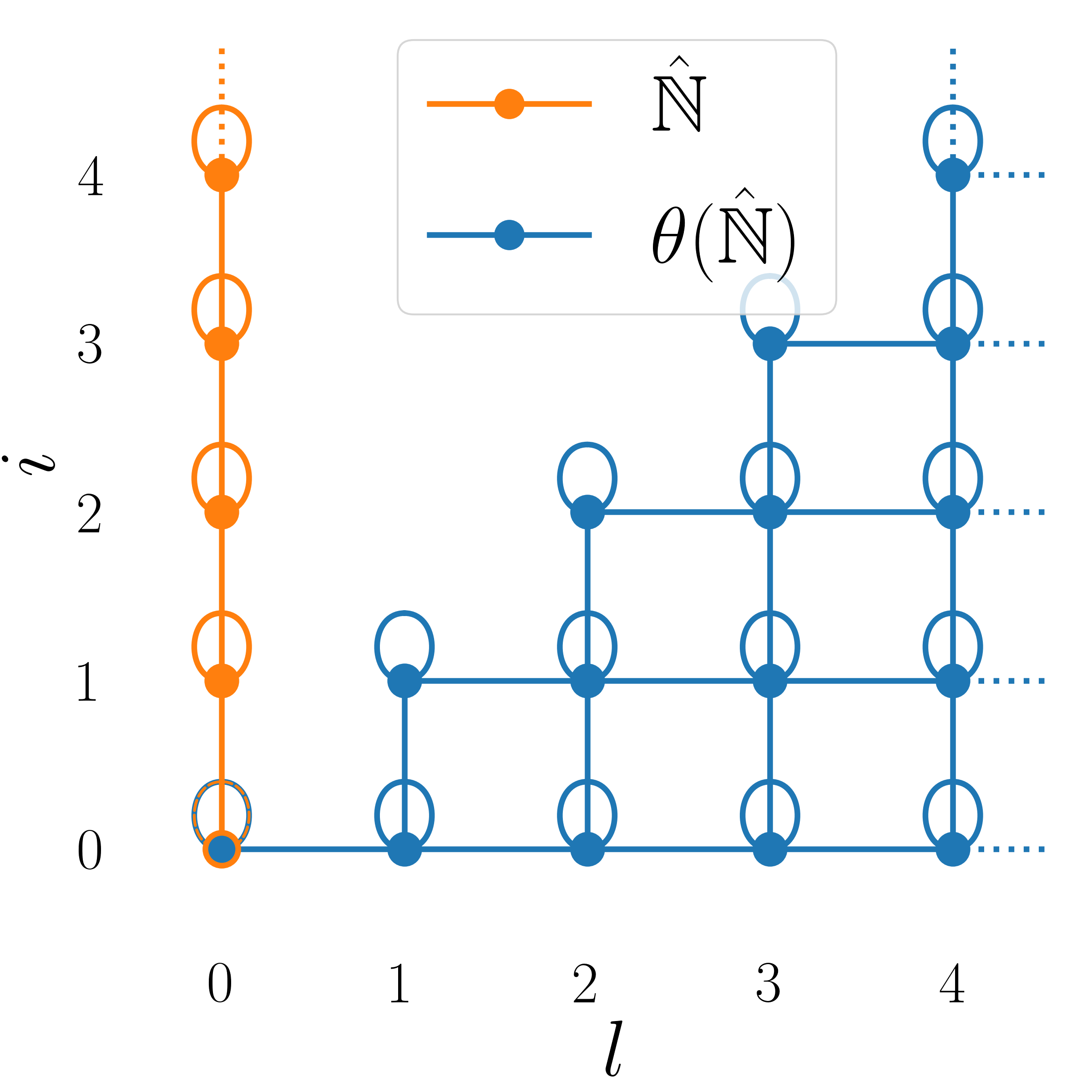}
    \caption{Graphs $\hat{\mathbb N}$ (indexed by grading number $i$ 
    in this illustration and in Equation~\ref{N_components}) and
    $\theta(\hat{\mathbb N})$ (indexed by both $l$ and $i$ here and in Equation~\ref{theta_N_components}), though now $l$ is the grading number.) 
    Undirected graph case is illustrated; otherwise directed edges ``point'' towards increasing index values. 
    This special case also serves as the template for the thickening operation in general,
    by way of a graded graph's level numbers.
    General graph lineages could in principle be added into this figure
    by adding sets of additional orthogonal axes $\{ j_i \}$, 
    that are projected out by $f$ and $\varphi_G$ here.
    Then intralevel graphs $G$ would project in these orthogonal directions
    to self-loops and to single horizontal edges,
    and interlevel graphs $S$ would project to to single vertical edges here.}
    \label{fig:thick_lineage_illus}
\end{figure}

Note that $\pi_{\theta(\hat{\mathbb N})}: (l,i) \mapsto i$ maps 
$\delta_{i \, i^\prime}$ from the second and third lines
of Equation~(\ref{theta_N_components}) to 
the second line of Equation~(\ref{N_components}), and
$\psi_{i \, i^\prime}$ from the second line
of Equation~(\ref{theta_N_components}) to the third line of Equation~(\ref{N_components}).
See illustration in Figure~\ref{fig:thick_lineage_illus}.

{\bf Definition.} 
The {\it thickening operator} $\theta$ 
is extended from $\hat{\mathbb N}$
to all other graded graphs $G$
(and consistently with the foregoing definition in the special case of $G = \theta(\hat{\mathbb N})$),
thus defining $\theta : G \mapsto \theta(G)$,
by the commutative Diagram 
\ref{diagram_thick_operator}
in the category of graphs,
universal in the $\theta(G)$ position.
This {definition} uses $\theta(\hat{\mathbb N})$
as a template for all other $\theta(G)$.

Such a universal diagram is a ``pullback'', indicated by the extra right-angle notation 
in the upper left of the square.
Since the lower horizontal arrow is $\pi$ rather than
$\varphi$, this diagram does not hold in the category of graded graphs.

{\bf Proposition 1}: The graded graph $\theta(\hat{\mathbb N}) $ exists, and
has formula in component notation (with Kronecker delta functions):
\begin{equation}
\label{theta_components}
\begin{split}
V(\theta(G)) &= 
\{ (l:{\mathbb N}, i:{\mathbb N}|_{i \leq l}, j_i : \{1, \ldots |V(G^{(i)})| \} ) \} \\
G[\theta(G)]_{(l,i, j_i) \; (l^{\prime},i^{\prime}, j^{\prime}_{i^{\prime}})}  
&= 
\delta_{l \; l^{\prime}} \;
G[\theta(G)]^{(l)}_{(i, j_i) \; (i^{\prime}, j^{\prime}_{i^{\prime}})} \\
G[\theta(G)]^{(l)}_{(i, j_i) \; (i^{\prime}, j^{\prime}_{i^{\prime}})}
& =
\big[ \delta_{i \; i^{\prime}} G^{(i)}_{(j_i, j^{\prime}_{i^{\prime}})} 
    + \delta_{i \; i^{\prime}+1} S^{(i, i^{\prime})}_{(j_i,  j^{\prime}_{i^{\prime}})}
        + \delta_{i+1 \; i^{\prime}} S^{(i, i^{\prime})}_{(j_i,  j^{\prime}_{i^{\prime}})}
    \big] \Theta(i \le l) \\
S[\theta(G)]_{(l,i, j_i) \; (l^{\prime},i^{\prime}, j^{\prime}_{i^{\prime}})}
&=  
S[\theta(G)]^{(l, l^\prime)}_{(i, j_i) \; (i^{\prime}, j^{\prime}_{i^{\prime}})}
= \psi_{l\; l^{\prime}} 
    \delta_{i \; i^{\prime}}  \; G^{(i)}_{(j_i, j^{\prime}_{i^{\prime}})} 
  \Theta(i \le l)  \; , \\
\end{split}
\end{equation}
where the corresponding formula for assembly of the
graded graph $(G,S)$ from its constituent levelwise graphs 
$G^{(i)}$ and $S^{(i, i^{\prime})}$ is, 
in component notation:
\begin{equation}
\label{G_components}
\begin{split}
V(G) &= \{ (i:{\mathbb N},  j_i  : \{1, \ldots |V(G^{(i)})| \}  ) \} \\
G_{(i, j_i) \; (i^{\prime}, j^{\prime}_{i^{\prime}})} &= \delta_{i \; i^{\prime}} G^{(i)}_{(j_i, j^{\prime}_{i^{\prime}})} \\
S_{(i, j_i) \; (i^{\prime}, j^{\prime}_{i^{\prime}})} &=  
    \psi_{i \; i^{\prime}} 
    S^{(i, i^{\prime})}_{(j_i,  j^{\prime}_{i^{\prime}})} 
    \; ,
\end{split}
\end{equation}
and $\Theta(\text{true}) =1, \Theta(\text{false})=0$.
Note that now $p: (l,i,j_i) \mapsto (i,j_i)$ maps 
$G^{(i)}_{(j_i, j^{\prime}_{i^{\prime}})}$ from the third and fourth lines
of Equation~(\ref{theta_components}) to 
the third line of Equation~(\ref{G_components}), and
$S^{(i, i^{\prime})}_{(j_i,  j^{\prime}_{i^{\prime}})}$ from the second line
of Equation~(\ref{theta_components}) to the third line of Equation~(\ref{G_components}).

{\it Proof of Proposition 1}: We will follow this outline:

\begin{enumerate}
\item We will first prove that the given indexing of vertices
and the given formula for their edges satisfies 
Diagram~\ref{diagram_thick_operator}. 
\item Considering any other rival occupant $X$ for
the position of $\theta(G)$ in Diagram \ref{diagram_thick_operator}, 
show that the vertices of $X$ can be indexed by
$V(X) =  \{ (l:{\mathbb N}, i:{\mathbb N}|_{i \leq l}, j_i : \{1, \ldots |V(G^{(i)})| \} ) , a \} $
where the domain of $a$ depends on $(l,i,j_i)$.
The maps from vertex set of $X$ to those of $G$ and $\theta({\mathbb N})$ preserve
the indices of the target vertex sets.
This indexing also establishes a mapping from vertices of $X$ to those of $\theta(G)$.
\item Show that the edges of $X$ are at most those
compatible with the formula for the edges of $\theta(G)$,
regardless of the index $a$.
\item Thus, establish a graph homomorphism from $X$ to $\theta(G)$, QED.
\end{enumerate} 

Step (1) is clearly satisfied by the component formulae,
by (a) projecting out the indices $l$ and $l^\prime$ in the horizontal direction
of Diagram~\ref{diagram_thick_operator}
($(l,i) \mapsto i$, obtaining Equation~(\ref{G_components}) from (\ref{theta_components}),
where each $G$ maps to a $G$ and each $S$ maps to an $S$,
and (\ref{N_components}) from (\ref{theta_N_components}));
and also (b) projecting out $j_i$ and $j^{\prime}_{i^{\prime}}$ in the vertical direction
($(i, j_i) \mapsto i$, obtaining Equation~(\ref{N_components}) from (\ref{G_components}) 
and (\ref{theta_N_components}) from (\ref{theta_components}),
where each $\delta_{i \; i^{\prime}}$ maps to a $\delta_{i \; i^{\prime}}$
and each 
 $\psi_{i \; i^{\prime}}$ maps to a $\psi_{i \; i^{\prime}}$.
In the horizontal direction projection from $\theta(G)$ to $G$ 
or $\theta(\hat{\mathbb N})$ to $\hat{\mathbb N}$, 
edges between corresponding nodes of different copies of the same graph 
get mapped into self-loop edges.

Step (2): The indexing of vertices in Diagram~\ref{diagram_thick_operator} 
is as shown in Diagram~\ref{diag:thick_operator_indexing}.
\begin{diagfig}
\centering
\begin{tikzcd}
\{ (l:{\mathbb N}, i:{\mathbb N}|_{i \leq l}, j_i : \{1, \ldots |V(G^{(i)})| \} ) \}  \arrow[rr, "p" above] \arrow[dd, "f" left] &  & \{ (i:{\mathbb N}, j_i : \{1, \ldots |V(G^{(i)})| \} ) \} \arrow[dd, "\varphi_G" left]     \\
                                             &  &                  \\
\{ (l:{\mathbb N}, i:{\mathbb N}|_{i \leq l} ) \}  \arrow[rr, "\pi_{\theta(\hat{\mathbb{N}})}" below]          &  & \{ (i:{\mathbb N} ) \}
\end{tikzcd}
\caption{The indexing of the vertices in Diagram \ref{diagram_thick_operator}.}
\label{diag:thick_operator_indexing}
\end{diagfig}

Consider instead the vertex set of $X$ to occupy the upper left corner of Diagram \ref{diag:thick_operator_indexing}.
Then for any fixed value of index $ i:{\mathbb N}$ in the lower right corner,
there is an inverse image $\phi_G^{-1}$ of vertices of $G$
indexed by $ (i:{\mathbb N}, j_i : \{1, \ldots |V(G^{(i)})| \} )$
in the upper right corner, and an inverse image $\pi_{\theta(\hat{\mathbb N})}^{-1}$
of vertices indexed by $(l:{\mathbb N}|_{l \geq i}, i:{\mathbb N} )$
in the lower left. 
In the category of sets, the pullback of mappings 
$\varphi_G$ and $\pi_{\theta(\hat{\mathbb{N}})}$
is the union over the lower right corner set $i:{\mathbb N}$ of the cross product
of the inverse images of each $i$ under the two mappings, i.e.
it is 
\begin{equation}
\begin{split}
\bigcup_{i:{\mathbb N}} & \{ (l:{\mathbb N} | {i \leq l},i) \}
\times  \{ (i:{\mathbb N}, j_i : \{1, \ldots |V(G^{(i)})| \} ) \}
\\ & 
\simeq
 \{ (l:{\mathbb N} , i:{\mathbb N}_{i \leq l}, j_i : \{1, \ldots |V(G^{(i)})| \} ) \}
 \end{split}
\end{equation}
which is the upper left corner of Diagram~\ref{diag:thick_operator_indexing}.
Here ``$\simeq$'' denotes an isomorphism in the category of sets, i.e. a 1-1 onto mapping.

Step (3) is forced by the graph homomorphisms in the variant
of Diagram \ref{diagram_thick_operator} obtained by replacing the upper left corner with
graph $X$. Step (4) is then automatic, completing the proof.

\vspace{0.1 in}

An important feature of Diagram \ref{diagram_thick_operator} is that it is not a graded graph
diagram, because the horizontal arrows change the
axis used for the grading from index $l$ to index $i$.
That is how each grade of $\theta(G)$
comes to include copies of not only the corresponding grade but also
all earlier grades of $G$, justifying the name ``thickening''
for this operation.

\begin{figure}
\begin{minipage}{.47\linewidth}
\centering
\includegraphics[width=\linewidth]{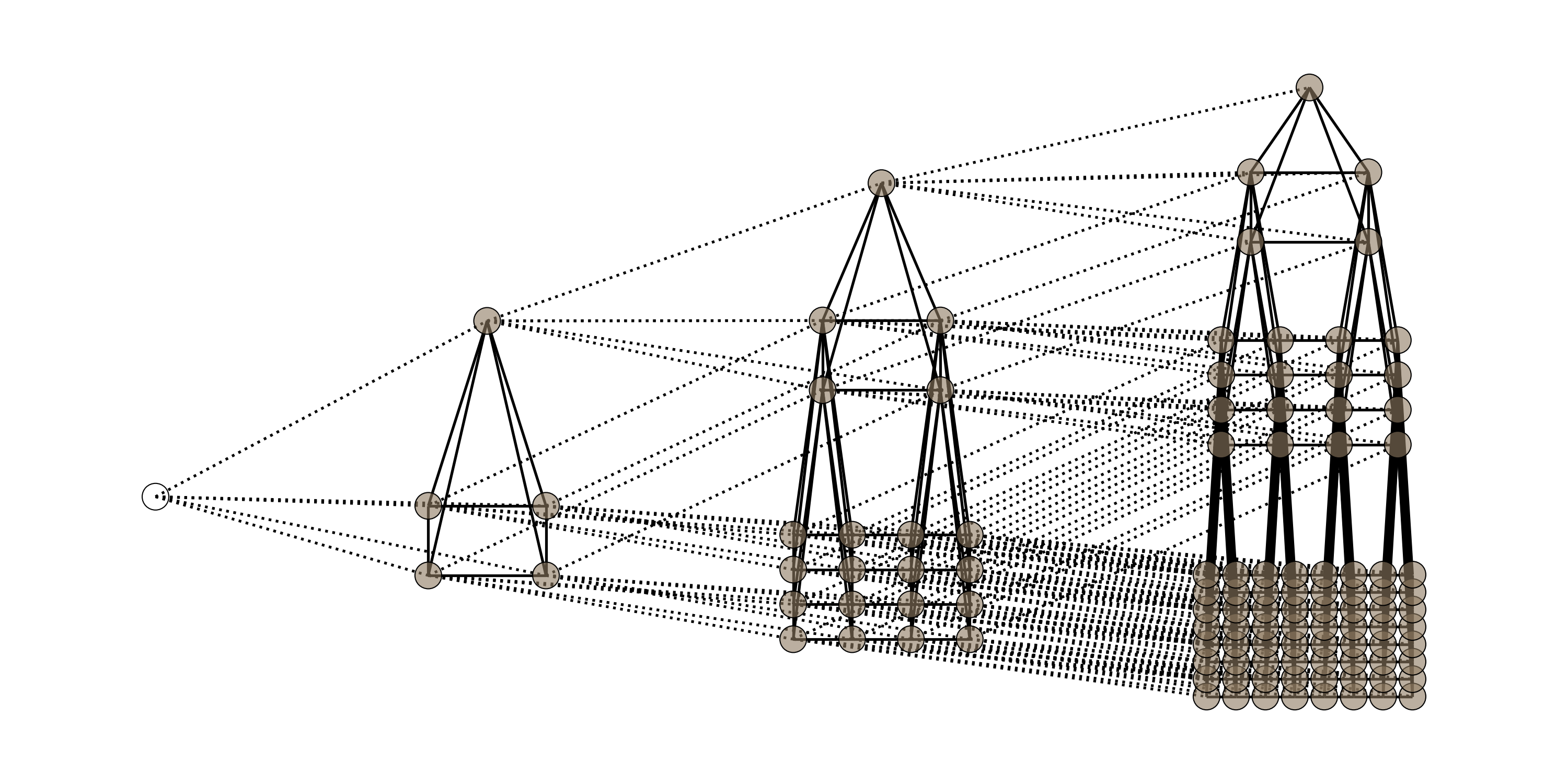}
\caption{A lineage of growing pyramid graphs produced by thickening the grid graph lineage in Figure \ref{fig:glin_grids}.}
\label{fig:glin_1x_pyramid}
\end{minipage}\hfil
\begin{minipage}{.47\linewidth}
\centering
\includegraphics[width=\linewidth]{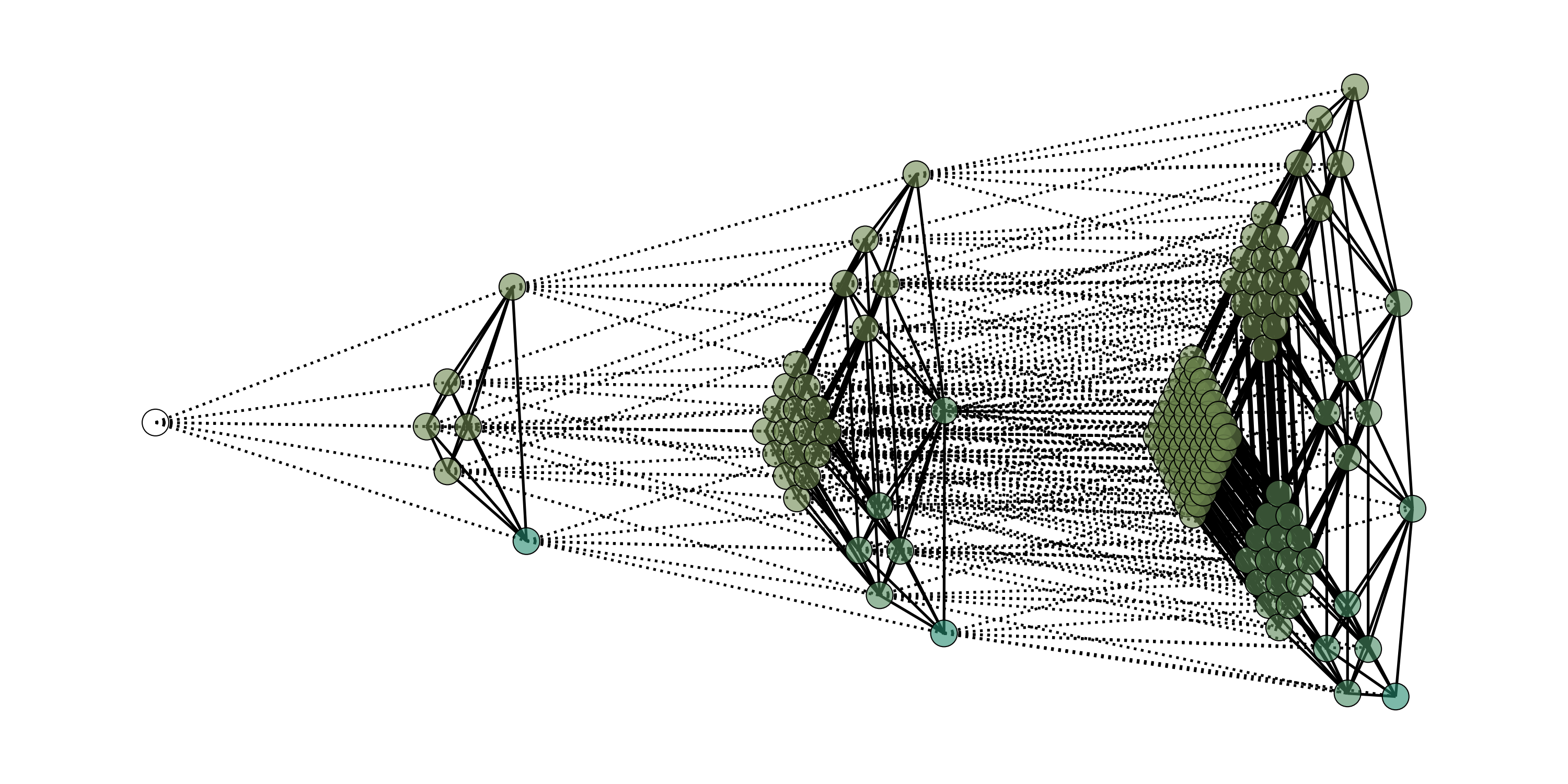}
\caption{A lineage of bipyramids produced by thickening the lineage in Figure \ref{fig:glin_1x_pyramid}, i.e. by thickening the grid graph lineage twice.}
\label{fig:glin_2x_pyramid}
\end{minipage}
\end{figure}

It is possible to iterate thickenings,
making successively thicker pyramids,
and to apply them to skeletal products,
making thicker bipyramids and multipyramids. 
See Figures \ref{fig:glin_1x_pyramid} and \ref{fig:glin_2x_pyramid} for examples of a pyramid and bipyramid lineage constructed by applying $\theta$ once and twice, respectively. 
The cost is extra factors of $l$ i.e. just logarithms of
the graph sizes,
so the bases $b_{\pm}^{v \text{ or } e}$ of cost exponentials
will increase by at most ``+ $\epsilon$'', and not at all
if they were already ``+ $\epsilon$''.
In particular the complexity class $\log |G| = o(n^{1+\epsilon})$ is
invariant under $n \rightarrow n \log n$ and is therefore
preserved under the thickening operation.

\subsection{Products of graded graphs}

We now seek analogs for the cross and box graph products in the case of graded graphs,
so as to better control the quadratic explosion in the space cost of the product of two graphs
and the exponential (in $n$) explosion in the space cost of the
product of $n$ graphs.

One idea would be to just define the (box or cross) product of two graded
graphs by starting with the corresponding product of two graphs without
the integer labels grading structure, and then to restore it somehow.
But how? And once the grading is reimposed, if it can be,
what properties can we expect of the resulting graded graph?
To answer these questions we formulate new diagrams related to the old ones
and also
related to the slice category structure  of graded graphs (Diagram \ref{diagram_graded_graph_hom}),
in the manner of Diagram \ref{diagram_thick_operator}.
We also aim to produce explicitly computable component-notation formulae for these products,
akin to Equations (\ref{box_prod_coordinates}) and (\ref{cross_prod_coordinates}).

The simplest grade level to assign to the vertex $(v_1,v_2)$ whose
vertices had grades $l_1, l_2$ would be $l=l_1+l_2$.
This additive rule will work for the box product to be defined below ({Proposition 2}),
but not the cross product due to the presence of edges that connect
levels $l$ and $l \pm 2$ for pairwise cross products. 
It will turn out ({Proposition 3}) that we can just truncate
these inconvenient edges.

Perhaps the most important property of the resulting generalizations
of the box and cross products relates to cost:
that the box and cross products of the graded graphs corresponding
to two graph lineages are again graph lineages, i.e. they have
controlled spatial cost as a function of level number $l$ ({Proposition 4}).
The additive rule counts exponential cost more accurately than $l = \max(l_1,l_2)$ would,
since $b^{l_1} b^{l_2}  = b^{l_1+l_2} $,
permitting much better cost control when truncating at finite $l \le L$
as one needs to do in applications.
Intermediate rules between sum and max
could include $l^p \simeq l_1^p + l_2^p$ as will be discussed.

\subsection{Skeletal cross product, $\hat{\times}$}
\label{subsec:skel_cross_prod}

\begin{figure}
    \centering
    \includegraphics[width=0.7\linewidth]{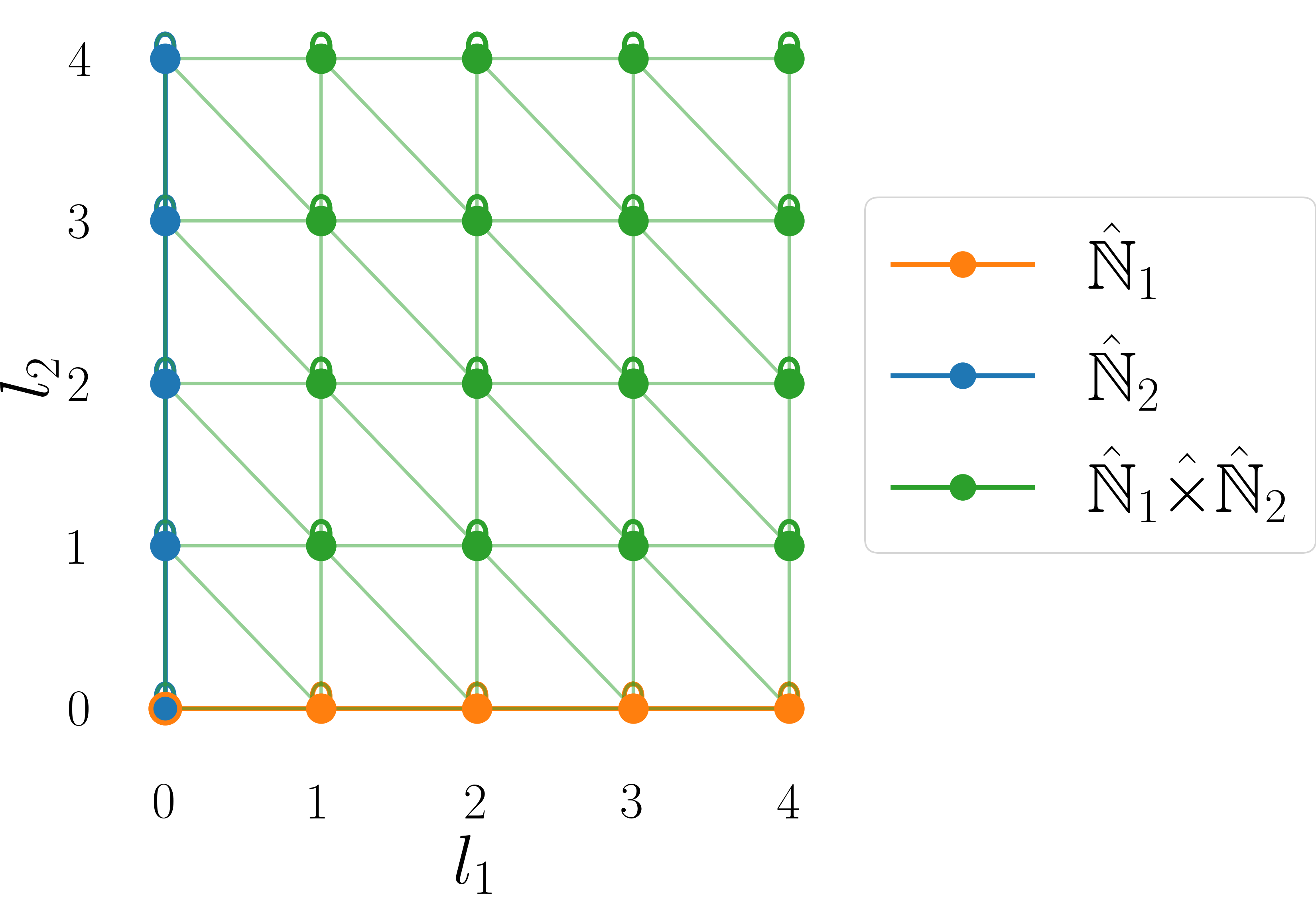}
    \caption{
    As in Figure ~\ref{fig:thick_lineage_illus}, but now for the skeletal cross product
    $\hat{\mathbb N} \hat{\times}\hat{\mathbb N}$,
    which also serves as the template for all other skeletal cross products
    by way of their level number indices.
     $\hat{\mathbb N} \hat{\Box}\hat{\mathbb N}$ is similar, but
     omits all the diagonal edges.}
    \label{fig:skel_cross_ullustration}
\end{figure}

For ${\times}$ we have a universal diagram characterization in the category of graphs and graph homomorphisms,
so we seek something similar for $\hat{\times}$
by using the ``template'' strategy from the unary
thickening operator.

We again assume our graded graphs are graph lineages
(and therefore hierarchical graph sequences),
with $\varphi({\text{roots}}) = 0$.
In the skeletal cross product of graded graphs, 
unlike for the skeletal box product,
there are edges in the full cross product of graphs
that would violate $|\Delta l \le 1|$
and therefore need to be dropped.

\subsubsection{Grading architecture}

We define the skeletal cross product $\hat{\times}$ first on $\hat{\mathbb{N}}$
and then on all graded graphs. For $\hat{\mathbb{N}}$ we define
\begin{equation}
\begin{split}
\hat{\mathbb{N}} \hat{\times} \hat{\mathbb{N}}  
&=  ( \mathbb{N} \otimes \mathbb{N}, {\cal E}_0 \cup {\cal E}_1 \cup {\cal E}_2) \quad \text{where} \\
& {\cal E}_0  = \{ \{(n,m) \} | n,m : \mathbb{N} \} \quad  \text{(singletons are vertex loop edges), and} \\
& {\cal E}_1  = \{ \{(n+1,m), (n, m+1) \} | n : \mathbb{N}  \}  
\cup  \{ \{(n,m+1), (n+1, m) \} | n : \mathbb{N}  \} 
\\ & \quad \quad \quad
    (\text{preserves } n+m \text{ among neighbors}) \\
& {\cal E}_2  = \{ \{(n,m), (n+1, m) \} | n : \mathbb{N} \} \cup \{ \{(n,m), (n, m+1) \} | n : \mathbb{N} \}  \\
& \quad \quad \quad  
(\text{increments } n+m \text{ among neighbors}) \\
& = (\hat{\mathbb{N}} {\Box} \hat{\mathbb{N}})  \cup {\cal E}_1 , \quad \text{with} \\
& \varphi_{  \hat{\mathbb{N}} \hat{\times} \hat{\mathbb{N}} }((n,m))  = n + m \\
\end{split}
\label{cal_E_defs}
\end{equation}
Here ${\cal E}_0$ and ${\cal E}_1$ preserve the level number $n+m$ and ${\cal E}_2$ increments or decrements it.
This is intended to be a model of and a means of imposing the $l=l_1 + l_2$ cost constraint 
on hierarchical graph families mentioned above, though $ \hat{\mathbb{N}}$ itself,
decomposed into core and halo and viewed as a hierarchical graph family, has base $= 1$.

For all graded graphs (including, one can prove, $\hat{\mathbb{N}}$) we will 
define the skeletal cross product $G_1 \hat{\times}  G_2$
using a universal diagram. In this diagram the morphism
$\varphi_{G_1 \hat{\times}  G_2} : (G_1 \hat{\times}  G_2) \rightarrow \hat{\mathbb{N}}$
factors as a composition of 
$h: (G_1 \hat{\times}  G_2) \rightarrow \hat{\mathbb{N}} \hat{\times} \hat{\mathbb{N}} $,
defined on vertices as $h=(\varphi_{G_1}, \varphi_{G_2})$, with 
$\varphi_{\hat{\mathbb{N}} \hat{\times} \hat{\mathbb{N}}} : (\hat{\mathbb{N}} \hat{\times} \hat{\mathbb{N}}) \rightarrow \hat{\mathbb{N}}$
defined by summation.
Then the Diagram~\ref{diagram_skel_cross_pullback}
{\it in the category of graded graphs}
commutes and is universal in the 
$(G_1 \hat{\times}  G_2)$ position.
If we incorporate extra triangles similar
to Diagram~\ref{diagram_graded_graph_hom},
we can translate to the {\it category of graphs}
as in Diagram~\ref{diag:skel_cross_commutative}.

\begin{diagfig}
\centering
\begin{tikzcd}[execute at end picture={\path (\tikzcdmatrixname-1-2) node[below right=.25 and .25]{\Huge $\lrcorner$};\path (\tikzcdmatrixname-1-2) node[below left=.25 and .25]{\Huge $\llcorner$};}]
G_1 \arrow[d, "\varphi_1" left]    & \left( G_1 \hat{\times} G_2 \right) \arrow[l, "\pi_1" above] \arrow[r,"\pi_2" above] \arrow[d, "h" left]                 & G_2 \arrow[d, "\varphi_2" left]    \\
\hat{\mathbb{N}} & \left( \hat{\mathbb{N}} \hat{\times} \hat{\mathbb{N}} \right) \arrow[l,"\hat{\pi}_1" above] \arrow[r,"\hat{\pi}_2" above] & \Tilde{\mathbb{N}}
\end{tikzcd}
\caption{The (universal) double pullback diagram that defines 
the skeletal cross product $G_1 \hat{\times} G_2$
in the category of graded graphs.
Note $h=(\varphi_{G_1}, \varphi_{G_2})$.}
\label{diagram_skel_cross_pullback}
\end{diagfig}

{\bf Definition.}
Diagram~\ref{diagram_skel_cross_pullback} is a 
known kind of gadget in category theory: 
a ``double pullback'', by analogy with ``double pushouts''.
We take this to be the diagram that defines 
$G_1 \hat{\times} G_2$
in the category of graded graphs.

Diagram \ref{diagram_skel_cross_pullback} is closely related to Diagram \ref{diagram_cat_prod} for the general categorical product
(which in the case of graph homomorphisms is the graph cross product)
in that the top line of Diagram \ref{diagram_skel_cross_pullback}
(i.e. Diagram \ref{diagram_skel_pullback_top}, now in the category of graded graphs)
is the bottom line of Diagram \ref{diagram_cat_prod},
and the double pullback universality of Diagram \ref{diagram_skel_cross_pullback} corresponds to
the rest of Diagram \ref{diagram_cat_prod}; but the middle and bottom lines of Diagram \ref{diagram_skel_cross_pullback}
are specific to the graded graph $G_1 \hat{\times} G_2$ construction.
They constrain the skeletal cross product 
to take the level numbering of $\hat{\mathbb{N}} \hat{\times} \hat{\mathbb{N}}$,
i.e. the addition of level numbers $l = l_1 + l_2$.

\begin{diagfig}
\centering
\begin{tikzcd}
G_1    & \left( G_1 \hat{\times} G_2 \right) \arrow[l,"\pi_1" above] \arrow[r,"\pi_2" above]    & G_2   \\
\end{tikzcd}
\caption{A closer look at Diagram \ref{diagram_skel_cross_pullback}, top line.
Category of graded graphs. Cf. Diagram \ref{diagram_cat_prod}, bottom line.}
\label{diagram_skel_pullback_top}
\end{diagfig}
\noindent

We can translate Diagram~\ref{diagram_skel_cross_pullback} into the category of graphs
using Diagram~\ref{diag:skel_cross_commutative},
which explicitly includes the level number map 
$\varphi_{ \left( \hat{\mathbb{N}} \hat{\times} \hat{\mathbb{N}} \right)}: (l_1, l_2) \mapsto= l_1 + l_2  $.
This map could be changed, at the risk of changing the properties 
(such as cost bounds by grade, 
near-associativity (cf. Section~\ref{section:associativity})
and up to equivalence, commutativity) 
of the resulting product.

\begin{diagfig}
\centering
\begin{tikzcd}
                 & G_1 \arrow[ld, "\varphi_1" {sloped, above}] \arrow[d, "\varphi_1" left]   & \left( G_1 \hat{\times} G_2 \right) \arrow[l, "\pi_1" above] \arrow[r, "\pi_2" above] \arrow[d, "h" left] \arrow[rdd, "\varphi_{\left( G_1 \hat{\times} G_2 \right)}" {sloped,above, near start}]                & G_2 \arrow[rd, "\varphi_2" {sloped, above}] \arrow[d, "\varphi_2" left]   &                  \\
\hat{\mathbb{N}} & \hat{\mathbb{N}} \arrow[l, "\varphi_\mathbb{\hat{N}}" {above}] & \left( \hat{\mathbb{N}} \hat{\times} \hat{\mathbb{N}} \right) \arrow[l, "\hat{\pi}_1" above] \arrow[r, "\hat{\pi}_2" {above, near end}] \arrow[rd,"\varphi_{ \left( \hat{\mathbb{N}} \hat{\times} \hat{\mathbb{N}} \right)}" {sloped,below}] & \hat{\mathbb{N}} \arrow[r, "\varphi_\mathbb{\hat{N}}" above] & \hat{\mathbb{N}} \\
                 &                            &                                                                                              & \hat{\mathbb{N}}           &                 
\end{tikzcd}
\caption{Universal commutative diagram for the skeletal cross product $\Hat{\times}$,
in the category of graphs.}
\label{diag:skel_cross_commutative}
\end{diagfig}
\noindent

In Diagram~\ref{diagram_skel_cross_pullback} all maps are interpreted as graph homomorphisms.
Note that $\varphi_{\hat{\mathbb{N}}} = \text{id}$ so the leftmost and rightmost triangles are trivial.
They do allow us however to recognize that $h$, $\varphi_1$, and $\varphi_2$ are
all graded graph homomorphisms.
Likewise the commutative triangle formed by 
$\pi_\alpha$, $\varphi_\alpha$ (vertical),
and the (unshown) composition $\hat{\pi} \circ h$
allows us to conclude that $\pi_1$ and $\pi_2$ 
are graded graph homomorphisms,
and the commutative triangle formed by 
$\hat{\pi}_\alpha$, $\varphi_{\hat{\mathbb{N}}}$ ,
and their (unshown) composition
allows us to conclude that $\hat{\pi}_1$ and $\hat{\pi}_2$ 
are graded graph homomorphisms.
The two squares remain commutative so we can create a
commutative diagram in the category of graded graphs
and their homomorphisms (Diagram \ref{diagram_skel_cross_pullback}).
We will show (Proposition 2 below) that
this diagram is also universal in the
$G_1 \hat{\times} G_2$ position,
as indicated in standard categorical diagram notation
by the right angle brackets.
%
%


What remains is to show that such a universal object exists, since if it does exist
it would be unique up to isomorphism by universality.

\vspace{10pt}

{\bf Proposition 2:} 
$G_1 \hat{\times} G_2$ as defined above always exists.

{\it Proof:}
The idea of the construction of $G_1 \hat{\times} G_2$
is to give it the same vertices as 
$G_1 \times G_2$ has
(viewing $G_1$ and $G_2$ as potentially infinite-sized but otherwise ordinary graphs),
but labeled by the summed level number 
$L=\varphi(v_1 \in G_1) + \varphi(v_2 \in G_2)$;
and for edges to include
include at each level
$L$ {\it all} of the edges 
in the ordinary non-skeletal graph cross product $G_1 \times G_2$
(again viewing $G_1$ and $G_2$ as potentially infinite-sized but otherwise ordinary graphs)
that {\it also} have level number indices $L=l_1+l_2$ or $(L \pm 1,L)$
that are consistent with the rules of a graded graph. 
This maximality will also lead to universality of the cross-product diagram,
since every vertex and every allowed edge in $f(X)$ 
finds a preimage in $G_1 \hat{\times} G_2$.

Diagram \ref{diagram_skel_cross_proof},
in which every arrow is at least a graph homomorphism
(some are also graded graph homomorphisms),
serves to bootstrap
the construction of $G_1 \hat{\times} G_2$
from $G_1 \times G_2$
via the graph homomorphism $\iota$ required by
the universal property of $G_1 \hat{\times} G_2$.
For this purpose we ignore $X$ and regard
$G_1 \hat{\times} G_2$ as the interloper
to $G_1 {\times} G_2$ in the remaining portion
of Diagram \ref{diagram_skel_cross_proof} which is equivalent to 
Diagram \ref{diagram_tensor_prod}.
Then we simply take $\iota$ to be an injective graph homomorphism,
surjective on vertices but not edges
since we omit $|\Delta l|>1$ edges,
and reintroduce the level number structure
that was suppressed to produce $G_1 {\times} G_2$.
Clearly Diagram \ref{diag:skel_cross_commutative} 
and hence Diagram~\ref{diagram_skel_cross_pullback}
without their universality property
can be satisfied this way, with $h = (\varphi_1,\varphi_2)$
and $\varphi_{\hat{\mathbb N}} = id$.

Diagram \ref{diagram_skel_cross_proof} including $X$ also serves to construct 
the mapping $g$ required to establish the universal property
of Diagram \ref{diagram_skel_cross_pullback},
given the universal property of
Diagram \ref{diagram_cat_prod}
which implies the existence of $f$ and its commuting $\pi$ triangles,
assuming the rest of Diagram \ref{diagram_skel_cross_pullback} is satisfied as we just proved.

Mapping $g$ is first defined on vertices,
using the fact that $G_1 \times G_2$ and ${G}_1 \hat{\times} {G}_2$
have the same vertices.
This diagram crosses categories:
maps $\tilde{\pi}_1, \tilde{\pi}_2, f, \iota$ 
are in the category of graphs and graph homomorphisms,
and maps 
${\pi}_1, {\pi}_2, {\pi}^\ast_1, {\pi}^\ast_2$ 
are in the category of
graded graphs and (vertex level-preserving) graded graph homomorphisms.
Map $g$ too will turn out to be a graph homomorphism 
by the universality property
of Diagram \ref{diagram_cat_prod}.
By hypothesis,
$X$ also participates as the top middle node in
a version of Diagram~\ref{diag:skel_cross_commutative} 
with $h$ replaced by $h^{\ast}$
and $\varphi_{{G}_1 \hat{\times} {G}_2}$ replaced by $\varphi_X$.
There $\varphi_X$ rules out edges in $X$
that have $|\Delta l| >1$.

By construction ${\tilde{\pi}}_i \circ \iota = \pi_i$
regardless of $X$ and its maps.
By hypothesis all $X$ edges have $|\Delta l| \le 1$.
Thus if $h^\ast$ is the graph homomorphism from $X$ to
${\hat{\mathbb N} \hat{\times} \hat{\mathbb N}}$
(in  the $X$ variant of Diagram~\ref{diagram_skel_cross_pullback}),
then $e \in E(X) \implies 
\sum_{i=1}^2 \Delta \varphi_i({\pi}^*_i(e))
= \sum_i \Delta \hat{\pi}_i(h^*(e))$
(by the $X$ variant of Diagram~\ref{diagram_skel_cross_pullback})
$= \sum_i \Delta l_i(e) = \Delta l(e) \in \{-1, 0, 1 \}$.
For the image of an edge $e$ in $X$ under $f$,
$\Delta l(f(e))= \sum_i \Delta \varphi_i(\tilde{\pi}_i(f(e)))
=\sum_i \Delta \varphi_i({\pi}^*_i(e))$
(by Diagram \ref{diagram_skel_cross_proof})
$ = \sum_i \Delta l_i(e)$
which we just shown is in $\{-1, 0, 1\}$.
Thus the $f$ images in $G_1 \times G_2$ of all edges $X$ 
all have $|\Delta l| \le 1$,
so their preimages under $\iota$ exist in 
$G_1 \hat{\times} G_2$.
Invertibility of $\iota$ on vertices, 
and the fact that it is injective on edges, 
and that all edges in $X$ find preimages under $\iota$
in $G_1 \hat{\times} G_2$,
implies the existence of a graph homomorphism
$g$ with $f = \iota \circ g$.

Next we must show that map $g$ is a
also a graded graph homomorphism,
i.e. that 
$\varphi_X = \varphi_{{G}_1 \hat{\times} {G}_2} \circ g$.
From Diagram~\ref{diag:skel_cross_commutative} in the category of graphs,
if $h^\ast$ is the graph homomorphism from $X$ to
${\hat{\mathbb N} \hat{\times} \hat{\mathbb N}}$,
it will suffice to show
$h^{\ast}  =  h \circ g$
and then apply the mapping $\varphi_{\hat{\mathbb N} \hat{\times} \hat{\mathbb N}}$
to both sides of this equation.
That's because Diagram~\ref{diag:skel_cross_commutative} tells us that 
$\varphi_{\hat{\mathbb N} \hat{\times} \hat{\mathbb N}} \circ h = \varphi_{G_1 \hat{\times} G_2}$
and correspondingly for $X$ that 
$\varphi_{\hat{\mathbb N} \hat{\times} \hat{\mathbb N}} \circ h^{\ast} = \varphi_X$.
So if $h^{\ast}  =  h \circ g$ then
$\varphi_X = \varphi_{\hat{\mathbb N} \hat{\times} \hat{\mathbb N}} \circ h^{\ast}
= \varphi_{\hat{\mathbb N} \hat{\times} \hat{\mathbb N}} \circ h \circ g
= \varphi_{G_1 \hat{\times} G_2} \circ g$, as required.

To establish
$h^{\ast}  =  h \circ g$ as graph homomorphisms, 
note that $h^{\ast}$ takes values in
the graph ${\hat{\mathbb N} \hat{\times} \hat{\mathbb N}}$
which maps injectively as a graph
into
${\hat{\mathbb N} {\times} \hat{\mathbb N}}$,
so we need only examine both projections
$\hat{\pi}_i$ of this equation:
$\hat{\pi}_i \circ h^{\ast}  =  (\hat{\pi}_i \circ h) \circ g$.
But we know from universality (Diagram \ref{diagram_skel_cross_proof})
that as graph homomorphisms
$\pi^\ast_i = \tilde{\pi}_i \circ f = \tilde{\pi} \circ \iota \circ g =
\pi_i \circ g$,
hence $\varphi_i \circ \pi^\ast_i = (\varphi_i \circ \pi_i) \circ g$,
and from the rectangles in 
Diagram~\ref{diag:skel_cross_commutative}
(first with $X$ and then with $G_1 \hat{\times} G_2$ in the top central position),
we have 
$\hat{\pi}_i \circ h^{\ast}  = \varphi_i \circ \pi^\ast_i
  = (\varphi_i \circ \pi_i) \circ g
   = \hat{\pi}_i \circ h \circ g$, as required. 
This establishes the universality property of 
Diagrams \ref{diag:skel_cross_commutative}
and hence Diagram~\ref{diagram_skel_cross_pullback},
for the $G_1 \hat{\times} G_2$ constructed from $G_1 {\times} G_2$.
This concludes the proof of Proposition 2.

\begin{diagfig}
\centering
\begin{tikzcd}[row sep = 30]
    &                                          & X \arrow[lld, "\pi^\ast_1" above] \arrow[rrd, "\pi^\ast_2" above] \arrow[lddd, "f" sloped] \arrow[rddd, dotted, "g" sloped] &                                                                       &     \\
G_1 &                                          &                                                             &                                                                       & G_2 \\
    &                                          &                                                             &                                                                       &     \\
    & G_1 \times G_2 \arrow[luu, "\tilde{\pi}_1" {sloped, above}] \arrow[rrruu,"\tilde{\pi}_2" {sloped, above, very near start}] &                                                             & {G_1} \hat{\times} {G_2} \arrow[llluu, "{\pi}_1" {sloped, above, very near start}] \arrow[ruu, "{\pi}_2" {sloped, above}] \arrow[ll, "\iota", hook'] &    
\end{tikzcd}
\caption{Transmission of universality. 
Maps $\tilde{\pi}_1, \tilde{\pi}_2, f, \iota$ 
concern $G_1 \times G_2$ and are therefore
in the category of graphs,
and maps  ${\pi}_1, {\pi}_2, {\pi}^\ast_1, {\pi}^\ast_2, g$ 
are in the category of graded graphs.
$f$ is induced by the universality of
ordinary graph cross products. 
The existence of $g$ with $f = \iota \circ g$ is ensured
by the invertibility of inclusion map
$\iota$ on vertices and its injectivity on edges.
As a graph $G_1 \hat{\times} G_2$ is only missing some
($|\Delta l|=2$) edges of $G_1 \times G_2$,
so by construction ${\tilde{\pi}}_i \circ \iota = \pi_i$
regardless of $X$ 
and its maps.}
\label{diagram_skel_cross_proof}
\end{diagfig}


For concreteness,
we now compute the edge adjacency matrices.
We follow a similar indexing 
strategy 
as for Proposition 1 (regarding the thickening operator).
Let $(l_{\alpha}, i_{\alpha, l_{\alpha}})$ index the vertices of graded
graph $G_{\alpha}$, for $\alpha \in \{1, 2\}$. 
The corresponding indexing of elements in Diagram \ref{diagram_skel_cross_pullback} becomes
Diagram~\ref{diagram_modified_skel_cross_pullback},
where as usual $L=l_1+l_2$. Likewise for the $j_{\alpha, l_\alpha}$ indices required along with $i_{\alpha, l_\alpha}$ in order to specify edges.

\begin{diagfig}
\centering
\begin{tikzcd}[row sep=40, column sep=20,execute at end picture={\path (\tikzcdmatrixname-1-2) node[below right=.25 and .25]{\Huge $\lrcorner$};\path (\tikzcdmatrixname-1-2) node[below left=.25 and .25]{\Huge $\llcorner$};}]
(l_1, i_{1, l_1}) \arrow[d, "\varphi_1" {left, near end}]    & (L,   (l_1, i_{1, l_1}), (l_2, i_{2, l_2})  )   \arrow[l, "\pi_1" above] \arrow[r,"\pi_2" above] \arrow[d, "h" {left,near end}]                 & (l_2, i_{2, l_2}) \arrow[d, "\varphi_2" {left, near end}]    \\
l_1 & (L, ( l_1 , l_2))  \arrow[l,"\Tilde{\pi}_1" above] \arrow[r,"\hat{\pi}_2" above] & l_2
\end{tikzcd}
\caption{Vertex indexing for skeletal graph cross product.}
\label{diagram_modified_skel_cross_pullback}
\end{diagfig}

With this indexing, in component notation we
argue that 
the skeletal cross product defined above has vertices
at level $L$ indexed by:
$((l_1,i_{1,l_1}),(l_2,i_{2,l_2}))$ or equivalently $(L,i_{1,l_1},i_{2,l_2})$.
following the prescription of keeping all
the allowed elements of the ordinary cross product, but
obeying the constraints $0 \leq l_\alpha \leq L$ and  $l_1+l_2 = L$,
we can 
follow the architecture of level numbers in
Equation~\ref{cal_E_defs} in regard to
set ${\cal E}_0$ and the two subsets in ${\cal E}_1$, and
write the general skeletal product adjacency matrix as:
\begin{equation}
\begin{split}
(G^{(1)}  \hat{\times} G^{(2)} & )^{(L)}_{ ( ((l_1, i_{1, l_1}), \; (l_2, i_{2, l_2})), ( (l^\prime_1, j_{1, l^\prime_1}), \, (l^\prime_2, j_{2, l^\prime_2})) ) } \\
 =& \delta_{L, l_1 + l_2 } \Big\{
 \delta_{l_1, l^\prime_1} \delta_{l_2, l^\prime_2}
G^{(1)(l_1)}_{(i_{1, l_1}, j_{1, l^\prime_1})}  G^{(2) (l_2)}_{(i_{2, l_2},j_{2,l^\prime_2})}  \\ &
%
+(\delta_{l_1+1, l^\prime_1} \delta_{l_2-1, l^\prime_2} 
    +\delta_{l_1-1, l^\prime_1} \delta_{l_2+1, l^\prime_2})
S^{(1) (l_1,l^\prime_1)}_{(i_{1, l_1}, j_{1, l^\prime_1})}
S^{(2) (l_2 ,l^\prime_2)}_{(i_{2, l_2}, j_{2, l^\prime_2})}   \Big\} .
\end{split}
\label{G_cross_components}
\end{equation}
Here we rely on the facts that 
$S^{(1) (l_1+1,l_1)}_{(i_{1, l_1 \pm 1}, j_{1, l_1})}=0$ 
unless $l_1$ is in range, and likewise for $S^{(2)}$.
This notation makes explicit that the summed level number $L = l_1+l_2 = (l_1+\Delta l)+(l_2- \Delta l)$ stays
constant under the action of $G^{(1)} \hat{\times} G^{(2)}$.
The leading delta function just serves to enforce previously stated index constraints.

For the off-diagonal blocks $S$, 
following the architecture of level numbers in
Equation~\ref{cal_E_defs} in regard to
the two subsets in ${\cal E}_2$ 
yields:
\begin{equation}
\begin{split}
S[& G^{(1)} \hat{\times}  G^{(2)} ]^{(L \pm 1,L)}_{ ( ((l_1, i_{1, l_1}), \; (l_2, i_{2, l_2})), ( (l^\prime_1, j_{1, l^\prime_1}), \, (l^\prime_2, j_{2, l^\prime_2})) ) } \\
%
%
& = \delta_{L \pm 1, l_1 + l_2 }  \delta_{L, l_1^{\prime} + l_2^{\prime} } \Big\{
S^{(1) (l_1,l_1^{\prime})}_{(i_{1, l_1}, j_{1, l_1^{\prime}})}
G^{(2)} _{l_2, (i_{2, l_2}, j_{2,l^\prime_2 })} 
+ G^{(1)}_{l_1,(i_{1, l_1}, j_{1, l^\prime_1 })} 
S^{(2) (l_2,l_2^{\prime})}_{(i_{2, l_2}, j_{2, l_2^{\prime}})}  \Big\} ,
\end{split}
\end{equation}
and all other $S[G^{(1)} \hat{\times} G^{(2)} ]^{(L,L^{\prime})}$ entries are zero.
Here we rely on the facts that 
$S^{(1) (l_1,l_1^{\prime})}_{(i_{1, l_1}, j_{1, l_1^{\prime}})} =0$ unless 
$|l_1 - l_1^{\prime}|=1$ and likewise for $S^{(2)}$;
then the formula ensures a similar condition on $S[G^{(1)} \hat{\times} G^{(2)} ]^{(L,L^{\prime})}$.
The leading delta functions just serve to enforce previously stated index constraints.

See Figure \ref{fig:paths_times_paths} for an example of a skeletal cross product between two path lineages. See also 
Section ~\ref {matrix_skeletal_cross_product} of the Supplemental Information,
for a matrix presentation of the skeletal cross-product.

\subsubsection{Properties}

Conventional graph cross and box products are both
(a) commutative up to graph isomorphism,
i.e. $G \times H \simeq H \times G$ and
$G \Box H \simeq H \Box G$
where $\simeq$ denotes graph isomorphism (by permutation of vertices),
and associative, i.e.
$(G_1 \times G_2) \times G_3 = G_1 \times (G_2 \times G_3)\equiv \times_{i=1}^3 G_i$
and
$(G_1 \Box G_2) \Box G_3 = G_1 \Box (G_2 \Box G_3) \equiv \Box_{i=1}^3 G_i$.
Their space cost grows as the product of the constituent graph space costs.
What about skeletal products?

{\it 1. Commutativity up to isomorphism}
\label{section:commutativity}

From the fact that skeletal products may be viewed as
the usual graph products with a symmetric restriction
on edges $|\Delta l_1 + \Delta l_2| \le 1$,
where $\Delta l_i = l_i - l^{\prime}_i$
for nodes connected between levels $l_i$ and $l^{\prime}_i$,
we see that both skeletal products are commutative up to isomorphism:
$G \hat{\times} H \simeq H \hat{\times} G$ and
$G \hat{\Box} H \simeq H \hat{\Box} G$.

A similar argument applies to the 
commutativity 
algebraic property of the skeletal box product defined
in Section~\ref{skeletal_box_product}, for the same reasons.

{\it 2. Associativity}
\label{section:associativity}

The skeletal graph box product is associative, as we will see.
Perhaps unfortunately, 
the skeletal graph cross product is
not quite associative.
For the skeletal product
$(G_1 \hat{\times} G_2) \hat{\times} G_3$,
we have the level number restrictions
$|\Delta l_i| \le 1$ and $|\sum_{i=1}^3 \Delta l_i| \le 1$
which are symmetric under permutations,
but also $|\Delta l_1 + \Delta l_2| \le 1$
which is not. To compare with 
$G_1 \hat{\times} (G_2 \hat{\times} G_3)$
we compute
$|\Delta l_2 + \Delta l_3| = |(\sum_{i=1}^3 \Delta l_i) - \Delta l_1|
\in | \{-1,0,1\} - \{-1,0,1\}|$
which can take the forbidden value 2 for 
$(l_1, l_2, l_3) \in \{(-1, +1, +1), (+1, -1, -1) \}$
which are two out of 19 symmetric possibilities
(hence ``almost'' but not actually associative).
So we cannot define an unparenthesised 
$G_1 \hat{\times} G_2 \hat{\times} G_3$ 
as just either of the parenthesised ones.

What we can do instead is to  
define an unparenthesised 
$G_1 \hat{\times} G_2 \hat{\times} G_3$ 
by the permutation-symmetric constraint 
$|\sum_{i=1}^3 \Delta l_i| \le 1$ alone.
In that case we have the same set of vertices,
and a superset of edges, compared to the parenthesised triple products:
\begin{equation}
(G_1 \hat{\times} G_2) \hat{\times} G_3 \; \subseteq_{\text{Edges}} \; \hat{\times}_{i=1}^3 G_i 
\quad \text{and} \quad
G_1 \hat{\times} (G_2 \hat{\times} G_3) \; \subseteq_{\text{Edges}} \; \hat{\times}_{i=1}^3 G_i 
\end{equation}

More generally we define the $n$-way skeletal cross product
for $n > 2 $
by the level number constraint
\begin{equation}
\left |\sum_{i=1}^n \Delta l_i \right| \le 1 ,
\end{equation}
resulting in commutativity up to graph isomorphism:
\begin{equation}
\hat{\times}_{i=1}^n G_i \; \simeq \; \hat{\times}_{i=1}^n G_{\pi(i)} 
\end{equation}
where $\pi$ is any permutation on $n$ elements,
and also in edge superset ``near-associativity'' for which:
\begin{equation}
(\hat{\times}_{i=1}^{k, 1 \le k < n} G_i) 
    \hat{\times} 
        (\hat{\times}_{i=k+1}^{n} G_i) \; 
    \subseteq_{\text{Edges}} \; \hat{\times}_{i=1}^n G_i ,
\end{equation}
where the $n \in \{1,2 \}$ skeletal products are just
$\hat{\times}_{i=k}^{k} G_i = G_k$
and
$\hat{\times}_{i=k}^{k+1} G_i = (G_k \hat{\times} G_{k+1})$.
The reason is that the outermost skeletal product on the left hand side
requires 
\begin{equation}
|\Delta(\sum_{i=1}^{k \le n} l_i) + \Delta(\sum_{i=k+1}^{n} l_i)| \le 1,
\end{equation}
which is the same as the $|\Delta(\sum_{i=1}^{n} l_i) )| \le 1$
constraint for the right hand side. In addition there are further
left hand side constraints arising from the inner parenthesised
skeletal products, recursively.

An alternative definition reverses
all the foregoing $\subseteq_\text{Edges}$ set inclusions by
allowing only $[\Delta l_i|i]$ vectors that
take the 0-padded alternating sign pattern 
$(0^* \; \sigma \; 0^* \; -\sigma \; \ldots \; \pm \sigma \; 0^*$),
where  $\sigma \in \{\pm 1\}$ and $0^*$ represents zero or more
consecutive occurrences of $\Delta l_i = 0$.
Such a pattern can be parenthesized by any tree 
that respects the linear order of factors $G_i$, 
and still remain valid. By this means we obtain
an alternative n-way skeletal graph cross operator $\tilde{\times}$
for which 
\begin{equation}
    \tilde{\times}_{i=1}^n G_i \; 
        \subseteq_{\text{Edges}} \; 
    (\tilde{\times}_{i=1}^{k, 1 \le k < n} G_i) 
    \hat{\times} 
        (\tilde{\times}_{i=k+1}^{n} G_i) \; 
,
\end{equation}
where again the $n \in \{1,2 \}$ lower-bounding
skeletal products are just
$\tilde{\times}_{i=k}^{k} G_i = G_k$
and
$\tilde{\times}_{i=k}^{k+1} G_i = (G_k \hat{\times} G_{k+1})$.

By contrast,
the skeletal box product $\hat{\Box}$ defined
in Section~\ref{skeletal_box_product} below is (exactly)
associative. The level number constraint for 
the $n$-way skeletal box product is
\begin{equation}
\sum_{i=1}^n |\Delta l_i | \le 1 ,
\end{equation}
from which one deduces 
$
\left |\sum_{i=1}^n \Delta l_i \right| =
\sum_{i=1}^n |\Delta l_i |
$
whence 
\begin{equation}
|\sum_{i=1}^{k \le n} \Delta l_i| + |\sum_{i=k+1}^{n} \Delta l_i| 
= 
\sum_{i=1}^{k<n} |\Delta l_i| + \sum_{i=k+1}^{n} |\Delta l_i|
= \sum_{i=1}^{n} |\Delta l_i| \, ,
\end{equation}
which implies associativity of $\hat{\Box}$.

{\it 3. Distributivity} 

Graph sums (disjoint unions) can be taken levelwise for
graph lineages, in which case the usual
distributive laws of graph products over graph sums,
given in matrix notation in Appendix S1,
apply also for skeletal products
since level numbers correspond between summands.

{\it 4. Cost} 

The complexity class $\log |G| = O(n^{1+\epsilon})$
for hierarchical graded graphs is
invariant under $n \rightarrow n \log n$ and is therefore
preserved under the binary $\hat{\times}$ operation.
Section~\ref{product_space_cost} presents a more detailed analysis.


\subsection{Skeletal box product, $\hat{\Box}$}
\label{skeletal_box_product}

The universal diagram for $\Box$ in the category of graphs 
is a bit more complex than that for $\times$, but the box product
is more relevant for process models
since it describes independent nonoverlapping graph-local processes
such as diffusion on a graph, whose continuous-time
operator is the graph Laplacian.
We expect something similar for $\hat{\Box}$ and $\hat{\times}$:
that the virtues of $\hat{\Box}$ as a cost-saving substitute for ${\Box}$
will pertain to process models, including those used to
define process distances. In principle the goal is to define $\hat{\Box}$ so as
to minimize some directed process distance
\begin{equation}
\label{Dgoal}
D^2(G^{(1)} \hat{\Box} G^{(2)}, G^{(1)} \Box G^{(2)}).
\end{equation}

With that in mind we again define the skeletal product of graphs in terms
of an architecture given by the product of two copies of $\mathbb{N}$,
but this time for box products. Therefore the $\{(n,m)\} \in {\cal E}_0$ diagonal inverse image blocks should
be of the form $G^{(1)}_n \Box G^{(2)}_m$ rather than $G^{(1)}_n \times G^{(2)}_m$, and the ${\cal E}_2$ blocks should be omitted entirely.

The inverse images of non-self connections in $\hat{\mathbb{N}} \hat{\Box} \hat{\mathbb{N}} $
can be taken to be the corresponding elements of $S$ for graphs or $P$ for weighted graphs.
There is may also be an advantage to $P$ over $G$ in minimizing (\ref{Dgoal}).
But where such factors are to be multiplied by inverse images of self connections in $\hat{\mathbb{N}}$,
those self connection factors can be either $G$ or the identity matrix $I$,
in accordance with the constraints of the box conventional box product.


{\it Terminology}: The graph architectures (at any given level in a graded graph)
 resulting from the thickening operation are ``pyramids''
 or ``scale spaces''.
The graph architectures resulting from the binary $\hat{\Box}$ operation are ``bipyramids'',
and those resulting from iterated or many-way $\hat{\Box}$ are more generally ``multipyramids''.
Of course one can also thicken pyramids and multipyramids.


\subsubsection{Diagram for $\hat{\Box}$}

We redraw the tensor product universal diagram, 
with a view to complicating it:

\begin{diagfig}
\centering
\begin{tikzcd}
E_G \times E_H \arrow[rr, "\tau" above, "\text{bimorphism}" below] \arrow[rrrd, "\xi" {above,sloped}, "\text{bimorphism}" {below, sloped}]  &  & G \Box H  \arrow[rd, dotted, "\xi^*"{sloped, above}]  &   \\
                                             &  &          &   X    \\
\end{tikzcd}
\caption{The universal diagram for tensor products.}
\label{diagram_tensor_prod2}
\end{diagfig}
\noindent
where the bimorphism maps are in the category of sets,
rather than graphs or graph lineages,
and where $\tau$ is the identity map on vertices
(hence also on pairs of vertices).
Bimorphisms are for each fixed vertex $v_1 \in G_1$,
a graph homomorphism (taking graph edges to graph edges)
on $E_2$, and vice versa.
Combining maps from different categories makes sense since
there are forgetful functors from graphs to sets and
(for later use) from graph lineages to graphs.

Since $\tau$ is the identity map on vertices,
$\tau$ has an inverse function $\tau^{-1}$,
both on vertices and on vertex pairs,
and we may define $\xi^{\ast}=\xi \circ \tau^{-1}$ on graph vertices and edges.
$\xi^{\ast}$ is a graph homomorphism by virtue of the
bimorphism property
(\cite{KnauerAlgGraphs}, Theorem 4.3.5).

We now extend this diagram to add graph lineage conditions,
by analogy with the skeletal cross product case,
as shown in Diagram~\ref{diagram_box_product_pf1}.
\begin{diagfig}
\centering
\begin{tikzcd}[execute at end picture={\path (\tikzcdmatrixname-1-3) node[below left=.25 and .25]{\Huge $\llcorner$};}]
E_G \times E_H \arrow[rr, "\tau" above, "\text{bimorphism}" below] \arrow[dd, "(\varphi_G \text{,} \varphi_H)" left] &  & G \hat{\Box} H \arrow[dd,"h"]     \\
                                             &  &                  \\
E_{\hat{\mathbb{N}}} \times E_{\hat{\mathbb{N}}} \arrow[rr, "\text{bimorphism}" below, "\tilde{\tau}" above]          &  & \hat{\mathbb{N}} \hat{\Box}  \hat{\mathbb{N}}
\end{tikzcd}
\caption{Box product definition. $h$ is a graph lineage homomorphism, 
and the diagram is universal 
as in Diagram~\ref{diagram_tensor_prod2}
at the $G \hat{\Box} H$ corner.}
\label{diagram_box_product_pf1}
\end{diagfig}
This diagram
can be 
combined with Diagram~\ref{diagram_tensor_prod2} and
unpacked into graph homomorphisms
in the more detailed Diagram~\ref{diagram_box_product_pf3}.
\begin{diagfig}
\centering
\begin{tikzcd}[row sep=30, column sep=15]
E_G \times E_H \arrow[rr, "\tau" above, "\text{bimorphism}" below] \arrow[dd, "(\varphi_G \text{,} \varphi_H)" left] \arrow[rrrd, "\xi" {above,sloped}, "\text{bimorphism}" {below, sloped}]  &  & G \hat{\Box} H \arrow[ddr, "\varphi_{G \hat{\Box} H}" {sloped,below}] \arrow[dd, "h" near end] \arrow[rd, dotted, "\xi^*"{sloped, above}]  &   \\
                                             &  &          &   X \arrow[ld, "h'"{sloped, below, near end}] \arrow[d, "\varphi_X"]     \\
E_{\hat{\mathbb{N}}} \times E_{\hat{\mathbb{N}}} \arrow[rr, "\text{bimorphism}" below, "\tilde{\tau}" above]          &  & \hat{\mathbb{N}} \hat{\Box}  \hat{\mathbb{N}} \arrow[r, "\varphi_{\hat{\mathbb{N}} \hat{\Box} \hat{\mathbb{N}}}"below] & \hat{\mathbb{N}}
\end{tikzcd}
\caption{Detailed box product definition.
Detailed version of Diagram~\ref{diagram_box_product_pf1},
expanding out the universality property with
another graph lineage $X$ and its graph lineage homomorphism $h^{\prime}$,
and all graph lineage homomorphisms 
(namely $h, h^{\prime}, \text{ and } \xi^{\ast}$) 
expanded out into
$\varphi$ graph homomorphism
commutative triangles as in  Diagram~\ref{diagram_graded_graph_hom}. 
Every graph lineage homomorphism is also a graph homomorphism
via the forgetful functor from graph lineages to graphs.}
\label{diagram_box_product_pf3}
\end{diagfig}
Here $h, h^{\prime}$, and $\xi^{\ast}$ are graph lineage homomorphisms,
the $\varphi$ maps are graph homomorphisms to $\hat{\mathbb N}$,
and $\tau, \xi, \tilde{\tau}$, and $(\varphi_G, \varphi_H)$
are functions in the category of sets
between graphs as ordered pairs of vertex-sets and edge-sets, 
with $(\varphi_G, \varphi_H)$ respecting its own
domain Cartesian product structure.

The top three nodes 
in Diagram~\ref{diagram_box_product_pf3}
are the same as in Diagram~\ref{diagram_tensor_prod2},
but now mappings to the graph ${\hat{\mathbb N}}$
are used to supply level numbers and to restrict
them to successive integers for graph-adjacent vertices.
We define 
${\hat{\mathbb N}} \hat{\Box} {\hat{\mathbb N}} = {\hat{\mathbb N}} \Box {\hat{\mathbb N}} $
with additive level numbers,
which is a graded graph because the successive integer constraint
is already met by graph $\Box$, unlike graph $\times $.
Clearly this definition is also consistent if 
${\hat{\mathbb N}} \hat{\Box} {\hat{\mathbb N}}$
is substituted into $G \hat{\Box} H$ in this diagram.

As in the case of the skeletal cross product, we need only show 
that such a universal object exists, since if it does exist
it would be unique up to isomorphism by universality.

{\bf Proposition 3:} 
$G_1 \hat{\Box} G_2$ as defined above exists.

{\it Proof:}
To construct $G_1 \hat{\Box} G_2$, we deduce its vertices 
from the identity mapping $\tau$
to be simply the set Cartesian product of the vertices
of $G$ and $H$.
We compute the edges of $G \hat{\Box} H$ as before
by setting $X = G \Box H $,
treated as a box product of graphs except that
the graph homomorphisms $h^{\prime}$ and $\varphi_X$
are defined using level numbers $(l_G,l_H)$ and $l_G+l_H$ respectively.
$\varphi_X$ thus defined already obeys the restriction
that nodes connected in the graph box product differ in
level number by zero or one, because one of the two 
adjacency matrix factors involved in defining an edge
is always the identity matrix which has $\Delta_l=0$.

In this special case of $X$, by Diagram~\ref{diagram_tensor_prod2}
there must be a graph homomorphism
$\tau^{\ast}: G \Box H \rightarrow G \hat{\Box} H$,
with $\tau^{\ast} = \tau \circ \xi^{-1}$ on vertices.
Since the vertices correspond 1-1 (i.e. injectively) via $\tau^{\ast}$, 
the edges must correspond 1-1 via $\tau^{\ast}$ as well.
We take $G \hat{\Box} H$ to the the image of this map,
so $\tau^{\ast}$ is also onto (i.e. surjective).
Thus $\tau^{\ast}$ is invertible and we find
$\xi^{\ast} = (\tau^{\ast})^{-1}$,
$h = h^{\prime} \circ \xi^{\ast}$, and
$\varphi_{G \hat{\Box} H} = \varphi_X \circ\xi^{\ast}$.

Therefore if we uniquely index the vertices of $G \Box H$ e.g. as 
$$(L,i_L) = (m+n, \text{en}((m,j_{m}),(n,k_{n}))$$
where, for each $L=m+n$ and over the 
ranges $m$ and $n$ of $j_{m}$ and $k_{n}$ in $G$ and $H$,
``en'' is
a 1-1 encoding function onto an initial segment of the integers,
then we can index the vertices of $G \hat{\Box} H$ the same way
and the two adjacency matrices will be the same.
We will calculate the adjacency matrices below.
We have thus constructed a graph lineage $G \hat{\Box} H$.
It satisfies the main rectangle of 
Diagram~\ref{diagram_box_product_pf3}, 
excluding the generic $X$ and all its arrows
which are needed to establish universality,
because the main rectangle 
$\tau, h, (\varphi_G, \varphi_H), \tilde{\tau}$
simply enforces the level number rules that
$G \Box H$ with $(\varphi_G, \varphi_H)$ already obeys.

For universality, we again define  $\xi^{\ast}=\xi \circ \tau^{-1}$
and verify the graph and graph lineage homomorphism properties,
from the bimorphism, graph, and graph lineage homomorphism properties
of the other arrows.
Considering $G \hat{\Box} H$ as a graph rather than
a graded graph, it is isomorphic to $G {\Box} H$.
So, $\xi^{\ast}$ is a graph homomorphism by 
Diagram~\ref{diagram_tensor_prod2}
as it is contained in Diagram~\ref{diagram_box_product_pf3}.
To establish that it also a graded graph homomorphism
requires the commutative triangle
$\varphi_{G \hat{\Box} H}= \varphi_{X} \circ \xi^{\ast}$,
which would follow if we knew the commutative triangle
$h = h^{\prime} \circ \xi^{\ast}$ since then 
$\varphi_{G \hat{\Box} H} 
= \varphi_{\hat{\mathbb N} \hat{\Box} \hat{\mathbb N}} \circ h
= (\varphi_{\hat{\mathbb N} \hat{\Box} \hat{\mathbb N}} \circ h^{\prime}) 
    \circ \xi^{\ast}
=\varphi_{X} \circ \xi^{\ast}$ as required.
What we know though is that
$h \circ \tau = h^{\prime} \circ \xi$
are equal bimorphisms and 
$\tau$ is invertible, i.e.
$h = (h \circ \tau) \circ \tau^{-1}
= h^{\prime} \circ (\xi \circ \tau^{-1})
= h^{\prime} \circ \xi^{\ast}$, QED.

\vspace{10pt}

Another way to express of this outcome is that $G^{(1)} \hat{\Box} G^{(2)}$
is just $G^{(1)} \Box G^{(2)}$ as a graph, graded by $\varphi((x,y)) = \varphi_1(x) + \varphi_2(y)$.
Since edges in $G^{(1)} \Box G^{(2)}$ change only one of $x,y$ at a time,
the graded graph constraint $\Delta l \in \{0,\pm 1\}$ is preserved
under this combined operation.

We now calculate the adjacency matrix and hence the edges explicitly.
From Equation~\ref{box_prod_coordinates}
the adjacency matrix expression for graphs $G$ and $H$
with adjacency matrices $G_{i j}$ and $H_{k l}$
is $G_{i j} I_{k l} + I_{i j} H_{k l}$ where $I$ is the identity matrix.
Changing to the level number indexing
$G_{ij} \rightarrow G^{(1)(l_1)}_{(i_{1, l_1}, j_{1, l_1})}$
and $G_{k l} \rightarrow G^{(2)(l_2)}_{(i_{2, l_2},j_{2,l_2})}$,
with $l_1, l_2 \ge 0$,
the skeletal box product defined above has vertices
at level $L$ indexed by:
$((l_1,i_{1,l_1}),(l_2,i_{2,l_2}))$ or equivalently $(L,i_{1,l_1},i_{2,l_2})$,
obeying the constraints $0 \leq l_\alpha \leq L$ and  $l_1+l_2 = L$.
Its intra-level block $L$ edges are given by:
\begin{equation}
\begin{split}
(G^{(1)} \hat{\Box} G^{(2)} &)^{(L)}_{ ( ((l_1, i_{1, l_1}),(l_2,i_{2, l_2})), ((l_1, j_{1, l_1}), (l_2,j_{2, l_2}))  } \\
 =&\delta_{L, l_1 + l_2 } \Big\{%
G^{(1)(l_1)}_{(i_{1, l_1}, j_{1, l_1})}  I^{(2)}_{l_2,(i_{2, l_2},j_{2,l_2})} 
+ I^{(1)}_{l_1,(i_{1, l_1}, j_{1, l_1})}  G^{(2)(l_2)}_{(i_{2, l_2},j_{2,l_2})} 
\Big\}
\end{split}
\label{G_box_components}
\end{equation}
The leading delta function just serves to enforce previously stated index constraints.
This equation has a simpler form than Equation~\ref{G_cross_components} 
because the lack of $S$ terms
simplifies the $l_i$ indexing.

For the off-diagonal blocks we need to add in the connections due to
prolongation/restriction maps $P$ or at least their 0/1 sparsity
structures $S$ with the same indices as $P$:
\begin{equation}
\begin{split}
S[ &G^{(1)}  \hat{\Box} G^{(2)}  ]^{(L+1,L)}_{ ( ((l_1, i_{1, l_1}), \; (l_2, i_{2, l_2})), ( (l^\prime_1, j_{1, l^\prime_1}), \, (l^\prime_2, j_{2, l^\prime_2})) ) } \\
%
&= \delta_{L+1, l_1 + l_2 }  \delta_{L, l_1^{\prime} + l_2^{\prime} } \Big\{
S^{(1) (l_1,l_1^{\prime})}_{(i_{1, l_1}, j_{1, l_1^{\prime}})}
I^{(2)} _{l_2, (i_{2, l_2}, j_{2,l_2 })} 
+ I^{(1)}_{l_1,(i_{1, l_1}, j_{1, l_1 })} 
S^{(2) (l_2,l_2^{\prime})}_{(i_{2, l_2}, j_{2, l_2^{\prime}})}  \Big\}
\end{split}
\end{equation}
and all other $S[G^{(1)} \hat{\times} G^{(2)} ]^{(L,L^{\prime})}$ entries are zero.
Here we rely on the facts that 
$S^{(1) (l_1,l_1^{\prime})}_{(i_{1, l_1}, j_{1, l_1^{\prime}})} =0$ unless $l_1 = l_1^{\prime} \pm 1$
and $S^{(2) (l_2,l_2^{\prime})}_{(i_{2, l_2}, j_{2, l_2^{\prime}})} =0$ unless $l_2 = l_2^{\prime} \pm 1$,
and the formula ensures a similar condition on $S[G^{(1)} \hat{\Box} G^{(2)} ]^{(L,L^{\prime})}$.
The leading delta functions just serve to enforce previously stated index constraints.

\begin{figure}
    \hfill
    \begin{minipage}{.31\linewidth}
        \centering
        \includegraphics[width=\linewidth]{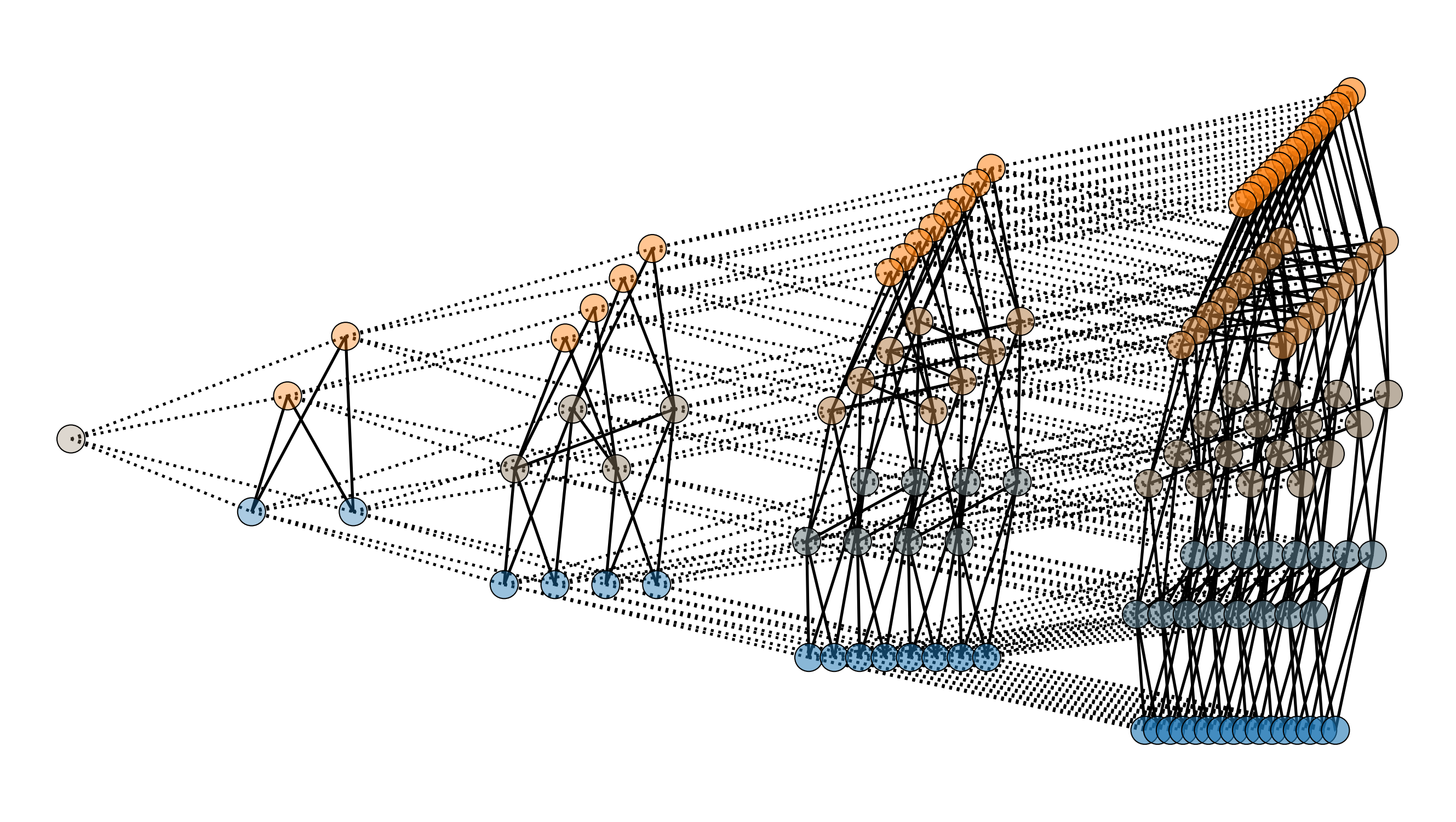}
        \caption{Paths $\hat{\times}$ Paths}
        \label{fig:paths_times_paths}
    \end{minipage} \hfill
    \begin{minipage}{.31\linewidth}
        \centering
        \includegraphics[width=\linewidth]{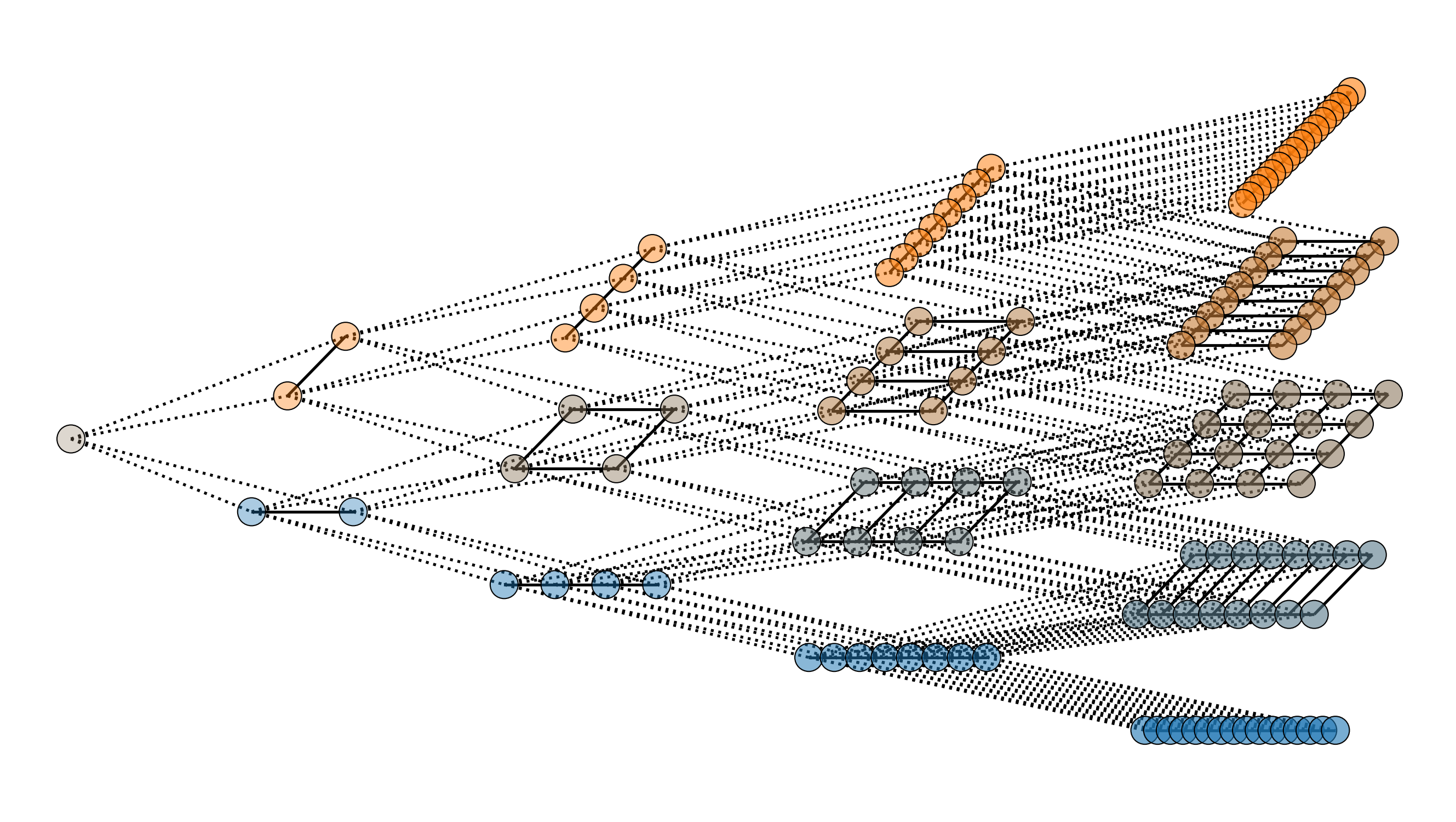}
        \caption{Paths $\hat{\Box}$ Paths}
        \label{fig:paths_box_paths}
    \end{minipage} \hfill 
    \begin{minipage}{.31\linewidth}
        \centering
        \includegraphics[width=\linewidth]{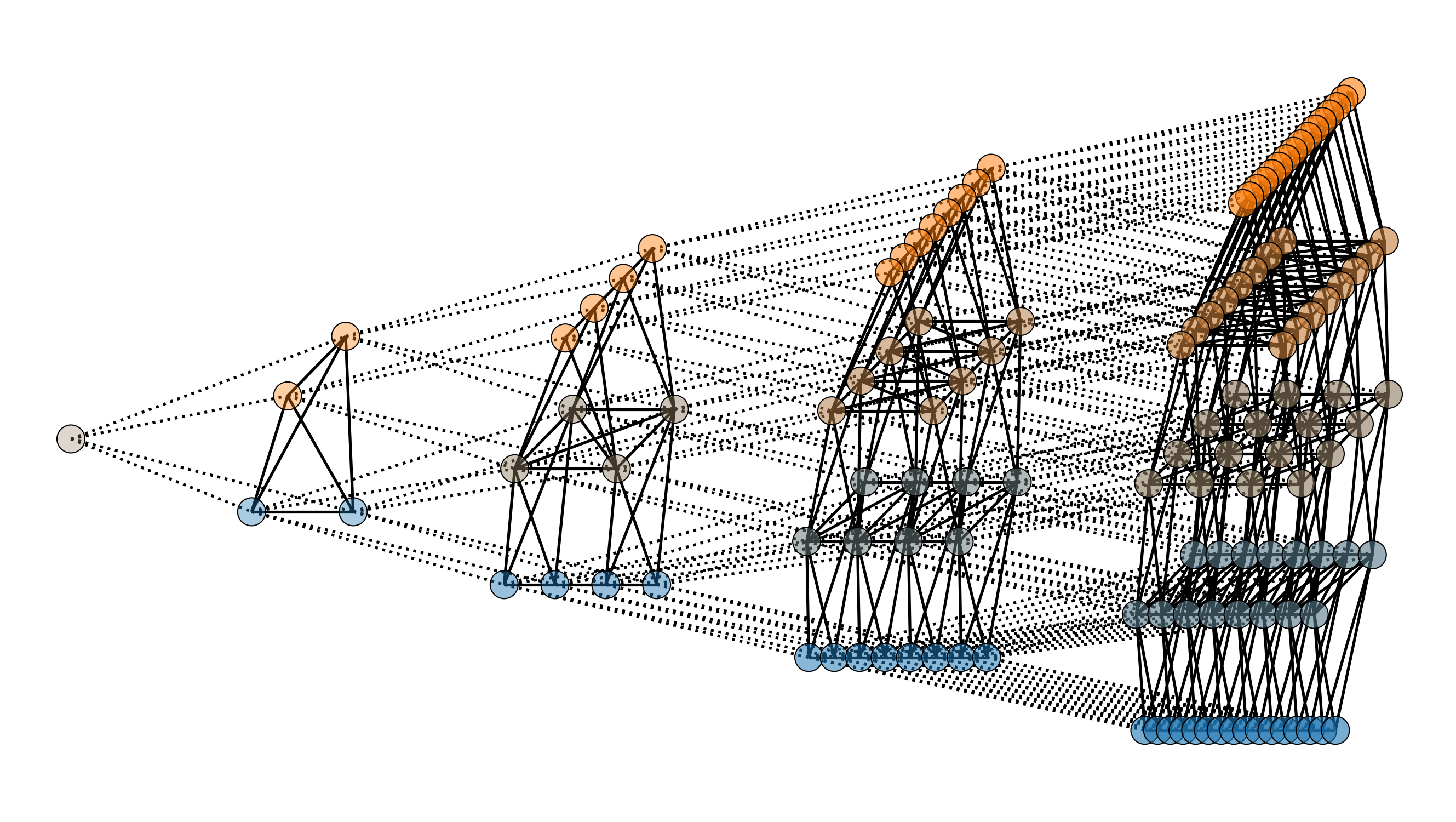}
        \caption{Paths $\hat{\boxtimes}$ Paths}
        \label{fig:paths_strong_paths}
    \end{minipage} \hfill 
    \caption{Three kinds of graph products. We illustrate the product of two path graph lineages for each of the graph products defined in this paper: the skeletal cross product $\hat{\times}$, the skeletal box product $\hat{\Box}$, and the skeletal box-cross or ``strong" product $\hat{\boxtimes}$.}
\end{figure}

See Figure \ref{fig:paths_box_paths} for an example of a skeletal cross product between two path lineages. See 
Section ~\ref{matrix_skeletal_box_product} of the Supplemental Information,
for a matrix presentation of the skeletal box product. Figure \ref{fig:paths_strong_paths}  shows a diagram of the union of the skeletal box and cross products i.e. the skeletal \emph{strong} product.

Within each grade, the graded graph 
$G_1  \hat{\Box} G_2$ we have defined is actually disconnected.
It could be reconnected within grades by considering related graph lineages
that incorporate the connecting prolongation structure of $G_1  \hat{\Box} G_2$,
such as $\theta(G_1  \hat{\Box} G_2)$, or the two-hop graph
$(G_1  \hat{\Box} G_2)^2$ truncated to include only the edges for which $\Delta l \in \{0,\pm 1\}$,
and thereby to exclude $\Delta l \in \{0,\pm 2\}$,
or by using the skeletal box-cross product defined by the union of the
edges in the skeletal box and cross products.

\subsection{Space cost of skeletal graph products}
\label{product_space_cost}

We now address space cost complexity,
the major advantage of skeletal graph products over
ordinary graph products.
We wish to show that the $O(b^{l^{1+\epsilon}})$ growth
of number of vertices is preserved by these products.
The cardinality of the number of vertices
in graded graph $\varphi_{G_1 \hat{\Box} G_2}$ grows
exponentially in level number, assuming the same is true
for $G_1$ and $G_2$, with base generally the larger
of their two bases. In particular, because $\tau$ is the 
identity map on vertices, for all level numbers $l \in {\mathbb{N}}$
we have
\begin{equation}
\begin{split}
|\varphi_{G_1  \hat{\Box} G_2}^{-1}(l)| &= \sum_{m=0}^l |\varphi_{G_1}^{-1}(m)| |\varphi_{G_2}^{-1}(l-m)| \\
&=  \sum_{m=0}^l O( b_1^{m^{(1+\epsilon)}}) O( b_2^{(l-m)^{(1+\epsilon)}}) 
=  O( (l+1)  \max(b_1,b_2)^{l^{(1+\epsilon)}}) \\
& = O( \max(b_1,b_2)^{l^{(1+\epsilon)}}) \\
\end{split}
\end{equation}
since the factor of $l+1 = b^{\log_b (l+1)} $ can be absorbed into the $\epsilon$.
Thus, the number of vertices in the resulting graded graph $G_1  \hat{\Box} G_2$
grows as required for a graph lineage.

By a similar calculation, the number of edges in the graded graph 
$G_1  \hat{\Box} G_2$
also grows as required for a graph lineage, 
assuming the same for its factors;
and the same is true for the edges of a skeletal cross product
$G_1  \hat{\times} G_2$.
Consequently:

{\bf Proposition 4} The graded graph $G_1  \hat{\Box} G_2$
is also a graph lineage, if its factors $G_1$ and $ G_2$ are.
The base $b$ of exponential cardinality growth with level number
is $\max(b_1,b_2)$.
The same is true for skeletal cross products.

In particular $\max(b_1,b_2)$ is much better than 
the product $b_1 b_2$ of a naive level-by-level
graph product of graph lineages, particular
for products of many lineages.
So for the repeated skeletal product
$\hat{\times}_{i=1}^n G_i$,
cardinalities grow exponentially 
as $O(b^{l^{1+\epsilon}})$ with base
$b = \max_{i=1}^n b_i \ll \prod_{i=1}^n b_i$,
and likewise for $\hat{\Box}_{i=1}^n G_i$.
This cost control
suggests that even a cost-affordable skeletal version
of a graph-to-graph function space is possible,
which we will argue in more detail in 
Section~\ref{section:skeletal_function_space}.

The choice of $O(b^{l^{1+\epsilon}})$ as the
definitional growth rate upper bound for graph lineages
is somewhat arbitrary, analogous to the choice
of logarithmic graph paper for plotting growth.
However it provides the correct scaling for
the skeletal products, so that the overhead
is just logarithmic, and for thickening, for which
the overhead is just a constant factor,
both of which essentially disappear into
the $O(b^{l^{1+\epsilon}})$ growth curve.
Slower-growing graded graphs such as those
resulting from rule firings in a local graph grammar
could be subsampled by omitting most levels,
forming a graph lineage with the standard growth curve.

\subsection{``Shaped'' and ``dilated" skeletal products}

\begin{figure}
    \centering
    \includegraphics[width=0.5\linewidth]{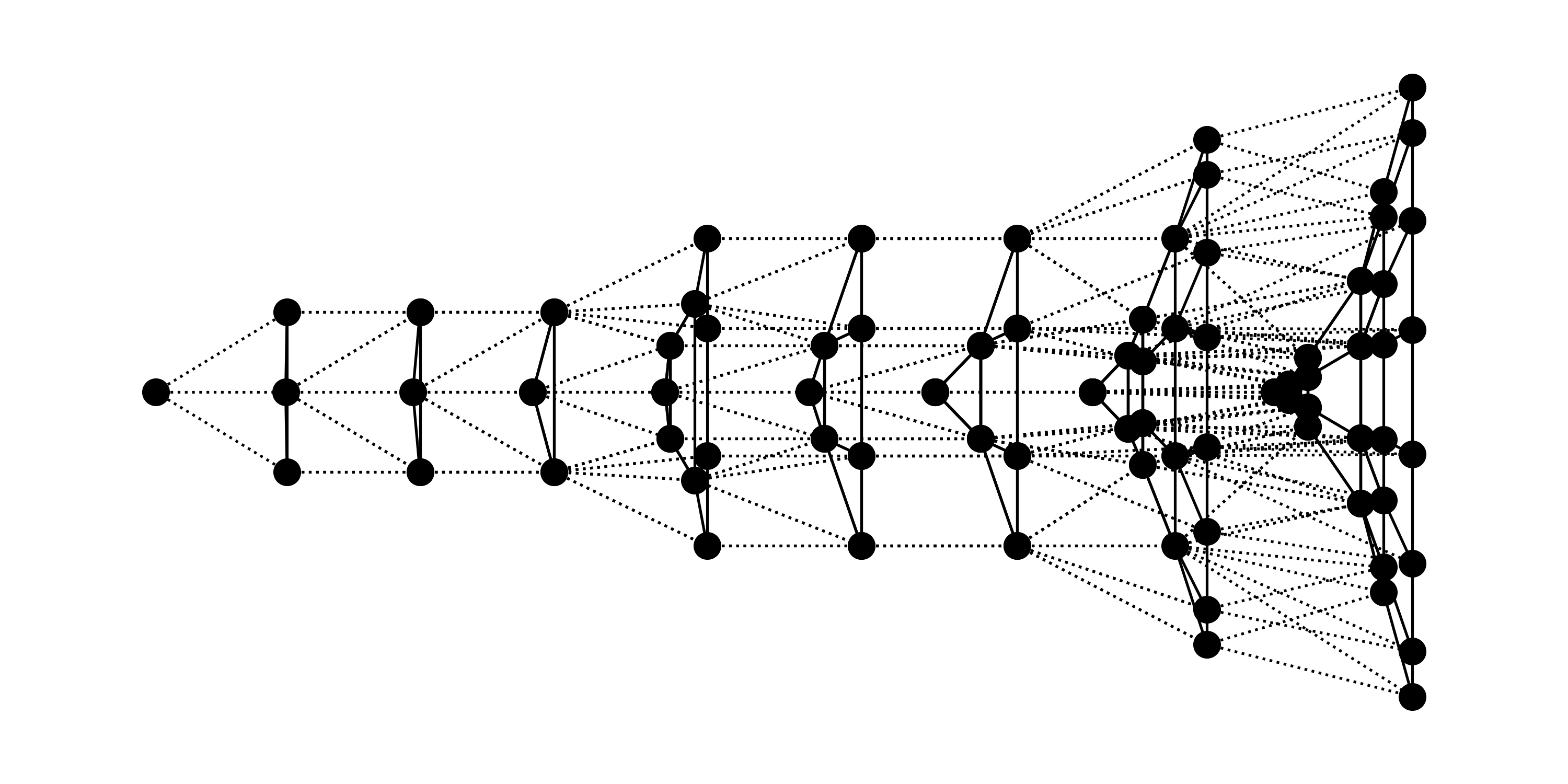}
    \caption{A dilated skeletal product between two path graph lineages.}
    \label{fig:skel_dilated_fft}
\end{figure}

For any of the preceding definitions, we can generalize the requirement that each level $L$ of a skeletal product consists of combinations of specific levels from the original two lineages, i.e. $l_1 + l_2 = L$. For example we can instead require that $f_1(l_1) + f_2(l_2) = L$ for any two monotonic functions $f_1, f_2$, or even $f_1(l_1) + f_2(l_2) = f(L)$. $f_i$ determine the ``shape'' of the network.
As a special case, we could have the constraint that $\lceil \rho_1 l_1 \rceil + \lceil \rho_2 l_2 \rceil = L$ for $\rho_i>0$. This version of the constraint allows for an ``exchange rate" of ${\rho_2}/{\rho_1}$ between changes of level number in two graph lineage factors in a skeletal product, and also a repetitiveness of $1/\max(\rho_1,\rho_2)$ in the number of levels or grades having the same given set of dimensions, as is useful in deep neural network architectures. The resulting network is ``dilated'' along the $L$ axis, as in Figure~\ref{fig:skel_dilated_fft}.
The $\lceil \ldots \rceil$ ceiling functions ensure that $L$ can be used to upper-bound the space cost of each level. However in this scheme the skeletal product may no longer be near-associative, nor commutative unless $\rho_1 = \rho_2$.

The algebraic form $l_1^p + l_2^p = L^p$ for $p>0$ is worth a look, since for $p=1$ it is the skeletal product, and for $p \rightarrow \infty$ it is $\max(l_1,l_2) = L$ i.e. the ordinary graph product (box or cross), and as $p\rightarrow 0^+$ one approaches a discrete graph sum. For $p=2$ one is hunting for Pythagorean triples, which are sparse, and for larger integer $p$ there aren't any solutions due to Fermat's Last Theorem. However one could accept instead solutions of $\lceil (l_1^p + l_2^p)^{1/p} \rceil = L$, or more generally $\lceil ((\rho l_1)^p + (\rho l_2)^p)^{1/p} \rceil = L$, and thereby successfully interpolate between ordinary and skeletal graph products. Or one can venture beyond, using $p<1$.


\section{Applications}
%
%
In this section, we discuss two applications of skeletal products to numerical algorithms. In Sections \ref{subsec:cnn_model_desc} and \ref{subsec:cnn_experiment}, we build a skeletal Convolutional Neural Network and evaluate its performance. In Section \ref{subsec:skel_multigrid} we explore the use of skeletal products in multigrid algorithms. Code for the CNN experiments is available at \url{https://github.com/cory-b-scott/SkelCNN}. The multigrid experiment code is available at \url{https://github.com/cory-b-scott/SkelMG}. Finally, a separate repository at \url{https://github.com/cory-b-scott/SkelProd} contains standalone code for generating graph lineages (represented as sparse tensors in PyTorch), as well as thickened lineages, skeletal box products, and skeletal cross products. 

\subsection{Building a CNN using skeletal products}
\label{subsec:cnn_model_desc}

\begin{figure}
    \centering
    \includegraphics[width=0.37\linewidth]{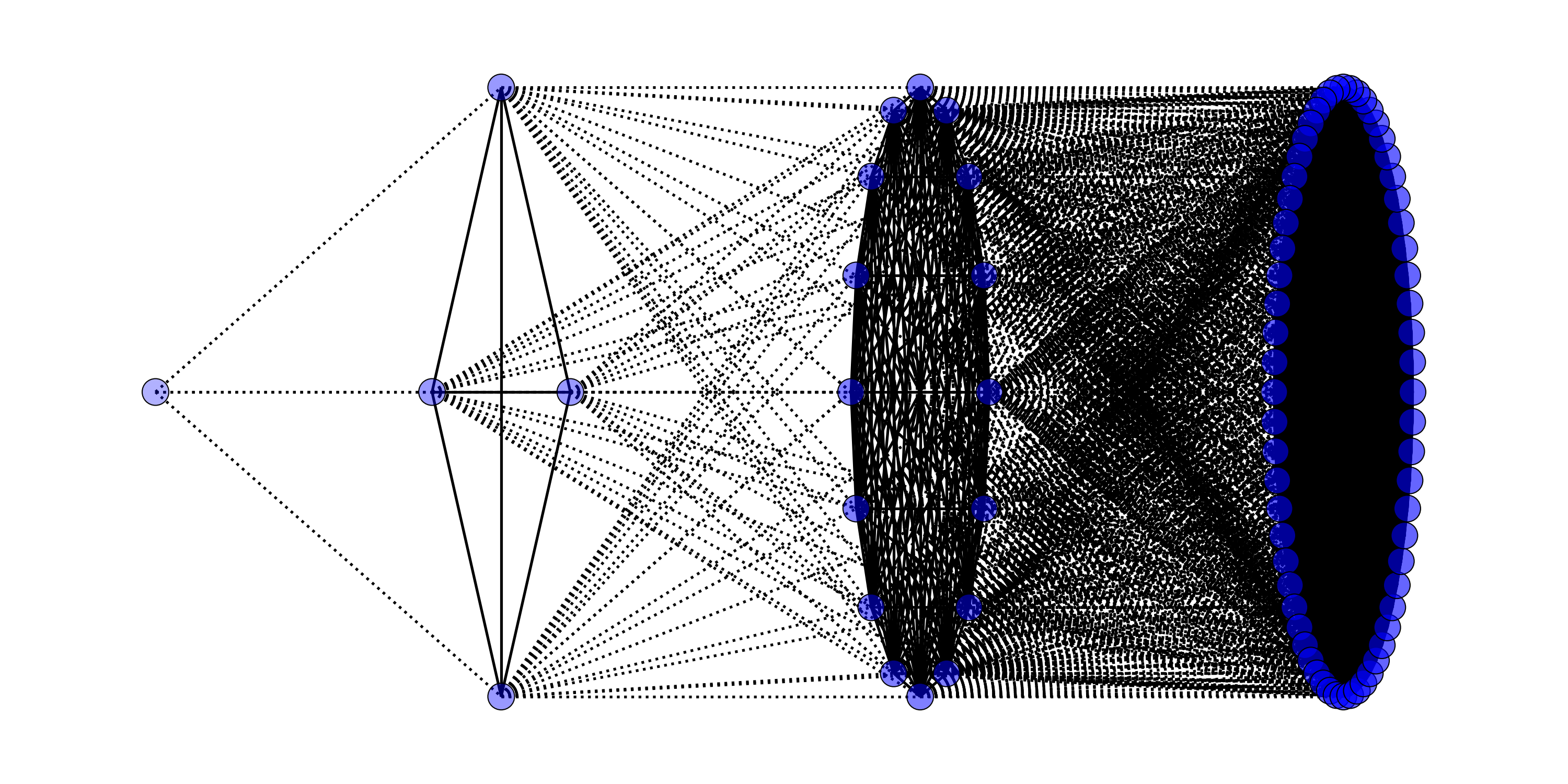} \hfill \includegraphics[width=0.57\linewidth]{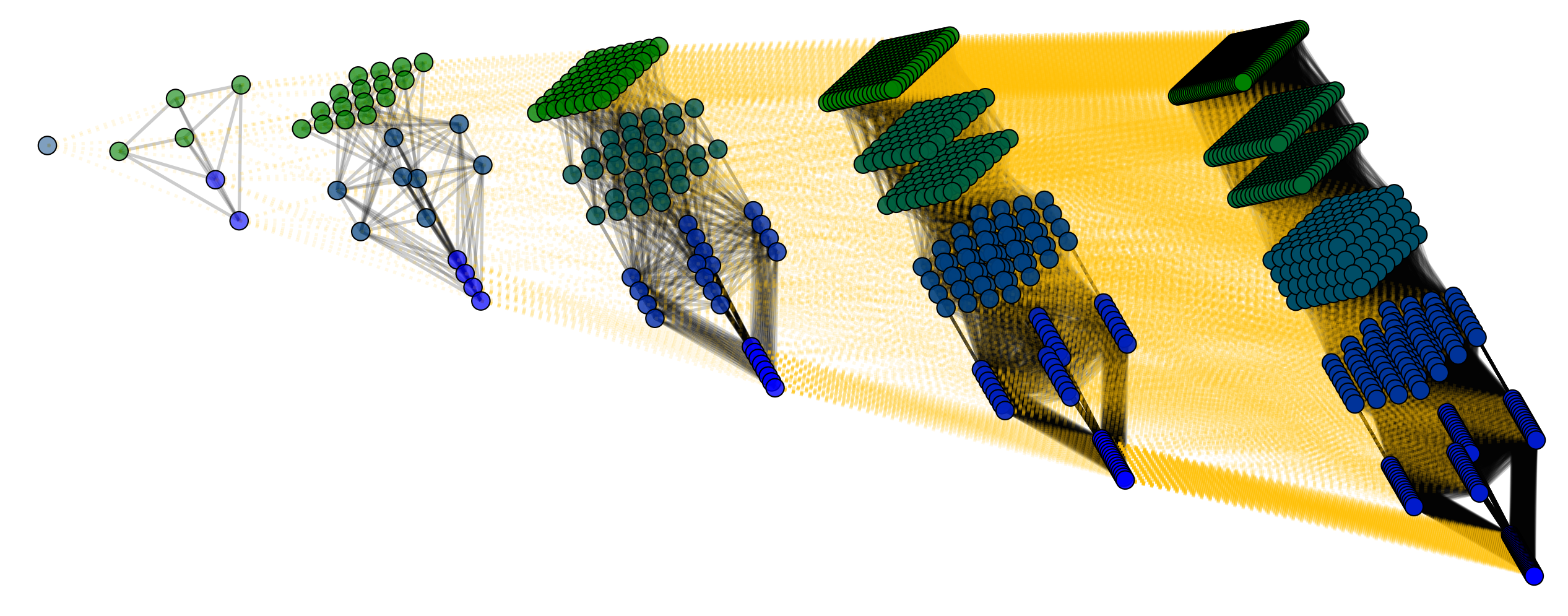}
    \caption{Left: a complete graph lineage, representing feature maps in a convolutional neural network. Right: the full convolutional neural network produced by taking the skeletal box product of the complete graph feature lineage with the grid graph lineage in Figure \ref{fig:glin_grids}. We see this produces a lineage of CNN-like structures. }
    \label{fig:conv_and_feat_lin}
\end{figure}

\begin{figure}
    \centering
    \includegraphics[width=\linewidth]{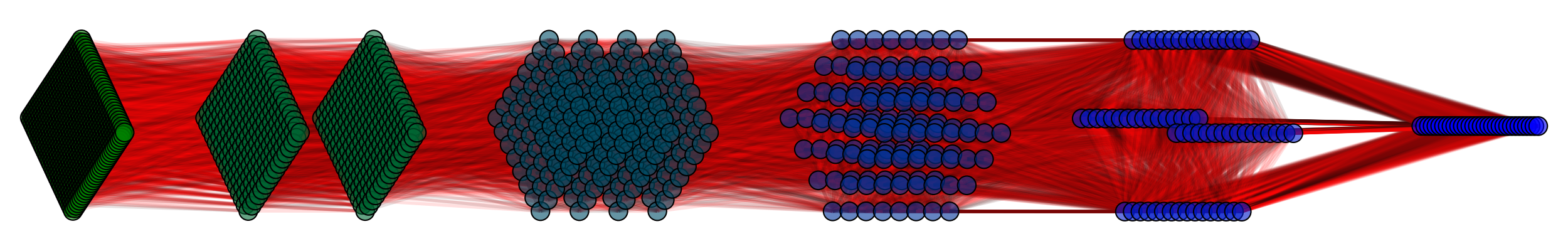}
    \caption{ The final graph in the convolutional neural net lineage from Figure \ref{fig:conv_and_feat_lin}. Note the structural similarity to classical depictions of convolutional neural nets e.g. Figure 2 of Fukushima et al. \cite{fukushima1988neocognitron}. }
    \label{fig:conv_model}
\end{figure}

In this subsection we demonstrate a practical application of the skeletal product of graph lineages. We construct a skeletal product that has a similar structure to a widely used machine learning model for image processing, a Convolutional Neural Network (CNN). As in the original construction of Convolutional Neural Networks, this architecture is inspired by the original \emph{Neocognitron} of Fukushima et al. \cite{fukushima1988neocognitron}. CNNs have enjoyed wide application in a variety of computer vision tasks.  

We construct a skeletal product CNN as a skeletal box-cross product of two lineages: 1) a spatial pyramid lineage, represented as a sequence of growing 2D grid graphs, and 2) a hierarchical lineage of composed feature maps, represented as a lineage of complete graphs. See Figure \ref{fig:conv_and_feat_lin}. By taking a skeletal product between a lineage of growing grid graphs (representing a spatial pyramid of images) and a lineage of growing complete graphs, the resulting model is able to both pool information across pixels and establish part-whole relationships between image features. 
%
%
%
%
\subsection{Comparison between a skeletal CNN and CNN}
\label{subsec:cnn_experiment}
\begin{figure}
    \centering
    \includegraphics[width=.47\linewidth]{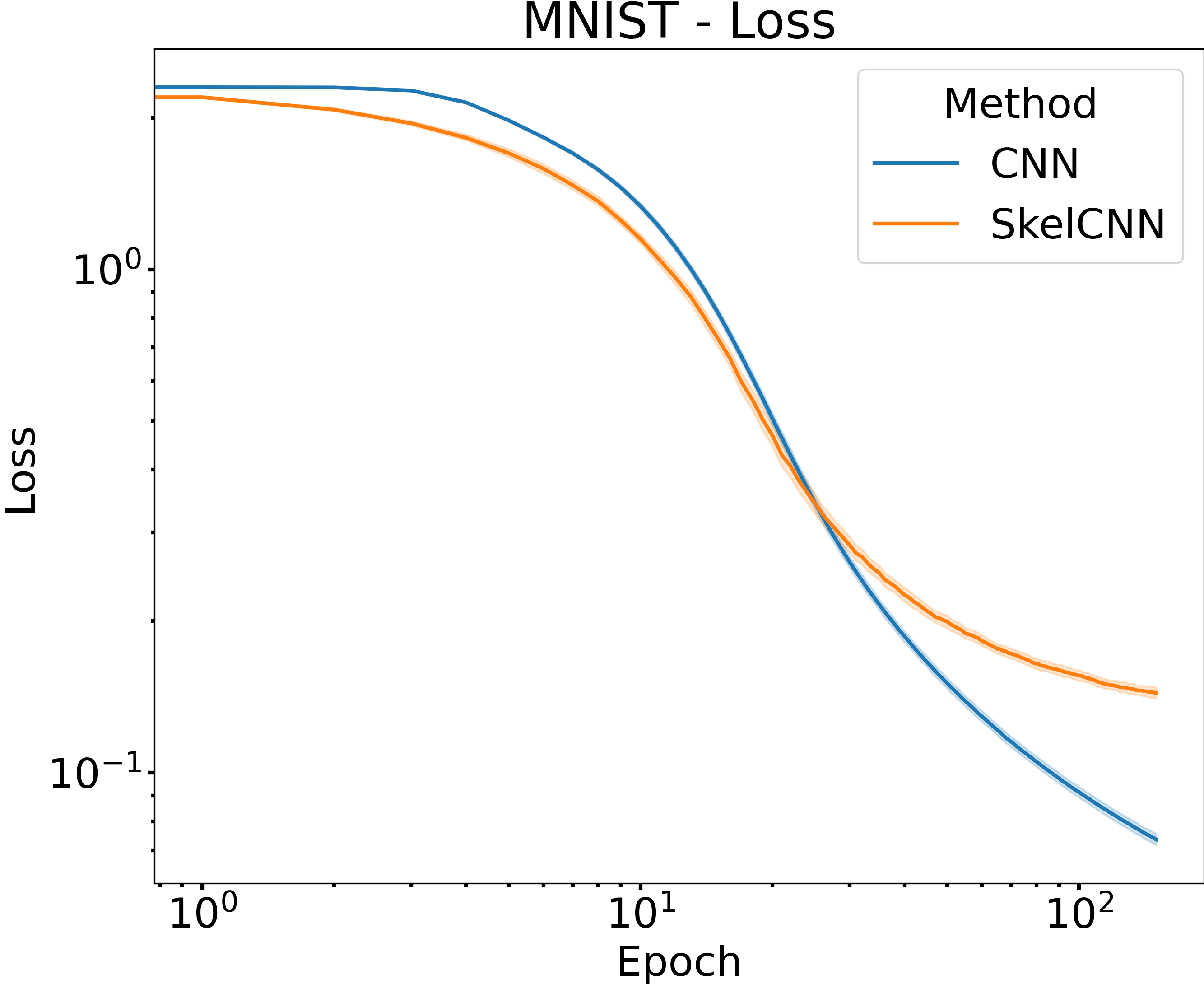} \hfill \includegraphics[width=.47\linewidth]{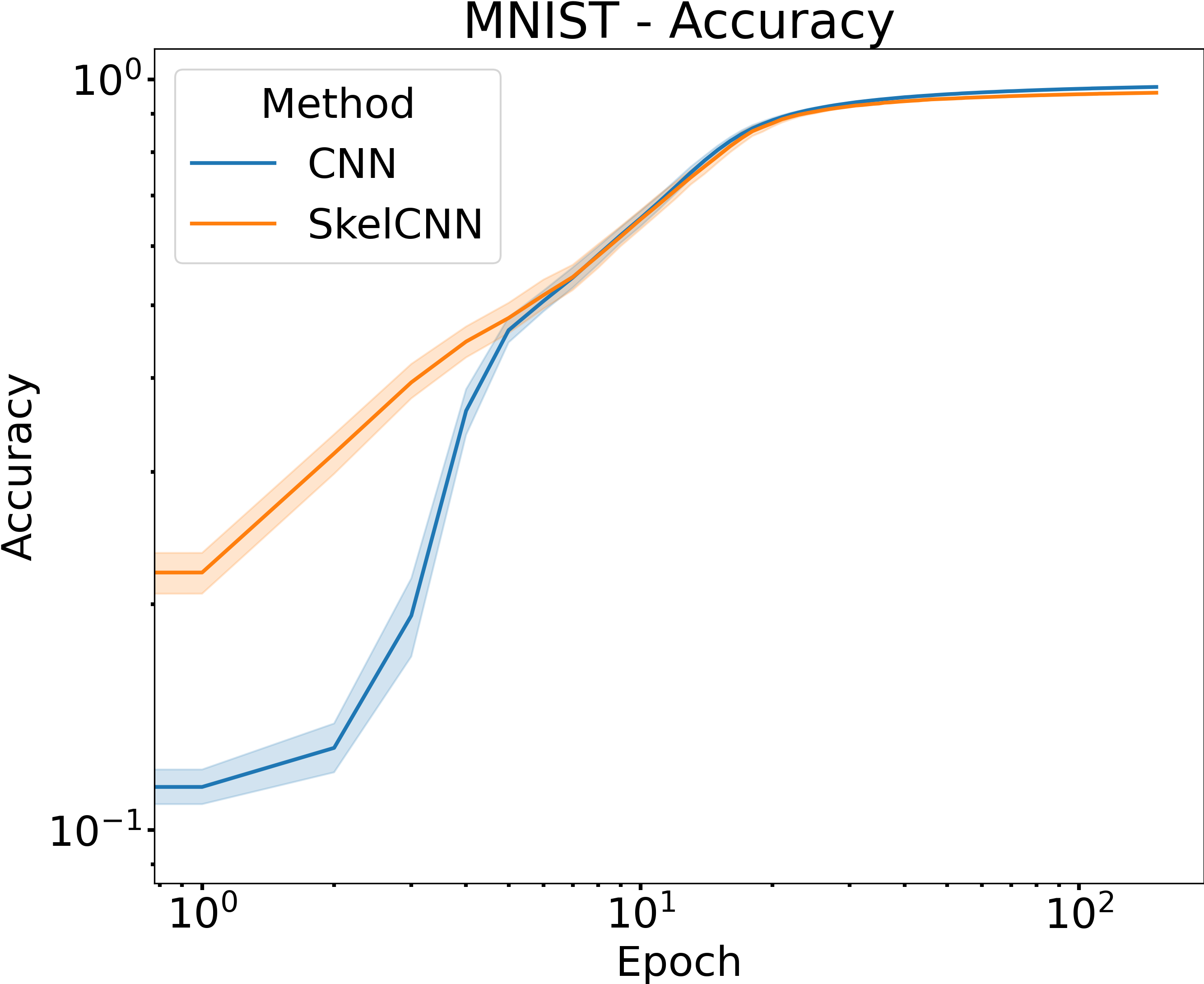} \\
    \includegraphics[width=.47\linewidth]{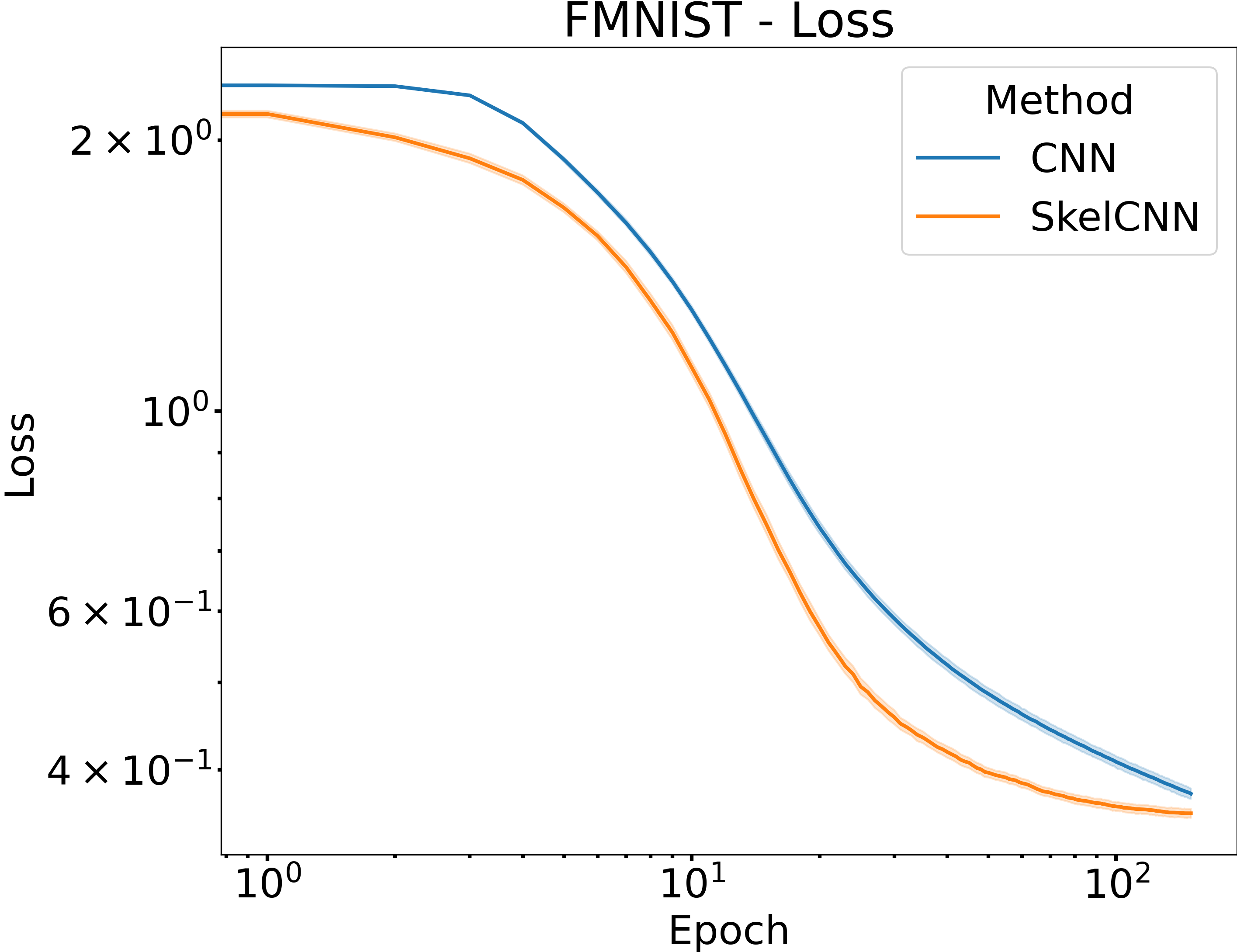} \hfill \includegraphics[width=.47\linewidth]{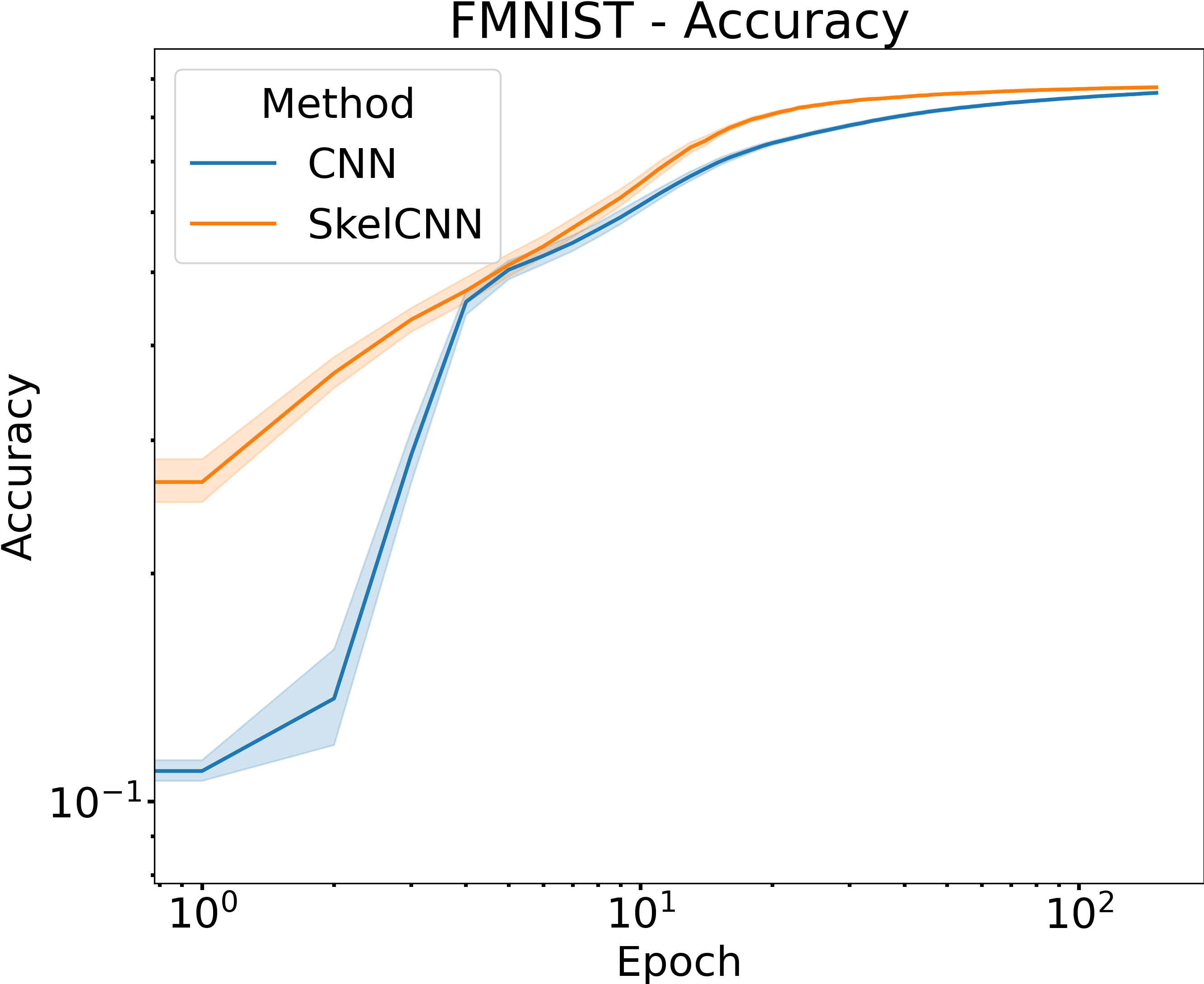} \\
    \caption{Comparison between a skeleton convolutional neural network, and a regular CNN with identical image sizes and number of filters. Each curve shows the average validation loss (as a function of epochs of training) over 32 random restarts with different random seeds. Shaded sections indicate 95\% confidence intervals. We see that the skeletal neural network learns solutions which are equivalent to those learned by a convolutional neural network. The Skeletal CNN has slightly different learning characteristics, as evidenced by the difference in loss and accuracy curves in each subplot.  }
    \label{fig:skel_vs_cnn_loss}
\end{figure}

\begin{table}
    \centering
    \caption{Results from our experiment comparing a typical CNN to one constructed using skeletal graph products. We see that the skeletal product achieves comparable performance (in terms of computational cost to train and accuracy) to the built-in pytorch CNN layers.}
    \begin{tabular}{c|c|c|c}
        Dataset & Model & Loss & Accuracy \\ \hline
        \multirow{2}{*}{MNIST} & CNN & $ 0.065 \pm 0.005 $ & $ 97.9 \pm .1  $ \\ \cline{2-4}
         & Skeletal CNN & $ 0.14 \pm 0.01 $ & $ 95.9 \pm 0.3 $ \\ \hline
        \multirow{2}{*}{\shortstack{Fashion\\MNIST}} & CNN & $ 0.40 \pm 0.01 $ & $ 85.4 \pm .3  $ \\ \cline{2-4}
         & Skeletal CNN & $ 0.39 \pm 0.01 $ & $ 86.7 \pm 0.4 $ \\  
    \end{tabular}
    \label{tab:test_loss_and_accuracy}\hfill
\end{table}
\begin{table}
    \centering
    \caption{Chararistics of both types of CNN (regular and Skeletal). We see that the two model variants have similar number of parameters. They also take similar amounts of time to train on a batch of data, despite the engineering advantage of the regular CNN which is implemented in highly-optimized GPU code. }
    \begin{tabular}{c|c|c}
         Model & \# Params & Time (s) \\ \hline
         CNN & 48900 &  $1.6 \pm .14$ \\ \hline
         Skeletal CNN & 56618 & $1.61 \pm 0.08 $ \\ 
    \end{tabular}
    \label{tab:model_char}

\end{table}

\begin{figure}
    \centering
    \includegraphics[width=\linewidth]{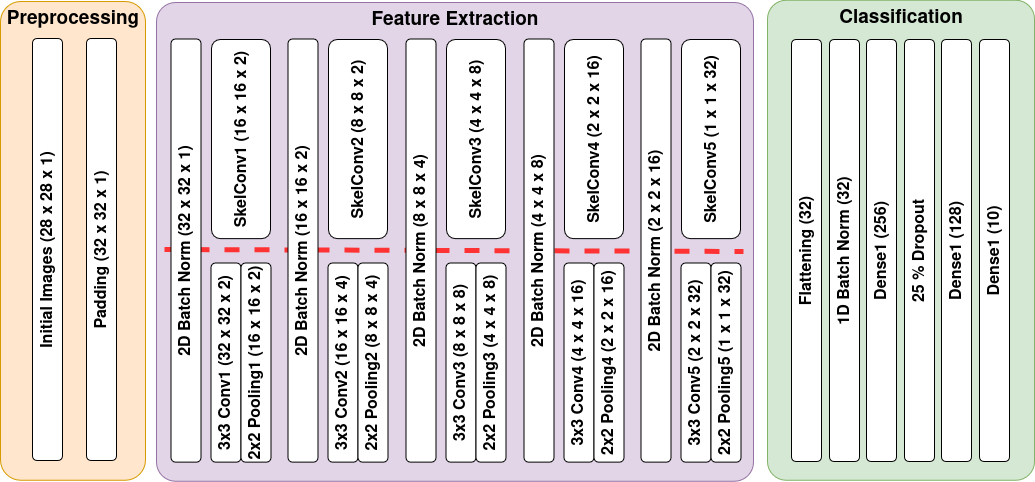}
    \caption{Schematic showing the entire pipeline used or the experiment in Section \ref{subsec:cnn_experiment}. The preprocessing and classification modules were identical between the two types of model considered. }
    \label{fig:skel_arch_detail_fig}
\end{figure}

To validate the model proposed in Section \ref{subsec:cnn_model_desc}, 
in Figure~\ref{subsec:cnn_experiment}
we examine its performance on the benchmark MNIST digit classification task \cite{deng2012mnist}, as well as the significantly more difficult ``Fashion MNIST" (or FMNIST) variant \cite{xiao2017fashion}. 
\paragraph{Architecture Details}
Each of the models we compare consisted of three main modules a) a preprocessing module, b) a feature extraction module (regular CNN or skeletal) and c) a classification module. Modules (a) and (c) were identical for all experiments: Preprocessing consisted of a single 2D batch norm layer, and classification consisted of several dense layers and a single 25\% dropout layer. See Figure \ref{fig:skel_arch_detail_fig} for details. All Convolution, Skeletal convolution, and dense layers included SiLU activations. 
\paragraph{Training Details}
We train both models for 150 epochs of 10 batches each (1 batch = 2048 examples). Models were trained to minimize cross-entropy loss using the built-in Adam \cite{kingma2014adam} implementation in Pytorch \cite{paszke2019pytorch} with default hyperparameters and initial learning rate of 0.001.  During each training run, we split the training data further into 50000 training samples and 10000 validation samples (not used for training). Random seeds were fixed to ensure these random splits were the same for both architectures. For each training run of each model, we utilized early stopping, and evaluated the model with the parameters that performed best on the validation data against the test set (see next paragraph). 
\paragraph{Evaluation}
We evaluate both the skeletal product-based CNN and the default CNN by computing the average loss and accuracy on the standard held-out MNIST (FMNIST) test set. The results are reported in Table \ref{tab:test_loss_and_accuracy}. Each number represents average performance ($\pm$ std. dev.) on the test set over 32 random initializations. We also report the average time per batch (over all batches processed) and count of parameters for each architecture in Table \ref{tab:model_char}.
\paragraph{Interpretation}
We see that the architecture constructed with a skeletal graph product performs similarly to the other CNN in terms of both loss and accuracy
in Figure~\ref{subsec:cnn_experiment}. 
The skeletal CNN is slightly slower 
by ``wall clock'' time
in current practice; 
we note here that the default CNN benefits from highly optimized CUDA instructions for convolution, whereas our model is implemented as a generic graph neural network. Despite this, the skeletal CNN is as fast as the default version. It also possess better scaling properties, as all of the skeletal product operations are implemented via sparse matrix multiplication. 

The main advantage of our model that it is specified algebraically in a way that could be manipulated symbolically, permitting the exploration of other highly structured
neural network architectures, either by human mathematicians
or automatically. For example one could start with conventional
graph product descriptions of high expressive power,
similar to Generalized Hough Transforms,
and skeletonize them in various alternative ways 
searching for both computational efficiency
and reduced parameter spaces conducive to statistical learning.
In addition, the skeletal product description may lead to continuum-limit geometrical descriptions as in Section~\ref{contin_skeletal} of the
neural architecture, with attendant mathematical benefits.

\subsection{Skeletal multigrid}
\label{subsec:skel_multigrid}

To further validate skeletal products as methods for constructing lightweight architectures for numerical algorithms, we investigate two methods for using the hierarchy of a skeletal product lineage in a multigrid-style solver. In this subsection we describe this algorithm and compare it to similar preexisting methods known in the multigrid literature. In the next subsection we demonstrate that this algorithm is competitive on two boundary value problems of varying difficulty. 

\subsubsection{Skeletal multigrid algorithms}
\label{subsubsec:skel_mg_defn}

We first describe two variant algorithms for a multigrid-like procedure that uses the skeletal graph product. We discuss this algorithm in the specific case of a boundary value problem on 2D grids, which we factor as the box product of two 1D grid graph lineages. In principle all of the ideas in this section could be easily extended to higher-dimensional graph products. 

The first variant resembles classical geometric multigrid: We want to solve the problem $\mathbb{A}_n x = b$ for a family of grids $\mathbb{A}_1, \mathbb{A}_2 \ldots \mathbb{A}_n $ (with $\mathbb{A}_n$ being the finest grid). Similar to the problem setup of Oosterlee et al. \cite{oosterlee1995robustness}, we assume that each $\mathbb{A}_i$ may be factored as the graph box product of two 1D grids i.e. $\mathbb{A}_i = A^{[1]}_i \Box  A^{[2]}_i$. Furthermore, assume that we have a set of orthogonal prolongation/restriction operators $P^{[d]}_{i,i+1}$ for each 1D lineage such that $ A^{[d]}_i = {\left(P^{[d]}_{i,i+1}\right)}^T A^{[1]}_{i+1} P^{[d]}_{i,i+1} $. As Osterlee et al. previously observed, coarsening along one spatial dimension at a time produces a semiordering of related 2D grids. Namely, for any $i_1 + i_2 = i_1' + i_2' + 1 $ we have 
\begin{equation}
A^{[1]}_{i_1} \Box  A^{[2]}_{i_2} = {\left( P^{[1]}_{i_1',i_1} \otimes P^{[2]}_{i_2',i_2} \right)}^T \left( A^{[1]}_{i_1'} \Box  A^{[2]}_{i_2'} \right) \left( P^{[1]}_{i_1',i_1} \otimes P^{[2]}_{i_2',i_2} \right)
\end{equation}
where we say that  $P^{[d]}_{i,i} = I$. For any $L$, the set of grids $A^{[1]}_{i_1} \Box  A^{[2]}_{i_2}$ such that $i_1 + i_2 = L$ is exactly the $\Delta l = 0$ edges of our skeletal product. Performing smoothing steps over one level of the skeletal product is therefore equivalent to smoothing over multiple resolutions (albeit at a proportionally increased cost). 
\begin{algorithm}
\caption{Skeletal Multigrid (Recursive)}
\TitleOfAlgo{SkelMultigrid}
\label{alg:sk_mg_recur}
\DontPrintSemicolon
\LinesNumbered
\KwIn{$u^{[l_1, l_2]}_{i}$ (state at step $i$ on grid $(l_1, l_2)$),$ A^{[1]}_{i_1}, A^{[2]}_{i_2}$ (linear operators), $b$ (target)}
\KwOut{$u^{[l_1, l_2]}_{i+1}$ (state at step $i+1$ on grid $(l_1, l_2)$)}
$v = \text{SmoothingStep}(u^{[l_1, l_2]}_{i}, A^{[1]}_{i_1} \Box  A^{[2]}_{i_2}, b)$ \\
$r = b - A v $ \\
$\Delta v = \mathbf{0}$\\
\If{$l_1 > 1$}{
$r^{[1]} = \left( P^{[1]}_{i_1-1,i_1} \otimes I \right) r $ \\
$ c^{[1]} = \text{SkelMultigrid}(0,A^{[1]}_{i_1-1}, A^{[2]}_{i_2}, r^{[1]}) $ \\
$ \Delta v = \Delta v + {\left( P^{[1]}_{i_1-1,i_1} \otimes I \right)}^T c^{[1]} $\\
}
\If{$l_2 > 1$}{
$r^{[2]} = \left( I \otimes P^{[2]}_{i_2-1,i_2} \right) r $ \\
$ c^{[2]} = \text{SkelMultigrid}(0,A^{[1]}_{i_1}, A^{[2]}_{i_2-1}, r^{[2]}) $ \\
$ \Delta v = \Delta v + {\left(I \otimes P^{[2]}_{i_2-1,i_2} \right)}^T c^{[2]} $ \\
}
$u^{[l_1, l_2]}_{i+1} = \text{SmoothingStep}(v + \Delta v, A^{[1]}_{i_1} \Box  A^{[2]}_{i_2}, b)$

\end{algorithm}

The second skeletal multigrid algorithm is classical multigrid \cite{brandt2006guide}, using each level of the skeletal product as the grids on which to perform smoothing. The hierarchy of nested grids and the prolongation operators between them are constructed according to the matrix equations described in Section \ref{matrix_skeletal_box_product}. Note that if the finest grid has level numbers $(i_1, i_2)$ with $i_1 + i_2 = L$, we only include those blocks $(i_1', i_2')$ of the skeletal products graph such that $i_1' \leq i_1$, $i_2' \leq i_2$, and $i_1' + i_2' \leq L$. 

\subsubsection{Skeletal multigrid comparison}
In this section, we compare the skeletal multigrid algorithms described in Section \ref{subsubsec:skel_mg_defn}. 

\begin{figure}
    \centering
    \includegraphics[width=0.23\linewidth]{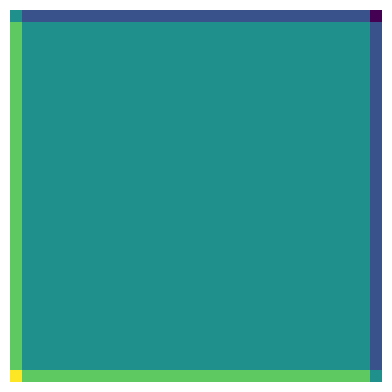} 
    \includegraphics[width=0.23\linewidth]{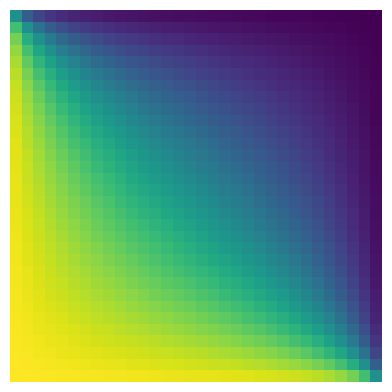} 
    \includegraphics[width=0.23\linewidth]{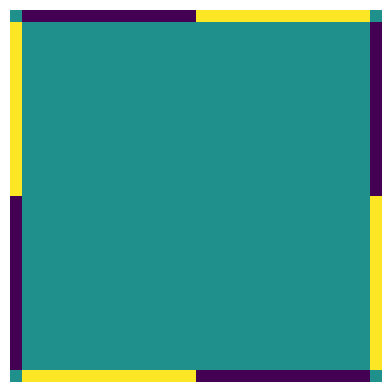} 
    \includegraphics[width=0.23\linewidth]{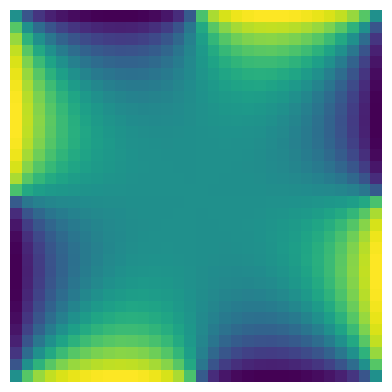} \\
    \caption{Boundary problems and solutions found by a multigrid solver. For comparisons of work performed by various algorithms to find these solutions, see Figure \ref{fig:skel_multigrid_comp}.}
    \label{fig:skel_multigrid_probs}
\end{figure}

\paragraph{Test problems}
We examine the performance of our skeletal multigrid algorithm on two boundary value problems in two dimensions. In each we seek to solve the sparse Dirichlet problem $A x = b$ where $b$ is a vector of boundary conditions, $x$ is our solution, and $A$ is the Laplacian of a 2D grid, i.e. a discretization of the unit square. Both problems are characterized by error modes that are $45\deg$ from the main axes of the grid. Boundary condition 1 has two adjacent (i.e. left and lower) edges filled with positive values; boundary condition 2 has an alternating pattern of positive-negative values around the circumference of the grid. Each problem, as well as the solution found by a Gauss-Seidel solver, are visualized in Figure \ref{fig:skel_multigrid_probs}.

We compare the performance of the following four algorithms:
\begin{itemize}
\item Gauss-Seidel smoothing;
\item classic Geometric Multigrid;
\item Skeletal Multigrid (recursive);
\item Skeletal Multigrid (levelwise).
\end{itemize}

\begin{figure}
    \centering
    \includegraphics[width=0.47\linewidth]{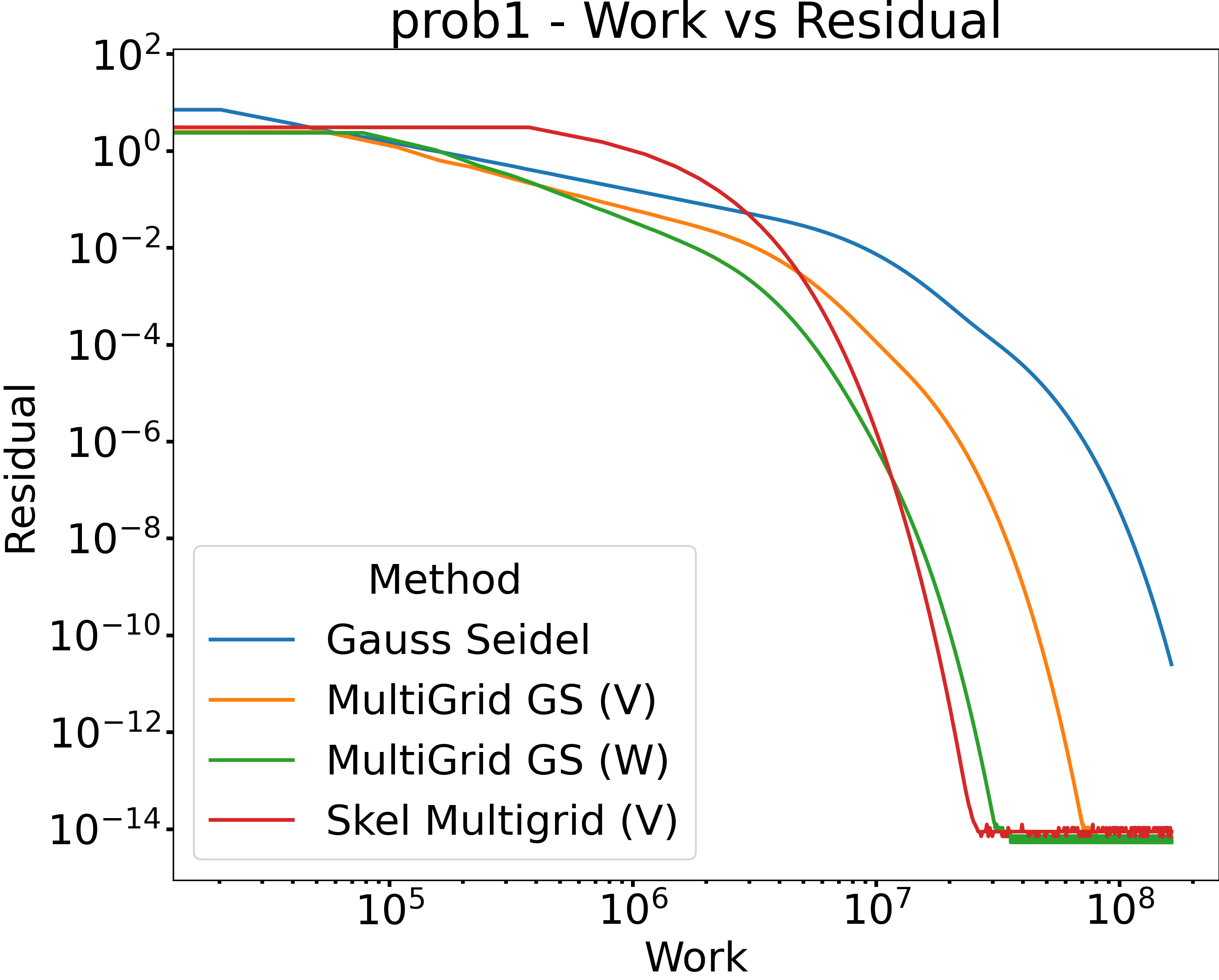} 
    \includegraphics[width=0.47\linewidth]{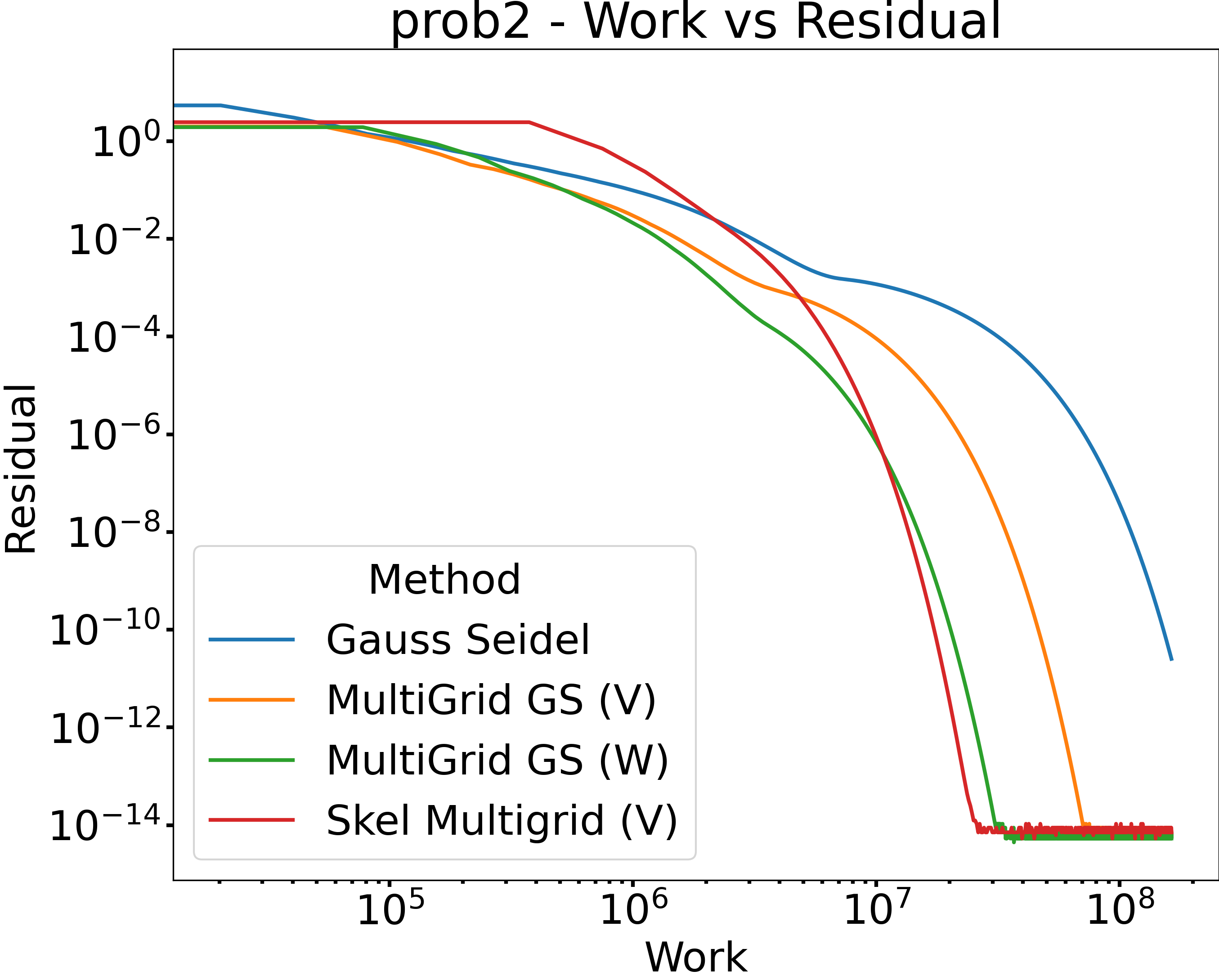} \\
    \caption{Comparison of our skeletal multigrid algorithm on the two boundary value problems shown in Figure \ref{fig:skel_multigrid_probs}. We see that the recursive skeletal multigrid V-cycle outperforms other methods, including both classical multigrid (V and W sycles) as well as Gauss-Seidel. Not shown: Skeletal Multigrid (Recursive) W-cycle and ``levelwise'' skeletal multigrid variant, neither of which were competitive with these methods.}
    \label{fig:skel_multigrid_comp}
\end{figure}

For each of the multigrid algorithms, we examine both V-cycle and W-cycle variants. We assume that the cost of a single smoothing step, at any scale, is proportional to the number of nonzero entries in the smoothing matrix. We compare work performed vs. finest-scale residual for each of the algorithms we consider. The results of this experiment are visualized in Figure \ref{fig:skel_multigrid_comp}. We see that the recursive skeletal multigrid approach outperforms classic multigrid (which of course outperforms Gauss-Seidel). The ``levelwise'' skeletal multigrid approach of smoothing on all grids with the same level number simultaneously converged more slowly than even Gauss-Seidel; in agreement with prior work on semi-coarsening \cite{oosterlee1995robustness}, we hypothesize that this is due to factoring the original grid in directions that are not aligned with the predominant error modes, but more experiments are needed to determine whether this is happening in our algorithm. Additionally, W-cycles of the recursive skeletal multigrid algorithm are prohibitively expensive due to the fact that each grid with level number $L$ has two descendants with level number $L-1$ and no work is shared between these coarser grids. We hope to ameliorate this with a dynamic programming approach in future work. 

\subsection{Continuous skeletal products}
\label{contin_skeletal}

Graphs can arise as discretizations of continuum geometries such as
manifolds or cell complexes, by way of meshes
that can be refined to varying spatial resolutions.
So it is interesting and potentially fruitful
to ask for the analog of a skeletal product, in continuous spaces.
We develop an example analogous to the graph lineage of
Figure~\ref{fig:glin_fourier}.

We begin with the continuous analog of a graph lineage
of path graphs that discretize the 1D real line.
It is the
Poincar\'e half-plane, with hyperbolic (negative curvature) 
metric: $ds^2 = (dx^2 + dz^2)/z^2$, 
for $0 < z < +\infty, -\infty < x < + \infty$.
This 2D manifold can be taken as a continuous scale space,
with $z$ as the scale axis.
A variety of local meshes can be embedded
in this space \cite{Margenstern}, generating approximating graph lineages.
Discretizations of the Poincar\'e half-plane have been used to construct 
neural networks \cite{Mjolsness_thesis}, 
with the continuum limit being a ``continuous computational medium''.


The conventional {\it Cartesian product} of two such half-planes 
is defined by the metric in a single global coordinate patch: 
$$ds^2 = (dx^2 + dz^2)/z^2 + (dy^2 + dw^2)/w^2 , $$
for $0 < z,w < +\infty, -\infty < x,y < + \infty$.

The {\it skeletal product} of two real line scale space half-planes, 
coordinatized by $(x,z)$ and $(y,w)$,
can be constructed from the Cartesian product of the Poincar\'e half-planes
by a 45 degree coordinate rotation in $(\log z, \log w)$. This rotation 
(times a $1/\sqrt{2}$ scaling factor)
to 
$(\log \mu , \log \nu)=$
$((\log z + \log w)/2, (\log z - \log w)/2)$ can be
expressed by $(\mu, \nu) = (\sqrt{z w}, \sqrt{z/w})$ with $\mu$
being the new joint scale axis
and $\nu$ the emergent $(x,y)$ tradeoff axis. Then
the continuous skeletal product metric becomes:
\begin{equation}
ds^2 
    = d x^2/(\mu \nu)^2 + dy^2/(\mu/\nu)^2 + 2 d\mu^2/\mu^2 + 2 d\nu^2/\nu^2.
\end{equation}
Slices of constant $\mu$ play the role of levels
or grades in a graded graph skeletal product.

\subsection{Skeletal function spaces}
\label{section:skeletal_function_space}

If $A, B$ are ordinary sets rather than graphs,
then a standard notation for the set
$A \rightarrow B$ of functions from $A$ to $B$ is:
$B^A = \times_{a \in A} B$ where $\times$ 
is the Cartesian product of sets.
In many type theories,
the space of functions $A \rightarrow B \simeq \prod_{a \in A}$
where $\prod$ is the categorical product.
Intrinsic to these notations is an ordering of the elements of $A$,
since the products are permutation-equivalent but not permutation invariant.

For graphs, we prefer the box (Cartesian) product

$$G_1 \rightarrow G_2 \equiv \Box_{v \in G_1} G_2$$

\noindent
because the edges in the resulting graph link two functions
that take the same value everywhere except on one vertex,
thus encoding a precursor of the idea of continuity 
(by a Hamming distance of 1) in the result graph.
Diffusion from a point source
on finite dimensional grid graphs (box product of 1D path graphs), 
for example, defines a distance measure that
tends towards circular contours as the grid 
graphs grow in extent. Of course the function definition above
grows exponentially in cardinality with the cardinality of $V(G_1)$.
The box product can be justified in category theoretical terms
not as the categorical product (which is the cross product of graphs)
but as the ``funny tensor product'' \cite{foltz1980algebraic,weber2013free}.
A particular node in $G_1 \rightarrow G_2$ is then a particular
function from nodes of $G_1$ to nodes of $G_2$,
and closely nearby functions are connected, 
in a ``graph function space''.

How can these ideas be generalized to graded graphs and graph lineages?

One simple step is, when $G_1$ and $G_2$ are graded graphs,
to replace the box product with the skeletal box product
and consider the object

$$G_1 \rightarrow G_2 \equiv \hat{\Box}_{v \in G_1} G_2 $$

\noindent
where the subscript $v$ runs over all vertices in $G_1$.
In this case the level numbers of the graded graph will
be $L = \sum_{v \in V(G_1)} l_{G_2(v)}$. But this cannot be
what we want because there are usually infinitely many vertices in
$V(G_1)$, so the cardinality of the box product index is not 
yet under control. If on the other hand we restrict the
index $v$ to run over one level $G_{1, \, l}$ of $G_1$ 
then we have control over the cost of the skeletal product
but we have only a fragment of the domain $G_1$, namely that specific level.
Each node of such a graded graph would only be an
{\it approximating function}, limited by the choice of 
levels $l$ of $G_1$ and $L$ of the skeletal product.

We could take a direct sum 
$\oplus_{l=0}^\infty \hat{\Box}_{v \in G_1^{(l)}} G_2$
of such box product graded graphs
as the level number of $G_1$ is varied, but then the cardinality
of the lowest level $L=0$ would be the number of levels in $G_1$
which is infinite. A solution to that problem is to shift
each skeletal box product summand upwards in the $L$ hierarchy by an integer
$S_{G_{1 l}} \ge l$ that increases monotonically with 
the $G_1$ level number $l$, so now
$L = S_{G_{1, \, l}} + \sum_{v \in G_1^{(l)}} l_{G_2(v)} .$
One example of such a function is $l$ itself,
though that choice underweights the importance of level numbers
in the domain vs. the range.

A better choice of shift $S_{G_{1 l}}$ is the $l^{\text{th}}$ ``volume'' $V_l$ =
the number of vertices of $G_1$ with level number $\le l$,
since that reflects the number of terms summed in ``$\sum_{v \in G_1^{(l)}}$'';
by construction it starts out at 1 and increases monotonically. 
Then 
\begin{equation}
L = V_l + \sum_{v \in G_1^{(l)}} l_{G_2(v)} .
\end{equation}

In favor of this choice is that ``frontiers'' $F$ of a 
suitably restricted graded graph, including but not limited to its levels,
themselves form a graded graph indexed by their volumes $V_F$.
Here a {\it frontier} is an edge cutset on the graded graph that separates
the corresponding (directed) graph into a finite component of the root 
(more generally finite components of the root nodes),
versus all the components that connect to infinity. 
Two frontiers of volumes $V$ and $V+1$ are connected
if they differ by the addition or deletion of just one node.
Each frontier has a corresponding set of leaf nodes, $\text{Lv}(F)$,
of the component DAG that includes the root.

In more detail, a {\it frontier} is an edge cutset whose removal disconnects the graded graph into one or more finite components each of whose nodes are connected by a path, increasing in level number by 1 at each edge, from a level zero node; and one or more infinite components each of whose nodes have nonzero level number and are connected by a path, increasing in level number by 1 at each edge, whose level numbers increase without bound; and for which all such paths to infinity in the graded graph, starting from a node in an infinite component, are contained in that infinite component. Each edge in the cutset must connect a finite component to an infinite component. The nodes on the finite side of any edge in the cutset comprise the “leaves” of the frontier.
The summed cardinality of the finite components is the ``volume'' of the frontier.

Nodes may be transferred one at a time from the infinite side of some frontier edge to the finite side (where they become leaves), locally redefining the cutset and hence the frontier. 
For example, nodes can be added in order of lower to higher level numbers while maintaining the frontier properties.
Frontiers related by such a single node transfer are connected to each other in the graded graph of frontiers indexed by their finite volume, 
which increases by one with each transfer.

Given this graded graph ${\cal F}$ of frontiers $F \in {\cal F}(F)$, 
replace each frontier-representing node with the corresponding 
skeletal box product $\hat{\Box}_{v \in \text{Lv}(F)} G_2(v)$.
We may write this new graded graph as:
\begin{equation}
\label{arrowhat}
(G_1 \hat{\rightarrow} G_2)
\cong 
\Bigg[ 
    \bigoplus_{V=0}^{+\infty}  
    \quad \bigoplus_{F \in {\cal F}(G_1) | V_F = V} 
    \hat{\Box}_{v \in \text{Lv}(F)} G_2(v) 
  \Bigg] , 
\end{equation}
where the $\bigoplus_{\cal F}$ over frontiers is the discrete sum 
of skeletal box product graded graphs,
but the $\bigoplus_V$ over integers $V$ shifts and organizes its summands
by level number $V$,
and connects up neighboring frontiers $F$,
following the template of the graded graph of frontiers.
Combined, $\bigoplus_V \bigoplus_{\cal F}$
sums over frontiers of the frontier graded graph,
with each summand being a skeletal box product over
the leaves of a frontier.

If we equip this graded graph $(G_1 \hat{\rightarrow} G_2)$ 
with a (single) label function
from each node to an ordered list of the nodes in its frontier $F$, 
then the $(G_1 \hat{\rightarrow} G_2)$ function values 
of those variables (for a skeletal product or range node)
can be read off. For a frontier-only (domain) node, the refinement
relationships between frontiers can instead be followed.
An additional bit of label information can be used to specify
which kind of node, domain or range, we are currently visiting.


The refinement links between similar frontiers can be extended
upwards to refinement links between approximating functions in
their box products, in the $\bigoplus$ over frontiers.
What is required is that for all the $G_1$ vertices
in common between two frontiers, the corresponding
function values in $G_2$ are the same; and also for the
one node that is added, the value in $G_2$ is 
the same as the value of its parental node in the other frontier.
Of course the next level of refinement of this value in $G_2$
can be different.
Because of these extra details, Equation (\ref{arrowhat})
isn't yet quite as algebraically self-sufficient as it looks.

Thus the vertices of the graded graph (\ref{arrowhat}) comprise a function space
between graded graphs; the edges within a level comprise the minimal
topology information for neighborliness in this space; and the
further edges between levels as specified above provide the minimal
topology information for the refinement of function approximations.

Finally, a function from the range-labelled nodes of 
$(G_1 \hat{\rightarrow} G_2)$
to some range such as ${\mathbb R}$
is a potentially cost-effective computational representation 
of a functional. One can impose Cauchy-like conditions on
such functions so that infinite-precision limits
can be taken along monotonic paths to infinity.

\section{Conclusions}

We have introduced a framework for defining families
of interrelated graphs,
called graph lineages, that grow with exponential-like
cost $O(b^{l^{1+\epsilon}})$ upper bound
for some $b\ge 1$ and all $\epsilon>0$
as a function of an integer-valued index $l$, and over which
a variety of useful architecture-forming operations
can be defined. The conventional graph sum
$\oplus$ and product
$\times$ and $ \Box$ binary operators can be defined, 
but these graph products are 
costly in
space cost measures since $b$ grows exponentially in the number of factors. So we used graph lineages 
and category theory diagrams (Propositions 2 and 3)
to define 
and prove universality properties of novel
``skeletal'' versions of
graph cross (or direct) and box (or Cartesian) products 
$\hat{\times}$ and $\hat{\Box}$,
which are much more affordable
than the conventional graph products
in their space cost ($b$ is just the maximum of the factor $b$s, Proposition 4). 

We showed that $\hat{\Box}$ is associative but
$\hat{\times}$ is not, so we defined n-way $\hat{\times}$
products that bound n-way combinations of the binary ones above and below
by edge subset inclusion.
We exhibited a continuous manifold analog of the skeletal box product,
for two Poincar\'e half-plane hyperbolic manifolds.
We demonstrated the utility of skeletal graph products
in neural network architecture specification
and in algebraic multigrid problems.
We also defined useful ``thickening'' (Proposition 1)
and ``frontier'' unary operators on graph lineages, 
and used the latter
to generalize from skeletal graph box products to skeletal
(hence relatively cost-limited)
graph functions.

\section*{Acknowledgements}
This work was supported in part by the
U.S. NIH/NIDA Brain Initiative Grant 1RF1DA055668-01;
the UC Multi-Campus Research Programs and Initiatives of the University of California, Grant Number M23PR5854; the UCI Donald Bren School of Information and Computer
Sciences; and the
UC Southern California Hub, with funding from the
UC National Laboratories division of the University
of California Office of the President. This work was additionally supported by the Office of the Dean of the College at Colorado College, as well as the Colorado College Department of Mathematics and Computer Science's Euclid Fund.

\let\oldsection\section
\newpage
\appendix
\makeatletter
\setcounter{section}{0}
\renewcommand \thesection{S\@arabic\c@section}
\renewcommand\thetable{S\@arabic\c@table}
\renewcommand \thefigure{S\@arabic\c@figure}
\makeatother

\section{Distributive properties of graph products}
\label{sec:dist_prop_graph_prod}
From the corresponding adjacency matrices, we have the distributive laws
\begin{equation}
G_1 \Box (G_2 \oplus G_2)  \cong  (G_1 \Box G_2) \oplus (G_1 \Box  G_3)
\end{equation}
and
\begin{equation}
G_1 \times (G_2 \oplus G_2)  \cong  (G_1 \times G_2) \oplus (G_1 \times  G_3) .
\end{equation}

{\it Proofs:} Using the matrix outer product $\otimes$,
we can transform the graph $G_1 \Box (G_2 \oplus G_3)$
into the graph $ (G_1 \Box G_2) \oplus (G_1 \Box  G_3)$
using matrix equational reasoning:
\begin{equation}
\begin{split}
G_1 \Box (G_2 \oplus G_3) 
\cong & G_1 \otimes 
\Big(
\begin{array}{cc}
I & 0 \\
0 & I
\end{array} \Big)
+ I \otimes
\Big(
\begin{array}{cc}
G_2 & 0 \\
0 & G_3
\end{array} \Big) \\
= & 
\Big(
\begin{array}{cc}
G_1 \otimes I & G_1 \otimes 0 = 0 \\
0 & G_1 \otimes I
\end{array} \Big)
+ I \otimes
\Big(
\begin{array}{cc}
I \otimes G_2 & I \otimes 0 = 0 \\
0 & I \otimes G_3
\end{array} \Big) \\
= & 
\Big(
\begin{array}{cc}
G_1 \otimes I + I \otimes G_2 & 0 \\
0 & G_1 \otimes I + I \otimes G_3
\end{array} \Big) \\
\cong & (G_1 \Box G_2) \oplus (G_1 \Box  G_3) . \\
\end{split}
\end{equation}
Likewise, using matrix notation for the calculation,
\begin{equation}
\begin{split}
G_1 \times (G_2 \oplus G_3) 
\cong & G_1 \otimes 
\Big(
\begin{array}{cc}
G_2 & 0 \\
0 & G_3
\end{array} \Big) \\
= & 
\Big(
\begin{array}{cc}
G_1 \otimes  G_2 & 0 \\ 
0 & G_1 \otimes G_3
\end{array} \Big)
= 
(G_1 \otimes G_2) \oplus (G_1 \otimes  G_3)\\
\cong & (G_1 \Box G_2) \oplus (G_1 \Box  G_3) . \\
\end{split}
\end{equation}

\section{Alternate proof via sparsification of graded graph products}
In this section, we will demonstrate how the skeletal graph product arises naturally as a sparsified version of the (box, cross, or boxcross) product of two graded graphs. 

Let $\mathcal{G}$ be a graded graph with grades $G_0, G_1, G_2 \ldots$ and inter-grade connections $S_{01}, S_{12}, S_{23}, \ldots$. Similarly, let $\mathcal{H}$ be a graded graph with grades $H_i$ and inter-grade connections $P_j$. The graded graphs $\mathcal{G}$ and $\mathcal{H}$ may be represented in adjacency matrix form by the following infinite-dimensional matrices:
\[
A(\mathcal{G}) = \left[ \begin{array}{cccccc}
G_0 & S_{01} & & & &\\
S_{01}^T & G_1 & S_{12} & & & \cdots\\
 & S_{12}^T & G_2 & S_{23} & & \\
&  & S_{23}^T & G_3 & & \\
& \vdots & & & & \ddots
\end{array} \right]
\]
\[
A(\mathcal{H}) = \left[ \begin{array}{cccccc}
H_0 & P_{01} & & & &\\
P_{01}^T & H_1 & P_{12} & & & \cdots\\
 & P_{12}^T & H_2 & P_{23} & & \\
&  & P_{23}^T & H_3 & & \\
& \vdots & & & & \ddots
\end{array} \right]
\]

For any graph $G$, let $I_{G}$ denote an identity matrix with the same dimensions as $G$'s adjacency matrix. For the restriction/prolongation maps $P_{ij}$ and $S_{ij}$, we will will reverse the indices to indicate the transpose map, i.e. $S_{ij} = S_{ji}^T$.

\subsection{Cross product}
\label{matrix_skeletal_cross_product} 

The cross product of $\mathcal{G}$ and $\mathcal{H}$ is constructed by taking the Kronecker product of their adjacency matrices,
in accordance with the relevant formula for graph adjacency matrices:

\begin{align*}
\mathcal{G} \times \mathcal{H} &= A(\mathcal{G}) \otimes A(\mathcal{H)} \\
&= \left[ \begin{array}{cccccc}
G_0 & S_{01} & & & &\\
S_{01}^T & G_1 & S_{12} & & & \cdots\\
 & S_{12}^T & G_2 & S_{23} & & \\
&  & S_{23}^T & G_3 & & \\
& \vdots & & & & \ddots
\end{array} \right] \otimes \left[ \begin{array}{cccccc}
H_0 & P_{01} & & & &\\
P_{01}^T & H_1 & P_{12} & & & \cdots\\
 & P_{12}^T & H_2 & P_{23} & & \\
&  & P_{23}^T & H_3 & & \\
& \vdots & & & & \ddots
\end{array} \right] \\
\end{align*}
\noindent=\resizebox{\linewidth}{!}{$\left[\begin{array}{c||c||c}
 \begin{array}{cccccc}
G_0 \otimes H_0 & G_0 \otimes P_{01} & & & &\\
G_0 \otimes P_{01}^T & G_0 \otimes H_1 & G_0 \otimes P_{12} & & & \cdots\\
 & G_0 \otimes P_{12}^T & G_0 \otimes H_2 & G_0 \otimes P_{23} & & \\
&  & G_0 \otimes P_{23}^T & G_0 \otimes H_3 & & \\
& \vdots & & & & \ddots
\end{array}  & \begin{array}{cccccc}
S_{01} \otimes H_0 & S_{01} \otimes P_{01} & & & &\\
S_{01} \otimes P_{01}^T & S_{01} \otimes H_1 & S_{01} \otimes P_{12} & & & \cdots\\
 & S_{01} \otimes P_{12}^T & S_{01} \otimes H_2 & S_{01} \otimes P_{23} & & \\
&  & S_{01} \otimes P_{23}^T & S_{01} \otimes H_3 & & \\
& \vdots & & & & \ddots
\end{array}  & \cdots \\
\hline
\hline
 \begin{array}{cccccc}
S_{01}^T \otimes H_0 & S_{01}^T \otimes P_{01} & & & &\\
S_{01}^T \otimes P_{01}^T & S_{01}^T \otimes H_1 & S_{01}^T \otimes P_{12} & & & \cdots\\
 & S_{01}^T \otimes P_{12}^T & S_{01}^T \otimes H_2 & S_{01}^T \otimes P_{23} & & \\
&  & S_{01}^T \otimes P_{23}^T & S_{01}^T \otimes H_3 & & \\
& \vdots & & & & \ddots
\end{array} &   \begin{array}{cccccc}
G_1 \otimes H_0 & G_1 \otimes P_{01} & & & &\\
G_1 \otimes P_{01}^T & G_1 \otimes H_1 & G_1 \otimes P_{12} & & & \cdots\\
 & G_1 \otimes P_{12}^T & G_1 \otimes H_2 & G_1 \otimes P_{23} & & \\
&  & G_1 \otimes P_{23}^T & G_1 \otimes H_3 & & \\
& \vdots & & & & \ddots
\end{array}& \cdots \\
\hline
\hline
\vdots & \vdots & \ddots
\end{array} \right]$
}

The result is a new graph, whose nodes represent pairs of nodes $(u,v)$ where $u \in \mathcal{G}$ and $v \in \mathcal{H}$. We define level numbers in this new graph to be sums of each pair's level number in $\mathcal{G}$ and $\mathcal{H}$. 
If we permute the rows and columns of the above block matrix to gather diagonal blocks of the same level number, we have:\\
\resizebox{\textwidth}{!}{
$\begin{array}{|c|cc|ccc|cccc|ccccc|ccc|}
\hline
G_{0} \otimes H_{0} & G_{0} \otimes P_{01} & S_{01} \otimes H_{0} &   & S_{01} \otimes P_{01} &   &   &   &   &   &   &   &   &   &   &   &   & \cdots \\
\hline
G_{0} \otimes P_{10} & G_{0} \otimes H_{1} & S_{01} \otimes P_{10} & G_{0} \otimes P_{12} & S_{01} \otimes H_{1} &   &   & S_{01} \otimes P_{12} &   &   &   &   &   &   &   &   &   & \cdots \\
S_{10} \otimes H_{0} & S_{10} \otimes P_{01} & G_{1} \otimes H_{0} &   & G_{1} \otimes P_{01} & S_{12} \otimes H_{0} &   &   & S_{12} \otimes P_{01} &   &   &   &   &   &   &   &   & \cdots \\
\hline
  & G_{0} \otimes P_{21} &   & G_{0} \otimes H_{2} & S_{01} \otimes P_{21} &   & G_{0} \otimes P_{23} & S_{01} \otimes H_{2} &   &   &   & S_{01} \otimes P_{23} &   &   &   &   &   & \cdots \\
S_{10} \otimes P_{10} & S_{10} \otimes H_{1} & G_{1} \otimes P_{10} & S_{10} \otimes P_{12} & G_{1} \otimes H_{1} & S_{12} \otimes P_{10} &   & G_{1} \otimes P_{12} & S_{12} \otimes H_{1} &   &   &   & S_{12} \otimes P_{12} &   &   &   &   & \cdots \\
  &   & S_{21} \otimes H_{0} &   & S_{21} \otimes P_{01} & G_{2} \otimes H_{0} &   &   & G_{2} \otimes P_{01} & S_{23} \otimes H_{0} &   &   &   & S_{23} \otimes P_{01} &   &   &   & \cdots \\
  \hline
  &   &   & G_{0} \otimes P_{32} &   &   & G_{0} \otimes H_{3} & S_{01} \otimes P_{32} &   &   & G_{0} \otimes P_{34} & S_{01} \otimes H_{3} &   &   &   &   & S_{01} \otimes P_{34} & \cdots \\
  & S_{10} \otimes P_{21} &   & S_{10} \otimes H_{2} & G_{1} \otimes P_{21} &   & S_{10} \otimes P_{23} & G_{1} \otimes H_{2} & S_{12} \otimes P_{21} &   &   & G_{1} \otimes P_{23} & S_{12} \otimes H_{2} &   &   &   &   & \cdots \\
  &   & S_{21} \otimes P_{10} &   & S_{21} \otimes H_{1} & G_{2} \otimes P_{10} &   & S_{21} \otimes P_{12} & G_{2} \otimes H_{1} & S_{23} \otimes P_{10} &   &   & G_{2} \otimes P_{12} & S_{23} \otimes H_{1} &   &   &   & \cdots \\
  &   &   &   &   & S_{32} \otimes H_{0} &   &   & S_{32} \otimes P_{01} & G_{3} \otimes H_{0} &   &   &   & G_{3} \otimes P_{01} & S_{34} \otimes H_{0} &   &   & \cdots \\
  \hline
  &   &   &   &   &   & G_{0} \otimes P_{43} &   &   &   & G_{0} \otimes H_{4} & S_{01} \otimes P_{43} &   &   &   & G_{0} \otimes P_{45} & S_{01} \otimes H_{4} & \cdots \\
  &   &   & S_{10} \otimes P_{32} &   &   & S_{10} \otimes H_{3} & G_{1} \otimes P_{32} &   &   & S_{10} \otimes P_{34} & G_{1} \otimes H_{3} & S_{12} \otimes P_{32} &   &   &   & G_{1} \otimes P_{34} & \cdots \\
  &   &   &   & S_{21} \otimes P_{21} &   &   & S_{21} \otimes H_{2} & G_{2} \otimes P_{21} &   &   & S_{21} \otimes P_{23} & G_{2} \otimes H_{2} & S_{23} \otimes P_{21} &   &   &   & \cdots \\
  &   &   &   &   & S_{32} \otimes P_{10} &   &   & S_{32} \otimes H_{1} & G_{3} \otimes P_{10} &   &   & S_{32} \otimes P_{12} & G_{3} \otimes H_{1} & S_{34} \otimes P_{10} &   &   & \cdots \\
  &   &   &   &   &   &   &   &   & S_{43} \otimes H_{0} &   &   &   & S_{43} \otimes P_{01} & G_{4} \otimes H_{0} &   &   & \cdots \\
  \hline
  &   &   &   &   &   &   &   &   &   & G_{0} \otimes P_{54} &   &   &   &   & G_{0} \otimes H_{5} & S_{01} \otimes P_{54} & \cdots \\
  &   &   &   &   &   & S_{10} \otimes P_{43} &   &   &   & S_{10} \otimes H_{4} & G_{1} \otimes P_{43} &   &   &   & S_{10} \otimes P_{45} & G_{1} \otimes H_{4} & \cdots \\
\vdots & \vdots & \vdots & \vdots & \vdots & \vdots & \vdots & \vdots & \vdots & \vdots & \vdots & \vdots & \vdots & \vdots & \vdots & \vdots & \vdots & \ddots \\
\hline
\end{array}$
}
And finally, we delete (set to zero) any blocks whose level number differs by 2 or more (e.g. we are only keeping edges whose new level number differs by -1, 0, or 1).

\noindent\resizebox{\textwidth}{!}{
$
\begin{array}{|c|cc|ccc|cccc|ccccc|ccc|}
\hline
G_{0} \otimes H_{0} & G_{0} \otimes P_{01} & S_{01} \otimes H_{0} &   &   &   &   &   &   &   &   &   &   &   &   &   &   & \cdots \\
\hline
G_{0} \otimes P_{10} & G_{0} \otimes H_{1} & S_{01} \otimes P_{10} & G_{0} \otimes P_{12} & S_{01} \otimes H_{1} &   &   &   &   &   &   &   &   &   &   &   &   & \cdots \\
S_{10} \otimes H_{0} & S_{10} \otimes P_{01} & G_{1} \otimes H_{0} &   & G_{1} \otimes P_{01} & S_{12} \otimes H_{0} &   &   &   &   &   &   &   &   &   &   &   & \cdots \\
\hline
  & G_{0} \otimes P_{21} &   & G_{0} \otimes H_{2} & S_{01} \otimes P_{21} &   & G_{0} \otimes P_{23} & S_{01} \otimes H_{2} &   &   &   &   &   &   &   &   &   & \cdots \\
  & S_{10} \otimes H_{1} & G_{1} \otimes P_{10} & S_{10} \otimes P_{12} & G_{1} \otimes H_{1} & S_{12} \otimes P_{10} &   & G_{1} \otimes P_{12} & S_{12} \otimes H_{1} &   &   &   &   &   &   &   &   & \cdots \\
  &   & S_{21} \otimes H_{0} &   & S_{21} \otimes P_{01} & G_{2} \otimes H_{0} &   &   & G_{2} \otimes P_{01} & S_{23} \otimes H_{0} &   &   &   &   &   &   &   & \cdots \\
  \hline
  &   &   & G_{0} \otimes P_{32} &   &   & G_{0} \otimes H_{3} & S_{01} \otimes P_{32} &   &   & G_{0} \otimes P_{34} & S_{01} \otimes H_{3} &   &   &   &   &   & \cdots \\
  &   &   & S_{10} \otimes H_{2} & G_{1} \otimes P_{21} &   & S_{10} \otimes P_{23} & G_{1} \otimes H_{2} & S_{12} \otimes P_{21} &   &   & G_{1} \otimes P_{23} & S_{12} \otimes H_{2} &   &   &   &   & \cdots \\
  &   &   &   & S_{21} \otimes H_{1} & G_{2} \otimes P_{10} &   & S_{21} \otimes P_{12} & G_{2} \otimes H_{1} & S_{23} \otimes P_{10} &   &   & G_{2} \otimes P_{12} & S_{23} \otimes H_{1} &   &   &   & \cdots \\
  &   &   &   &   & S_{32} \otimes H_{0} &   &   & S_{32} \otimes P_{01} & G_{3} \otimes H_{0} &   &   &   & G_{3} \otimes P_{01} & S_{34} \otimes H_{0} &   &   & \cdots \\
  \hline
  &   &   &   &   &   & G_{0} \otimes P_{43} &   &   &   & G_{0} \otimes H_{4} & S_{01} \otimes P_{43} &   &   &   & G_{0} \otimes P_{45} & S_{01} \otimes H_{4} & \cdots \\
  &   &   &   &   &   & S_{10} \otimes H_{3} & G_{1} \otimes P_{32} &   &   & S_{10} \otimes P_{34} & G_{1} \otimes H_{3} & S_{12} \otimes P_{32} &   &   &   & G_{1} \otimes P_{34} & \cdots \\
  &   &   &   &   &   &   & S_{21} \otimes H_{2} & G_{2} \otimes P_{21} &   &   & S_{21} \otimes P_{23} & G_{2} \otimes H_{2} & S_{23} \otimes P_{21} &   &   &   & \cdots \\
  &   &   &   &   &   &   &   & S_{32} \otimes H_{1} & G_{3} \otimes P_{10} &   &   & S_{32} \otimes P_{12} & G_{3} \otimes H_{1} & S_{34} \otimes P_{10} &   &   & \cdots \\
  &   &   &   &   &   &   &   &   & S_{43} \otimes H_{0} &   &   &   & S_{43} \otimes P_{01} & G_{4} \otimes H_{0} &   &   & \cdots \\
  \hline
  &   &   &   &   &   &   &   &   &   & G_{0} \otimes P_{54} &   &   &   &   & G_{0} \otimes H_{5} & S_{01} \otimes P_{54} & \cdots \\
  &   &   &   &   &   &   &   &   &   & S_{10} \otimes H_{4} & G_{1} \otimes P_{43} &   &   &   & S_{10} \otimes P_{45} & G_{1} \otimes H_{4} & \cdots \\
 \vdots & \vdots & \vdots & \vdots & \vdots & \vdots & \vdots & \vdots & \vdots & \vdots & \vdots & \vdots & \vdots & \vdots & \vdots & \vdots & \vdots & \ddots \\
\hline
\end{array}
$
}

\subsection{Box product}
\label{matrix_skeletal_box_product} 

The box product of $\mathcal{G}$ and $\mathcal{H}$ is constructed by taking the Kronecker sum of their adjacency matrices,
in accordance with the relevant formula for graph adjacency matrices:

\begin{align*}
\mathcal{G} \Box \mathcal{H} &= A(\mathcal{G}) \otimes I_{A(\mathcal{H)}} + I_{A(\mathcal{G})} \otimes A(\mathcal{H)} \\
&= \left[ \begin{array}{cccccc}
G_0 & S_{01} & & & &\\
S_{01}^T & G_1 & S_{12} & & & \cdots\\
 & S_{12}^T & G_2 & S_{23} & & \\
&  & S_{23}^T & G_3 & & \\
& \vdots & & & & \ddots
\end{array} \right] \otimes \left[ \begin{array}{cccccc}
I_{H_0} & & & & &\\
 & I_{H_1} & & & & \cdots\\
 &   & I_{H_2} &   & & \\
&  & & I_{H_3} & & \\
& \vdots & & & & \ddots
\end{array} \right] \\
&+ \left[ \begin{array}{cccccc}
I_{G_0} & & & & &\\
 & I_{G_1} & & & & \cdots\\
 &   & I_{G_2} &   & & \\
&  & & I_{G_3} & & \\
& \vdots & & & & \ddots
\end{array} \right] \otimes \left[ \begin{array}{cccccc}
H_0 & P_{01} & & & &\\
P_{01}^T & H_1 & P_{12} & & & \cdots\\
 & P_{12}^T & H_2 & P_{23} & & \\
&  & P_{23}^T & H_3 & & \\
& \vdots & & & & \ddots
\end{array} \right]
\end{align*}

\noindent=\resizebox{.95\textwidth}{!}{
$\begin{array}{|cccccccccccccccccc|}
\hline
G_{0} \otimes I_{H_{0}} &   &   &   &   &   &   &   &   & S_{01} \otimes I_{H_{0}} &   &   &   &   &   &   &   & \cdots \\
  & G_{0} \otimes I_{H_{1}} &   &   &   &   &   &   &   &   & S_{01} \otimes I_{H_{1}} &   &   &   &   &   &   & \cdots \\
  &   & G_{0} \otimes I_{H_{2}} &   &   &   &   &   &   &   &   & S_{01} \otimes I_{H_{2}} &   &   &   &   &   & \cdots \\
  &   &   & G_{0} \otimes I_{H_{3}} &   &   &   &   &   &   &   &   & S_{01} \otimes I_{H_{3}} &   &   &   &   & \cdots \\
  &   &   &   & G_{0} \otimes I_{H_{4}} &   &   &   &   &   &   &   &   & S_{01} \otimes I_{H_{4}} &   &   &   & \cdots \\
  &   &   &   &   & G_{0} \otimes I_{H_{5}} &   &   &   &   &   &   &   &   & S_{01} \otimes I_{H_{5}} &   &   & \cdots \\
  &   &   &   &   &   & G_{0} \otimes I_{H_{6}} &   &   &   &   &   &   &   &   & S_{01} \otimes I_{H_{6}} &   & \cdots \\
  &   &   &   &   &   &   & G_{0} \otimes I_{H_{7}} &   &   &   &   &   &   &   &   & S_{01} \otimes I_{H_{7}} & \cdots \\
  &   &   &   &   &   &   &   & G_{0} \otimes I_{...} &   &   &   &   &   &   &   &   & \cdots \\
S_{10} \otimes I_{H_{0}} &   &   &   &   &   &   &   &   & G_{1} \otimes I_{H_{0}} &   &   &   &   &   &   &   & \cdots \\
  & S_{10} \otimes I_{H_{1}} &   &   &   &   &   &   &   &   & G_{1} \otimes I_{H_{1}} &   &   &   &   &   &   & \cdots \\
  &   & S_{10} \otimes I_{H_{2}} &   &   &   &   &   &   &   &   & G_{1} \otimes I_{H_{2}} &   &   &   &   &   & \cdots \\
  &   &   & S_{10} \otimes I_{H_{3}} &   &   &   &   &   &   &   &   & G_{1} \otimes I_{H_{3}} &   &   &   &   & \cdots \\
  &   &   &   & S_{10} \otimes I_{H_{4}} &   &   &   &   &   &   &   &   & G_{1} \otimes I_{H_{4}} &   &   &   & \cdots \\
  &   &   &   &   & S_{10} \otimes I_{H_{5}} &   &   &   &   &   &   &   &   & G_{1} \otimes I_{H_{5}} &   &   & \cdots \\
  &   &   &   &   &   & S_{10} \otimes I_{H_{6}} &   &   &   &   &   &   &   &   & G_{1} \otimes I_{H_{6}} &   & \cdots \\
  &   &   &   &   &   &   & S_{10} \otimes I_{H_{7}} &   &   &   &   &   &   &   &   & G_{1} \otimes I_{H_{7}} & \cdots \\
 \hline
\end{array}$ \\
\par
}

\noindent+\resizebox{.95\textwidth}{!}{

$\begin{array}{|cccccccccccccccccc|}
\hline
I_{G_{0}} \otimes H_{0} & I_{G_{0}} \otimes P_{01} &   &   &   &   &   &   & ... &   &   &   &   &   &   &   &   & \cdots \\
I_{G_{0}} \otimes P_{10} & I_{G_{0}} \otimes H_{1} & I_{G_{0}} \otimes P_{12} &   &   &   &   &   & ... &   &   &   &   &   &   &   &   & \cdots \\
  & I_{G_{0}} \otimes P_{21} & I_{G_{0}} \otimes H_{2} & I_{G_{0}} \otimes P_{23} &   &   &   &   & ... &   &   &   &   &   &   &   &   & \cdots \\
  &   & I_{G_{0}} \otimes P_{32} & I_{G_{0}} \otimes H_{3} & I_{G_{0}} \otimes P_{34} &   &   &   & ... &   &   &   &   &   &   &   &   & \cdots \\
  &   &   & I_{G_{0}} \otimes P_{43} & I_{G_{0}} \otimes H_{4} & I_{G_{0}} \otimes P_{45} &   &   & ... &   &   &   &   &   &   &   &   & \cdots \\
  &   &   &   & I_{G_{0}} \otimes P_{54} & I_{G_{0}} \otimes H_{5} & I_{G_{0}} \otimes P_{56} &   & ... &   &   &   &   &   &   &   &   & \cdots \\
  &   &   &   &   & I_{G_{0}} \otimes P_{65} & I_{G_{0}} \otimes H_{6} & I_{G_{0}} \otimes P_{67} & ... &   &   &   &   &   &   &   &   & \cdots \\
  &   &   &   &   &   & I_{G_{0}} \otimes P_{76} & I_{G_{0}} \otimes H_{7} & ... &   &   &   &   &   &   &   &   & \cdots \\
... & ... & ... & ... & ... & ... & ... & ... & ... & ... & ... & ... & ... & ... & ... & ... & ... & \cdots \\
  &   &   &   &   &   &   &   & ... & I_{G_{1}} \otimes H_{0} & I_{G_{1}} \otimes P_{01} &   &   &   &   &   &   & \cdots \\
  &   &   &   &   &   &   &   & ... & I_{G_{1}} \otimes P_{10} & I_{G_{1}} \otimes H_{1} & I_{G_{1}} \otimes P_{12} &   &   &   &   &   & \cdots \\
  &   &   &   &   &   &   &   & ... &   & I_{G_{1}} \otimes P_{21} & I_{G_{1}} \otimes H_{2} & I_{G_{1}} \otimes P_{23} &   &   &   &   & \cdots \\
  &   &   &   &   &   &   &   & ... &   &   & I_{G_{1}} \otimes P_{32} & I_{G_{1}} \otimes H_{3} & I_{G_{1}} \otimes P_{34} &   &   &   & \cdots \\
  &   &   &   &   &   &   &   & ... &   &   &   & I_{G_{1}} \otimes P_{43} & I_{G_{1}} \otimes H_{4} & I_{G_{1}} \otimes P_{45} &   &   & \cdots \\
  &   &   &   &   &   &   &   & ... &   &   &   &   & I_{G_{1}} \otimes P_{54} & I_{G_{1}} \otimes H_{5} & I_{G_{1}} \otimes P_{56} &   & \cdots \\
  &   &   &   &   &   &   &   & ... &   &   &   &   &   & I_{G_{1}} \otimes P_{65} & I_{G_{1}} \otimes H_{6} & I_{G_{1}} \otimes P_{67} & \cdots \\
  &   &   &   &   &   &   &   & ... &   &   &   &   &   &   & I_{G_{1}} \otimes P_{76} & I_{G_{1}} \otimes H_{7} & \cdots \\
 \vdots & \vdots & \vdots & \vdots & \vdots & \vdots & \vdots & \vdots & \vdots & \vdots & \vdots & \vdots & \vdots & \vdots & \vdots & \vdots & \vdots & \ddots \\
 \hline
\end{array}$
}

\noindent=\resizebox{.95\textwidth}{!}{

$\begin{array}{|cccccccccccccccccc|}
\hline
G_{0} \otimes I_{H_{0}} + I_{G_{0}} \otimes H_{0} & I_{G_{0}} \otimes P_{01} &   &   &   &   &   &   & ... & S_{01} \otimes I_{H_{0}} &   &   &   &   &   &   &   & \cdots \\
I_{G_{0}} \otimes P_{10} & G_{0} \otimes I_{H_{1}} + I_{G_{0}} \otimes H_{1} & I_{G_{0}} \otimes P_{12} &   &   &   &   &   & ... &   & S_{01} \otimes I_{H_{1}} &   &   &   &   &   &   & \cdots \\
  & I_{G_{0}} \otimes P_{21} & G_{0} \otimes I_{H_{2}} + I_{G_{0}} \otimes H_{2} & I_{G_{0}} \otimes P_{23} &   &   &   &   & ... &   &   & S_{01} \otimes I_{H_{2}} &   &   &   &   &   & \cdots \\
  &   & I_{G_{0}} \otimes P_{32} & G_{0} \otimes I_{H_{3}} + I_{G_{0}} \otimes H_{3} & I_{G_{0}} \otimes P_{34} &   &   &   & ... &   &   &   & S_{01} \otimes I_{H_{3}} &   &   &   &   & \cdots \\
  &   &   & I_{G_{0}} \otimes P_{43} & G_{0} \otimes I_{H_{4}} + I_{G_{0}} \otimes H_{4} & I_{G_{0}} \otimes P_{45} &   &   & ... &   &   &   &   & S_{01} \otimes I_{H_{4}} &   &   &   & \cdots \\
  &   &   &   & I_{G_{0}} \otimes P_{54} & G_{0} \otimes I_{H_{5}} + I_{G_{0}} \otimes H_{5} & I_{G_{0}} \otimes P_{56} &   & ... &   &   &   &   &   & S_{01} \otimes I_{H_{5}} &   &   & \cdots \\
  &   &   &   &   & I_{G_{0}} \otimes P_{65} & G_{0} \otimes I_{H_{6}} + I_{G_{0}} \otimes H_{6} & I_{G_{0}} \otimes P_{67} & ... &   &   &   &   &   &   & S_{01} \otimes I_{H_{6}} &   & \cdots \\
  &   &   &   &   &   & I_{G_{0}} \otimes P_{76} & G_{0} \otimes I_{H_{7}} + I_{G_{0}} \otimes H_{7} & ... &   &   &   &   &   &   &   & S_{01} \otimes I_{H_{7}} & \cdots \\
... & ... & ... & ... & ... & ... & ... & ... & G_{0} \otimes I_{...} + ... & ... & ... & ... & ... & ... & ... & ... & ... & \cdots \\
S_{10} \otimes I_{H_{0}} &   &   &   &   &   &   &   & ... & G_{1} \otimes I_{H_{0}} + I_{G_{1}} \otimes H_{0} & I_{G_{1}} \otimes P_{01} &   &   &   &   &   &   & \cdots \\
  & S_{10} \otimes I_{H_{1}} &   &   &   &   &   &   & ... & I_{G_{1}} \otimes P_{10} & G_{1} \otimes I_{H_{1}} + I_{G_{1}} \otimes H_{1} & I_{G_{1}} \otimes P_{12} &   &   &   &   &   & \cdots \\
  &   & S_{10} \otimes I_{H_{2}} &   &   &   &   &   & ... &   & I_{G_{1}} \otimes P_{21} & G_{1} \otimes I_{H_{2}} + I_{G_{1}} \otimes H_{2} & I_{G_{1}} \otimes P_{23} &   &   &   &   & \cdots \\
  &   &   & S_{10} \otimes I_{H_{3}} &   &   &   &   & ... &   &   & I_{G_{1}} \otimes P_{32} & G_{1} \otimes I_{H_{3}} + I_{G_{1}} \otimes H_{3} & I_{G_{1}} \otimes P_{34} &   &   &   & \cdots \\
  &   &   &   & S_{10} \otimes I_{H_{4}} &   &   &   & ... &   &   &   & I_{G_{1}} \otimes P_{43} & G_{1} \otimes I_{H_{4}} + I_{G_{1}} \otimes H_{4} & I_{G_{1}} \otimes P_{45} &   &   & \cdots \\
  &   &   &   &   & S_{10} \otimes I_{H_{5}} &   &   & ... &   &   &   &   & I_{G_{1}} \otimes P_{54} & G_{1} \otimes I_{H_{5}} + I_{G_{1}} \otimes H_{5} & I_{G_{1}} \otimes P_{56} &   & \cdots \\
  &   &   &   &   &   & S_{10} \otimes I_{H_{6}} &   & ... &   &   &   &   &   & I_{G_{1}} \otimes P_{65} & G_{1} \otimes I_{H_{6}} + I_{G_{1}} \otimes H_{6} & I_{G_{1}} \otimes P_{67} & \cdots \\
  &   &   &   &   &   &   & S_{10} \otimes I_{H_{7}} & ... &   &   &   &   &   &   & I_{G_{1}} \otimes P_{76} & G_{1} \otimes I_{H_{7}} + I_{G_{1}} \otimes H_{7} & \cdots \\
 \hline
\end{array}$
}

Again reordering by level number and deleting edges with $|\Delta L| \ge 2$, we have:

\noindent=\resizebox{.95\textwidth}{!}{

$\begin{array}{|c|cc|ccc|cccc|ccccc|ccc|}
\hline
G_{0} \Box H_{0} & I_{G_{0}} \otimes P_{01} & S_{01} \otimes I_{H_{0}} &   &   &   &   &   &   &   &   &   &   &   &   &   &   & \cdots \\
\hline
I_{G_{0}} \otimes P_{10} & G_{0} \Box H_{1} &   & I_{G_{0}} \otimes P_{12} & S_{01} \otimes I_{H_{1}} &   &   &   &   &   &   &   &   &   &   &   &   & \cdots \\
S_{10} \otimes I_{H_{0}} &   & G_{1} \Box H_{0} &   & I_{G_{1}} \otimes P_{01} & S_{12} \otimes I_{H_{0}} &   &   &   &   &   &   &   &   &   &   &   & \cdots \\
\hline
  & I_{G_{0}} \otimes P_{21} &   & G_{0} \Box H_{2} &   &   & I_{G_{0}} \otimes P_{23} & S_{01} \otimes I_{H_{2}} &   &   &   &   &   &   &   &   &   & \cdots \\
  & S_{10} \otimes I_{H_{1}} & I_{G_{1}} \otimes P_{10} &   & G_{1} \Box H_{1} &   &   & I_{G_{1}} \otimes P_{12} & S_{12} \otimes I_{H_{1}} &   &   &   &   &   &   &   &   & \cdots \\
  &   & S_{21} \otimes I_{H_{0}} &   &   & G_{2} \Box H_{0} &   &   & I_{G_{2}} \otimes P_{01} & S_{23} \otimes I_{H_{0}} &   &   &   &   &   &   &   & \cdots \\
  \hline
  &   &   & I_{G_{0}} \otimes P_{32} &   &   & G_{0} \Box H_{3} &   &   &   & I_{G_{0}} \otimes P_{34} & S_{01} \otimes I_{H_{3}} &   &   &   &   &   & \cdots \\
  &   &   & S_{10} \otimes I_{H_{2}} & I_{G_{1}} \otimes P_{21} &   &   & G_{1} \Box H_{2} &   &   &   & I_{G_{1}} \otimes P_{23} & S_{12} \otimes I_{H_{2}} &   &   &   &   & \cdots \\
  &   &   &   & S_{21} \otimes I_{H_{1}} & I_{G_{2}} \otimes P_{10} &   &   & G_{2} \Box H_{1} &   &   &   & I_{G_{2}} \otimes P_{12} & S_{23} \otimes I_{H_{1}} &   &   &   & \cdots \\
  &   &   &   &   & S_{32} \otimes I_{H_{0}} &   &   &   & G_{3} \Box H_{0} &   &   &   & I_{G_{3}} \otimes P_{01} & S_{34} \otimes I_{H_{0}} &   &   & \cdots \\
  \hline
  &   &   &   &   &   & I_{G_{0}} \otimes P_{43} &   &   &   & G_{0} \Box H_{4} &   &   &   &   & I_{G_{0}} \otimes P_{45} & S_{01} \otimes I_{H_{4}} & \cdots \\
  &   &   &   &   &   & S_{10} \otimes I_{H_{3}} & I_{G_{1}} \otimes P_{32} &   &   &   & G_{1} \Box H_{3} &   &   &   &   & I_{G_{1}} \otimes P_{34} & \cdots \\
  &   &   &   &   &   &   & S_{21} \otimes I_{H_{2}} & I_{G_{2}} \otimes P_{21} &   &   &   & G_{2} \Box H_{2} &   &   &   &   & \cdots \\
  &   &   &   &   &   &   &   & S_{32} \otimes I_{H_{1}} & I_{G_{3}} \otimes P_{10} &   &   &   & G_{3} \Box H_{1} &   &   &   & \cdots \\
  &   &   &   &   &   &   &   &   & S_{43} \otimes I_{H_{0}} &   &   &   &   & G_{4} \Box H_{0} &   &   & \cdots \\
  \hline
  &   &   &   &   &   &   &   &   &   & I_{G_{0}} \otimes P_{54} &   &   &   &   & G_{0} \Box H_{5} &   & \cdots \\
  &   &   &   &   &   &   &   &   &   & S_{10} \otimes I_{H_{4}} & I_{G_{1}} \otimes P_{43} &   &   &   &   & G_{1} \Box H_{4} & \cdots \\
 \vdots & \vdots & \vdots & \vdots & \vdots & \vdots & \vdots & \vdots & \vdots & \vdots & \vdots & \vdots & \vdots & \vdots & \vdots & \vdots & \vdots & \ddots \\
 \hline
\end{array}$
}

\subsection{Box-cross product}
The logic for the box-cross or ``strong'' graph product is identical to the prior sections for the individual products. We are left with the matrix: \\
\noindent\resizebox{\textwidth}{!}{

$\begin{array}{|c|cc|ccc|cccc|ccccc|ccc|}
\hline
G_{0} \boxtimes H_{0} & I_{G_{0}} \otimes P_{01} + G_{0} \otimes P_{01} & S_{01} \otimes I_{H_{0}} + S_{01} \otimes H_{0} &   &   &   &   &   &   &   &   &   &   &   &   &   &   & \cdots \\
\hline
I_{G_{0}} \otimes P_{10} + G_{0} \otimes P_{10} & G_{0} \boxtimes H_{1} & S_{01} \otimes P_{10} & I_{G_{0}} \otimes P_{12} + G_{0} \otimes P_{12} & S_{01} \otimes I_{H_{1}} + S_{01} \otimes H_{1} &   &   &   &   &   &   &   &   &   &   &   &   & \cdots \\
S_{10} \otimes I_{H_{0}} + S_{10} \otimes H_{0} & S_{10} \otimes P_{01} & G_{1} \boxtimes H_{0} &   & I_{G_{1}} \otimes P_{01} + G_{1} \otimes P_{01} & S_{12} \otimes I_{H_{0}} + S_{12} \otimes H_{0} &   &   &   &   &   &   &   &   &   &   &   & \cdots \\
\hline
  & I_{G_{0}} \otimes P_{21} + G_{0} \otimes P_{21} &   & G_{0} \boxtimes H_{2} & S_{01} \otimes P_{21} &   & I_{G_{0}} \otimes P_{23} + G_{0} \otimes P_{23} & S_{01} \otimes I_{H_{2}} + S_{01} \otimes H_{2} &   &   &   &   &   &   &   &   &   & \cdots \\
  & S_{10} \otimes I_{H_{1}} + S_{10} \otimes H_{1} & I_{G_{1}} \otimes P_{10} + G_{1} \otimes P_{10} & S_{10} \otimes P_{12} & G_{1} \boxtimes H_{1} & S_{12} \otimes P_{10} &   & I_{G_{1}} \otimes P_{12} + G_{1} \otimes P_{12} & S_{12} \otimes I_{H_{1}} + S_{12} \otimes H_{1} &   &   &   &   &   &   &   &   & \cdots \\
  &   & S_{21} \otimes I_{H_{0}} + S_{21} \otimes H_{0} &   & S_{21} \otimes P_{01} & G_{2} \boxtimes H_{0} &   &   & I_{G_{2}} \otimes P_{01} + G_{2} \otimes P_{01} & S_{23} \otimes I_{H_{0}} + S_{23} \otimes H_{0} &   &   &   &   &   &   &   & \cdots \\
  \hline
  &   &   & I_{G_{0}} \otimes P_{32} + G_{0} \otimes P_{32} &   &   & G_{0} \boxtimes H_{3} & S_{01} \otimes P_{32} &   &   & I_{G_{0}} \otimes P_{34} + G_{0} \otimes P_{34} & S_{01} \otimes I_{H_{3}} + S_{01} \otimes H_{3} &   &   &   &   &   & \cdots \\
  &   &   & S_{10} \otimes I_{H_{2}} + S_{10} \otimes H_{2} & I_{G_{1}} \otimes P_{21} + G_{1} \otimes P_{21} &   & S_{10} \otimes P_{23} & G_{1} \boxtimes H_{2} & S_{12} \otimes P_{21} &   &   & I_{G_{1}} \otimes P_{23} + G_{1} \otimes P_{23} & S_{12} \otimes I_{H_{2}} + S_{12} \otimes H_{2} &   &   &   &   & \cdots \\
  &   &   &   & S_{21} \otimes I_{H_{1}} + S_{21} \otimes H_{1} & I_{G_{2}} \otimes P_{10} + G_{2} \otimes P_{10} &   & S_{21} \otimes P_{12} & G_{2} \boxtimes H_{1} & S_{23} \otimes P_{10} &   &   & I_{G_{2}} \otimes P_{12} + G_{2} \otimes P_{12} & S_{23} \otimes I_{H_{1}} + S_{23} \otimes H_{1} &   &   &   & \cdots \\
  &   &   &   &   & S_{32} \otimes I_{H_{0}} + S_{32} \otimes H_{0} &   &   & S_{32} \otimes P_{01} & G_{3} \boxtimes H_{0} &   &   &   & I_{G_{3}} \otimes P_{01} + G_{3} \otimes P_{01} & S_{34} \otimes I_{H_{0}} + S_{34} \otimes H_{0} &   &   & \cdots \\
  \hline
  &   &   &   &   &   & I_{G_{0}} \otimes P_{43} + G_{0} \otimes P_{43} &   &   &   & G_{0} \boxtimes H_{4} & S_{01} \otimes P_{43} &   &   &   & I_{G_{0}} \otimes P_{45} + G_{0} \otimes P_{45} & S_{01} \otimes I_{H_{4}} + S_{01} \otimes H_{4} & \cdots \\
  &   &   &   &   &   & S_{10} \otimes I_{H_{3}} + S_{10} \otimes H_{3} & I_{G_{1}} \otimes P_{32} + G_{1} \otimes P_{32} &   &   & S_{10} \otimes P_{34} & G_{1} \boxtimes H_{3} & S_{12} \otimes P_{32} &   &   &   & I_{G_{1}} \otimes P_{34} + G_{1} \otimes P_{34} & \cdots \\
  &   &   &   &   &   &   & S_{21} \otimes I_{H_{2}} + S_{21} \otimes H_{2} & I_{G_{2}} \otimes P_{21} + G_{2} \otimes P_{21} &   &   & S_{21} \otimes P_{23} & G_{2} \boxtimes H_{2} & S_{23} \otimes P_{21} &   &   &   & \cdots \\
  &   &   &   &   &   &   &   & S_{32} \otimes I_{H_{1}} + S_{32} \otimes H_{1} & I_{G_{3}} \otimes P_{10} + G_{3} \otimes P_{10} &   &   & S_{32} \otimes P_{12} & G_{3} \boxtimes H_{1} & S_{34} \otimes P_{10} &   &   & \cdots \\
  &   &   &   &   &   &   &   &   & S_{43} \otimes I_{H_{0}} + S_{43} \otimes H_{0} &   &   &   & S_{43} \otimes P_{01} & G_{4} \boxtimes H_{0} &   &   & \cdots \\
  \hline
  &   &   &   &   &   &   &   &   &   & I_{G_{0}} \otimes P_{54} + G_{0} \otimes P_{54} &   &   &   &   & G_{0} \boxtimes H_{5} & S_{01} \otimes P_{54} & \cdots \\
  &   &   &   &   &   &   &   &   &   & S_{10} \otimes I_{H_{4}} + S_{10} \otimes H_{4} & I_{G_{1}} \otimes P_{43} + G_{1} \otimes P_{43} &   &   &   & S_{10} \otimes P_{45} & G_{1} \boxtimes H_{4} & \cdots \\
 \vdots & \vdots & \vdots & \vdots & \vdots & \vdots & \vdots & \vdots & \vdots & \vdots & \vdots & \vdots & \vdots & \vdots & \vdots & \vdots & \vdots & \ddots \\
 \hline
\end{array}$
}

\renewcommand{\section}{\oldsection}
\newpage
\bibliographystyle{etna}
\bibliography{references}

\begin{thebibliography}{100}

\bibitem{adelson1984pyramid}
{\sc E.~H. Adelson, C.~H. Anderson, J.~R. Bergen, P.~J. Burt, and J.~M. Ogden}, {\em Pyramid methods in image processing}, RCA engineer, 29 (1984), pp.~33--41.

\bibitem{alvarezpicallo2023functorial}
{\sc M.~Alvarez-Picallo, D.~Ghica, D.~Sprunger, and F.~Zanasi}, {\em {Functorial String Diagrams for Reverse-Mode Automatic Differentiation}}, in 31st EACSL Annual Conference on Computer Science Logic (CSL 2023), B.~Klin and E.~Pimentel, eds., Leibniz International Proceedings in Informatics (LIPIcs), 252, Schloss Dagstuhl -- Leibniz-Zentrum f{\"u}r Informatik, Dagstuhl, Germany, 2023, pp.~6:1--6:20.

\bibitem{araujo1997novel}
{\sc C.~P.~S. Araujo}, {\em Novel neural network models for computing homothetic invariances: An image algebra notation}, Journal of Mathematical Imaging and Vision, 7 (1997), pp.~69--83.

\bibitem{awodey2010category}
{\sc S.~Awodey}, {\em Category theory}, OUP Oxford, 2010.

\bibitem{babaud1986uniqueness}
{\sc J.~Babaud, A.~P. Witkin, M.~Baudin, and R.~O. Duda}, {\em Uniqueness of the gaussian kernel for scale-space filtering}, IEEE transactions on pattern analysis and machine intelligence,  (1986), pp.~26--33.

\bibitem{badias2024neural}
{\sc A.~Badías and A.~G. Banerjee}, {\em Neural network layer algebra: A framework to measure capacity and compression in deep learning}, IEEE Transactions on Neural Networks and Learning Systems, 35 (2024), pp.~10380--10393.

\bibitem{ballard1981generalizing}
{\sc D.~H. Ballard}, {\em Generalizing the {H}ough transform to detect arbitrary shapes}, Pattern recognition, 13 (1981), pp.~111--122.

\bibitem{bank1988hierarchical}
{\sc R.~E. Bank, T.~F. Dupont, and H.~Yserentant}, {\em The hierarchical basis multigrid method}, Numerische Mathematik, 52 (1988), pp.~427--458.

\bibitem{baydin2018automated}
{\sc A.~G. Baydin, B.~A. Pearlmutter, A.~A. Radul, and J.~M. Siskind}, {\em Automatic differentiation in machine learning: a survey}, Journal of Machine Learning Research, 18 (2018), pp.~1--43.
\newblock \\ \url{http://jmlr.org/papers/v18/17-468.html}.

\bibitem{bekkersb}
{\sc E.~J. Bekkers}, {\em B-spline cnns on lie groups}, in International Conference on Learning Representations, 2020.

\bibitem{bradley2023structure}
{\sc T.-D. Bradley, J.~L. Gastaldi, and J.~Terilla}, {\em The structure of meaning in language: parallel narratives in linear algebra and category theory}, Notices of the American Mathematical Society, 71 (2023).

\bibitem{brandt2006guide}
{\sc A.~Brandt}, {\em Guide to multigrid development}, in Multigrid Methods: Proceedings of the Conference Held at K{\"o}ln-Porz, November 23--27, 1981, Springer, 2006, pp.~220--312.

\bibitem{carlsson2020topological}
{\sc G.~Carlsson and R.~B. Gabrielsson}, {\em Topological approaches to deep learning}, in Topological Data Analysis: The Abel Symposium 2018, Springer, 2020, pp.~119--146.

\bibitem{chen2021transfer}
{\sc X.~Chen, C.~Gong, Q.~Wan, L.~Deng, Y.~Wan, Y.~Liu, B.~Chen, and J.~Liu}, {\em Transfer learning for deep neural network-based partial differential equations solving}, Advances in Aerodynamics, 3 (2021), p.~36.

\bibitem{chen2022meta}
{\sc Y.~Chen, B.~Dong, and J.~Xu}, {\em Meta-mgnet: Meta multigrid networks for solving parameterized partial differential equations}, Journal of computational physics, 455 (2022), p.~110996.

\bibitem{cruttwell2022categorical}
{\sc G.~S. Cruttwell, B.~Gavranovi{\'c}, N.~Ghani, P.~Wilson, and F.~Zanasi}, {\em Categorical foundations of gradient-based learning}, in European Symposium on Programming, Springer International Publishing Cham, 2022, pp.~1--28.

\bibitem{d2022underspecification}
{\sc A.~D'Amour, K.~Heller, D.~Moldovan, B.~Adlam, B.~Alipanahi, A.~Beutel, C.~Chen, J.~Deaton, J.~Eisenstein, M.~D. Hoffman, et~al.}, {\em Underspecification presents challenges for credibility in modern machine learning}, Journal of Machine Learning Research, 23 (2022), pp.~1--61.

\bibitem{daniely2016deeper}
{\sc A.~Daniely, R.~Frostig, and Y.~Singer}, {\em Toward deeper understanding of neural networks: The power of initialization and a dual view on expressivity}, in Advances in Neural Information Processing Systems, D.~Lee, M.~Sugiyama, U.~Luxburg, I.~Guyon, and R.~Garnett, eds., Curran Associates, Inc., 2016.

\bibitem{deng2017peephole}
{\sc B.~Deng, J.~Yan, and D.~Lin}, {\em Peephole: Predicting network performance before training}, arXiv preprint arXiv:1712.03351,  (2017).

\bibitem{deng2012mnist}
{\sc L.~Deng}, {\em The mnist database of handwritten digit images for machine learning research [best of the web]}, IEEE signal processing magazine, 29 (2012), pp.~141--142.

\bibitem{duda1972use}
{\sc R.~O. Duda and P.~E. Hart}, {\em Use of the {H}ough transformation to detect lines and curves in pictures}, Communications of the ACM, 15 (1972), pp.~11--15.

\bibitem{elsken2019neural}
{\sc T.~Elsken, J.~H. Metzen, and F.~Hutter}, {\em Neural architecture search: A survey}, Journal of Machine Learning Research, 20 (2019), pp.~1--21.

\bibitem{fabregat2023exploring}
{\sc A.~Fabregat-Hern{\'a}ndez, J.~Palanca, and V.~Botti}, {\em Exploring explainable ai: category theory insights into machine learning algorithms}, Machine Learning: Science and Technology, 4 (2023), p.~045061.

\bibitem{felzenszwalb2007hierarchical}
{\sc P.~F. Felzenszwalb and J.~D. Schwartz}, {\em Hierarchical matching of deformable shapes}, in 2007 IEEE conference on computer vision and pattern recognition, IEEE, 2007, pp.~1--8.

\bibitem{feurer2015automated}
{\sc M.~Feurer, A.~Klein, K.~Eggensperger, J.~Springenberg, M.~Blum, and F.~Hutter}, {\em Efficient and robust automated machine learning}, in Advances in Neural Information Processing Systems, C.~Cortes, N.~Lawrence, D.~Lee, M.~Sugiyama, and R.~Garnett, eds., Curran Associates, Inc., 2015.

\bibitem{fiedler1973algebraic}
{\sc M.~Fiedler}, {\em Algebraic connectivity of graphs}, Czechoslovak mathematical journal, 23 (1973), pp.~298--305.

\bibitem{flinkow2024towards}
{\sc T.~Flinkow, B.~A. Pearlmutter, and R.~Monahan}, {\em Towards correct-by-construction machine-learnt models}, in 19th International Conference on Integrated Formal Methods (iFM), 2024.

\bibitem{foltz1980algebraic}
{\sc F.~Foltz, C.~Lair, and G.~M. Kelly}, {\em Algebraic categories with few monoidal biclosed structures or none}, Journal of Pure and Applied Algebra, 17 (1980), pp.~171--177.

\bibitem{fukushima1988neocognitron}
{\sc K.~Fukushima}, {\em Neocognitron: A hierarchical neural network capable of visual pattern recognition}, Neural networks, 1 (1988), pp.~119--130.

\bibitem{gavranovic2024position}
{\sc B.~Gavranovi{\'c}, P.~Lessard, A.~J. Dudzik, T.~von Glehn, J.~G.~M. Ara{\'u}jo, and P.~Veli{\v{c}}kovi{\'c}}, {\em Position: Categorical deep learning is an algebraic theory of all architectures}, in Forty-first International Conference on Machine Learning, 2024.

\bibitem{gong2020geometrically}
{\sc S.~Gong, M.~Bahri, M.~M. Bronstein, and S.~Zafeiriou}, {\em Geometrically principled connections in graph neural networks}, in Proceedings of the IEEE/CVF Conference on Computer Vision and Pattern Recognition (CVPR), June 2020.

\bibitem{grundmann2010efficient}
{\sc M.~Grundmann, V.~Kwatra, M.~Han, and I.~Essa}, {\em Efficient hierarchical graph-based video segmentation}, in 2010 ieee computer society conference on computer vision and pattern recognition, IEEE, 2010, pp.~2141--2148.

\bibitem{guo2021hierarchical}
{\sc K.~Guo, Y.~Hu, Y.~Sun, S.~Qian, J.~Gao, and B.~Yin}, {\em Hierarchical graph convolution network for traffic forecasting}, in Proceedings of the AAAI conference on artificial intelligence, 2021, pp.~151--159.

\bibitem{hartmann2008adaptive}
{\sc D.~Hartmann, M.~Meinke, and W.~Schr{\"o}der}, {\em An adaptive multilevel multigrid formulation for cartesian hierarchical grid methods}, Computers \& Fluids, 37 (2008), pp.~1103--1125.

\bibitem{he2016deep}
{\sc K.~He, X.~Zhang, S.~Ren, and J.~Sun}, {\em Deep residual learning for image recognition}, in Proceedings of the IEEE conference on computer vision and pattern recognition, 2016, pp.~770--778.

\bibitem{huang2022learning}
{\sc R.~Huang, R.~Li, and Y.~Xi}, {\em Learning optimal multigrid smoothers via neural networks}, SIAM Journal on Scientific Computing, 45 (2022), pp.~S199--S225.

\bibitem{illingworth1987adaptive}
{\sc J.~Illingworth and J.~Kittler}, {\em The adaptive {H}ough transform}, IEEE Transactions on Pattern Analysis and Machine Intelligence,  (1987), pp.~690--698.

\bibitem{illingworth1988survey}
\leavevmode\vrule height 2pt depth -1.6pt width 23pt, {\em A survey of the {H}ough transform}, Computer vision, graphics, and image processing, 44 (1988), pp.~87--116.

\bibitem{ImrichKlavzar}
{\sc W.~Imrich and S.~Klavzar}, {\em Product graphs, structure and recognition}, John Wiley \& Sons, 2000.

\bibitem{istrate2019tapas}
{\sc R.~Istrate, F.~Scheidegger, G.~Mariani, D.~Nikolopoulos, C.~Bekas, and A.~C.~I. Malossi}, {\em Tapas: Train-less accuracy predictor for architecture search}, in Proceedings of the AAAI conference on artificial intelligence, 2019, pp.~3927--3934.

\bibitem{jackson2017algebraic}
{\sc E.~C. Jackson, J.~A. Hughes, M.~Daley, and M.~Winter}, {\em An algebraic generalization for graph and tensor-based neural networks}, in 2017 IEEE Conference on Computational Intelligence in Bioinformatics and Computational Biology (CIBCB), IEEE, 2017, pp.~1--8.

\bibitem{jansson2022scale}
{\sc Y.~Jansson and T.~Lindeberg}, {\em Scale-invariant scale-channel networks: Deep networks that generalise to previously unseen scales}, Journal of Mathematical Imaging and Vision, 64 (2022), pp.~506--536.

\bibitem{jin2006context}
{\sc Y.~Jin and S.~Geman}, {\em Context and hierarchy in a probabilistic image model}, in 2006 IEEE computer society conference on computer vision and pattern recognition (CVPR'06), IEEE, 2006, pp.~2145--2152.

\bibitem{joyce2024algebraic}
{\sc J.~Joyce and J.~Verschelde}, {\em Algebraic representations for faster predictions in convolutional neural networks}, in International Workshop on Computer Algebra in Scientific Computing, Springer, 2024, pp.~161--177.

\bibitem{kalogeropoulos2024scale}
{\sc I.~Kalogeropoulos, G.~Bouritsas, and Y.~Panagakis}, {\em Scale equivariant graph metanetworks}, in The Thirty-eighth Annual Conference on Neural Information Processing Systems, 2024.

\bibitem{kang2020model}
{\sc D.~Kang, D.~Raghavan, P.~Bailis, and M.~Zaharia}, {\em Model assertions for monitoring and improving ml models}, Proceedings of Machine Learning and Systems, 2 (2020), pp.~481--496.

\bibitem{ke2017multigrid}
{\sc T.-W. Ke, M.~Maire, and S.~X. Yu}, {\em Multigrid neural architectures}, in Proceedings of the IEEE Conference on Computer Vision and Pattern Recognition, 2017, pp.~6665--6673.

\bibitem{kingma2014adam}
{\sc D.~P. Kingma and J.~Ba}, {\em Adam: A method for stochastic optimization}, arXiv preprint arXiv:1412.6980,  (2014).

\bibitem{kirsch2018modular}
{\sc L.~Kirsch, J.~Kunze, and D.~Barber}, {\em Modular networks: Learning to decompose neural computation}, in Advances in Neural Information Processing Systems, S.~Bengio, H.~Wallach, H.~Larochelle, K.~Grauman, N.~Cesa-Bianchi, and R.~Garnett, eds., Curran Associates, Inc., 2018.

\bibitem{KnauerAlgGraphs}
{\sc U.~Knauer and K.~Knauer}, {\em Algebraic Graph Theory}, Walter de Gruyter GmbH, 2019.

\bibitem{krizhevsky2012imagenet}
{\sc A.~Krizhevsky, I.~Sutskever, and G.~E. Hinton}, {\em Imagenet classification with deep convolutional neural networks}, in Advances in Neural Information Processing Systems, F.~Pereira, C.~Burges, L.~Bottou, and K.~Weinberger, eds., Curran Associates, Inc., 2012.

\bibitem{lecun2015deep}
{\sc Y.~LeCun, Y.~Bengio, and G.~Hinton}, {\em Deep learning}, nature, 521 (2015), pp.~436--444.

\bibitem{lehner2023gauge}
{\sc C.~Lehner and T.~Wettig}, {\em Gauge-equivariant neural networks as preconditioners in lattice qcd}, Physical Review D, 108 (2023), p.~034503.

\bibitem{lei2017deriving}
{\sc T.~Lei, W.~Jin, R.~Barzilay, and T.~Jaakkola}, {\em Deriving neural architectures from sequence and graph kernels}, in Proceedings of the 34th International Conference on Machine Learning, D.~Precup and Y.~W. Teh, eds., Proceedings of Machine Learning Research, 70, PMLR, 06--11 Aug 2017, pp.~2024--2033.

\bibitem{li2019semi}
{\sc J.~Li, Y.~Rong, H.~Cheng, H.~Meng, W.~Huang, and J.~Huang}, {\em Semi-supervised graph classification: A hierarchical graph perspective}, in The World Wide Web Conference, 2019, pp.~972--982.

\bibitem{liang2024solving}
{\sc S.~Liang, S.~W. Jiang, J.~Harlim, and H.~Yang}, {\em Solving pdes on unknown manifolds with machine learning}, Applied and Computational Harmonic Analysis, 71 (2024), p.~101652.

\bibitem{lim2023graph}
{\sc D.~Lim, H.~Maron, M.~T. Law, J.~Lorraine, and J.~Lucas}, {\em Graph metanetworks for processing diverse neural architectures}, in The Twelfth International Conference on Learning Representations, 2023.

\bibitem{lindeberg2013scale}
{\sc T.~Lindeberg}, {\em Scale-space theory in computer vision}, Springer Science \& Business Media, 2013.

\bibitem{liu2018progressive}
{\sc C.~Liu, B.~Zoph, M.~Neumann, J.~Shlens, W.~Hua, L.-J. Li, L.~Fei-Fei, A.~Yuille, J.~Huang, and K.~Murphy}, {\em Progressive neural architecture search}, in Proceedings of the European conference on computer vision (ECCV), 2018, pp.~19--34.

\bibitem{liu2018learning}
{\sc S.~Liu, L.~Giles, and A.~Ororbia}, {\em Learning a hierarchical latent-variable model of 3d shapes}, in 2018 international conference on 3D vision (3DV), IEEE, 2018, pp.~542--551.

\bibitem{liu2023multiresolution}
{\sc Y.~Liu, C.~Ponce, S.~L. Brunton, and J.~N. Kutz}, {\em Multiresolution convolutional autoencoders}, Journal of Computational Physics, 474 (2023), p.~111801.

\bibitem{lovasz2012large}
{\sc L.~Lov{\'a}sz}, {\em Large networks and graph limits, volume 60 of american mathematical society colloquium publications}, American Mathematical Society, Providence, RI, 22 (2012).

\bibitem{luz2020learning}
{\sc I.~Luz, M.~Galun, H.~Maron, R.~Basri, and I.~Yavneh}, {\em Learning algebraic multigrid using graph neural networks}, in International Conference on Machine Learning, PMLR, 2020, pp.~6489--6499.

\bibitem{madan2022and}
{\sc S.~Madan, T.~Henry, J.~Dozier, H.~Ho, N.~Bhandari, T.~Sasaki, F.~Durand, H.~Pfister, and X.~Boix}, {\em When and how convolutional neural networks generalize to out-of-distribution category--viewpoint combinations}, Nature Machine Intelligence, 4 (2022), pp.~146--153.

\bibitem{Margenstern}
{\sc M.~Margenstern}, {\em An application of grossone to the study of a family of tilings of the hyperbolic plane}, Applied Mathematics and Computation, 218 (2012), pp.~8005--8018.
\newblock https://www.sciencedirect.com/science/article/pii/S0096300311005698, \\ \url{https://doi.org/10.1016/j.amc.2011.04.014}.

\bibitem{maroninvariant}
{\sc H.~Maron, H.~Ben-Hamu, N.~Shamir, and Y.~Lipman}, {\em Invariant and equivariant graph networks}, in International Conference on Learning Representations, 2019.

\bibitem{mcgreivy2024weak}
{\sc N.~McGreivy and A.~Hakim}, {\em Weak baselines and reporting biases lead to overoptimism in machine learning for fluid-related partial differential equations}, Nature machine intelligence, 6 (2024), pp.~1256--1269.

\bibitem{mellor2021neural}
{\sc J.~Mellor, J.~Turner, A.~Storkey, and E.~J. Crowley}, {\em Neural architecture search without training}, in International conference on machine learning, PMLR, 2021, pp.~7588--7598.

\bibitem{memin1998multigrid}
{\sc E.~Memin and P.~Perez}, {\em A multigrid approach for hierarchical motion estimation}, in Sixth International Conference on Computer Vision (IEEE Cat. No. 98CH36271), IEEE, 1998, pp.~933--938.

\bibitem{mjolsness1988neural}
{\sc E.~Mjolsness, G.~Gindi, and P.~Anandan}, {\em Neural networks for model matching and perceptual organization}, Advances in Neural Information Processing Systems, 1 (1988).

\bibitem{Mjolsness_thesis}
{\sc E.~D. Mjolsness}, {\em Neural Networks, Pattern Recognition, and Fingerprint Hallucination}, Ph.D.~thesis, California Institute of Technology, 1986.

\bibitem{NLab_FTV}
{\sc {nLab authors}}, {\em funny tensor product}.
\newblock \url{https://ncatlab.org/nlab/show/funny+tensor+product}, May 2025.
\newblock \href{https://ncatlab.org/nlab/revision/funny+tensor+product/12}{Revision 12}.

\bibitem{olah2017research}
{\sc C.~Olah and S.~Carter}, {\em Research debt}, Distill, 2 (2017), p.~e5.

\bibitem{oosterlee1995robustness}
{\sc C.~W. Oosterlee and P.~Wesseling}, {\em On the robustness of a multiple semi-coarsened grid method}, Zeitschrift Fur Angewandte Mathematik Und Mechanik, 75 (1995), pp.~251--251.

\bibitem{parekh2000constructive}
{\sc R.~Parekh, J.~Yang, and V.~Honavar}, {\em Constructive neural-network learning algorithms for pattern classification}, IEEE Transactions on Neural Networks, 11 (2000), pp.~436--451.

\bibitem{paszke2019pytorch}
{\sc A.~Paszke, S.~Gross, F.~Massa, A.~Lerer, J.~Bradbury, G.~Chanan, T.~Killeen, Z.~Lin, N.~Gimelshein, L.~Antiga, et~al.}, {\em Pytorch: An imperative style, high-performance deep learning library}, in Advances in Neural Information Processing Systems 32, 2019, pp.~8024--8035.

\bibitem{perera2024multiscale}
{\sc R.~Perera and V.~Agrawal}, {\em Multiscale graph neural networks with adaptive mesh refinement for accelerating mesh-based simulations}, Computer Methods in Applied Mechanics and Engineering, 429 (2024), p.~117152.

\bibitem{perret2019higra}
{\sc B.~Perret, G.~Chierchia, J.~Cousty, S.~J.~F. Guimaraes, Y.~Kenmochi, and L.~Najman}, {\em Higra: Hierarchical graph analysis}, SoftwareX, 10 (2019), p.~100335.

\bibitem{pham2018efficient}
{\sc H.~Pham, M.~Guan, B.~Zoph, Q.~Le, and J.~Dean}, {\em Efficient neural architecture search via parameters sharing}, in International conference on machine learning, PMLR, 2018, pp.~4095--4104.

\bibitem{pineau2021improving}
{\sc J.~Pineau, P.~Vincent-Lamarre, K.~Sinha, V.~Larivi{\`e}re, A.~Beygelzimer, F.~d'Alch{\'e} Buc, E.~Fox, and H.~Larochelle}, {\em Improving reproducibility in machine learning research (a report from the neurips 2019 reproducibility program)}, Journal of machine learning research, 22 (2021), pp.~1--20.

\bibitem{raff2019step}
{\sc E.~Raff}, {\em A step toward quantifying independently reproducible machine learning research}, Advances in Neural Information Processing Systems, 32 (2019).

\bibitem{reed2017parallel}
{\sc S.~Reed, A.~Oord, N.~Kalchbrenner, S.~G. Colmenarejo, Z.~Wang, Y.~Chen, D.~Belov, and N.~Freitas}, {\em Parallel multiscale autoregressive density estimation}, in International conference on machine learning, PMLR, 2017, pp.~2912--2921.

\bibitem{ritter2017deriving}
{\sc S.~Ritter, D.~G.~T. Barrett, A.~Santoro, and M.~M. Botvinick}, {\em Cognitive psychology for deep neural networks: A shape bias case study}, in Proceedings of the 34th International Conference on Machine Learning, D.~Precup and Y.~W. Teh, eds., Proceedings of Machine Learning Research, 70, PMLR, 06--11 Aug 2017, pp.~2940--2949.

\bibitem{ronneberger2015u}
{\sc O.~Ronneberger, P.~Fischer, and T.~Brox}, {\em U-net: Convolutional networks for biomedical image segmentation}, in International Conference on Medical image computing and computer-assisted intervention, Springer, 2015, pp.~234--241.

\bibitem{salehin2024automl}
{\sc I.~Salehin, M.~S. Islam, P.~Saha, S.~Noman, A.~Tuni, M.~M. Hasan, and M.~A. Baten}, {\em Automl: A systematic review on automated machine learning with neural architecture search}, Journal of Information and Intelligence, 2 (2024), pp.~52--81.

\bibitem{scott2020graph}
{\sc C.~B. Scott and E.~Mjolsness}, {\em Graph prolongation convolutional networks: explicitly multiscale machine learning on graphs with applications to modeling of cytoskeleton}, Machine Learning: Science and Technology, 2 (2020), p.~015009.

\bibitem{scott2021graph}
{\sc C.~B. Scott and E.~{M}jolsness}, {\em {G}raph {D}iffusion {D}istance: {P}roperties and {E}fficient {C}omputation}, {P}{L}o{S} {ONE}, 16 (2021), p.~e0249624.

\bibitem{sporring2013gaussian}
{\sc J.~Sporring, M.~Nielsen, L.~Florack, and P.~Johansen}, {\em Gaussian scale-space theory}, Springer Science \& Business Media, 2013.

\bibitem{srivastava2020empirical}
{\sc M.~Srivastava, B.~Nushi, E.~Kamar, S.~Shah, and E.~Horvitz}, {\em An empirical analysis of backward compatibility in machine learning systems}, in Proceedings of the 26th ACM SIGKDD International Conference on Knowledge Discovery \& Data Mining, 2020, pp.~3272--3280.

\bibitem{stuben2001review}
{\sc K.~St{\"u}ben}, {\em A review of algebraic multigrid}, Numerical Analysis: Historical Developments in the 20th Century,  (2001), pp.~331--359.

\bibitem{tian2024visual}
{\sc K.~Tian, Y.~Jiang, Z.~Yuan, B.~Peng, and L.~Wang}, {\em Visual autoregressive modeling: Scalable image generation via next-scale prediction}, Advances in neural information processing systems, 37 (2024), pp.~84839--84865.

\bibitem{trottenberg2001multigrid}
{\sc U.~Trottenberg, C.~W. Oosterlee, and A.~Schuller}, {\em Multigrid methods}, Academic press, 2001.

\bibitem{voulodimos2018deep}
{\sc A.~Voulodimos, N.~Doulamis, A.~Doulamis, and E.~Protopapadakis}, {\em Deep learning for computer vision: A brief review}, Computational intelligence and neuroscience, 2018 (2018), p.~7068349.

\bibitem{watanabe2001algebraic}
{\sc S.~Watanabe}, {\em Algebraic analysis for nonidentifiable learning machines}, Neural Computation, 13 (2001), pp.~899--933.

\bibitem{weber2013free}
{\sc M.~Weber}, {\em Free products of higher operad algebras}, Theory and applications of categories, 28 (2013), pp.~24--65.

\bibitem{wesseling1995introduction}
{\sc P.~Wesseling}, {\em Introduction to multigrid methods}, {T}ech. {R}eport, Institute for Computer Applications in Science and Engineering, 1995.

\bibitem{worrall2019deep}
{\sc D.~Worrall and M.~Welling}, {\em Deep scale-spaces: Equivariance over scale}, Advances in Neural Information Processing Systems, 32 (2019).

\bibitem{xiao2017fashion}
{\sc H.~Xiao, K.~Rasul, and R.~Vollgraf}, {\em Fashion-mnist: a novel image dataset for benchmarking machine learning algorithms}, arXiv preprint arXiv:1708.07747,  (2017).

\bibitem{xu2022neural}
{\sc T.~Xu and Y.~Maruyama}, {\em Neural string diagrams: a universal modelling language for categorical deep learning}, in Artificial General Intelligence: 14th International Conference, AGI 2021, Palo Alto, CA, USA, October 15--18, 2021, Proceedings 14, Springer, 2022, pp.~306--315.

\bibitem{yang2023reinforcement}
{\sc J.~Yang, T.~Dzanic, B.~Petersen, J.~Kudo, K.~Mittal, V.~Tomov, J.-S. Camier, T.~Zhao, H.~Zha, T.~Kolev, et~al.}, {\em Reinforcement learning for adaptive mesh refinement}, in International conference on artificial intelligence and statistics, PMLR, 2023, pp.~5997--6014.

\bibitem{yang2022amgnet}
{\sc Z.~Yang, Y.~Dong, X.~Deng, and L.~Zhang}, {\em Amgnet: multi-scale graph neural networks for flow field prediction}, Connection Science, 34 (2022), pp.~2500--2519.

\bibitem{yehudai2021local}
{\sc G.~Yehudai, E.~Fetaya, E.~Meirom, G.~Chechik, and H.~Maron}, {\em From local structures to size generalization in graph neural networks}, in International Conference on Machine Learning, PMLR, 2021, pp.~11975--11986.

\bibitem{ying2018hierarchical}
{\sc Z.~Ying, J.~You, C.~Morris, X.~Ren, W.~Hamilton, and J.~Leskovec}, {\em Hierarchical graph representation learning with differentiable pooling}, Advances in neural information processing systems, 31 (2018).

\bibitem{you2020structure}
{\sc J.~You, J.~Leskovec, K.~He, and S.~Xie}, {\em Graph structure of neural networks}, in Proceedings of the 37th International Conference on Machine Learning, H.~D. III and A.~Singh, eds., Proceedings of Machine Learning Research, 119, PMLR, 13--18 Jul 2020, pp.~10881--10891.

\bibitem{zhang2024blending}
{\sc E.~Zhang, A.~Kahana, A.~Kopani{\v{c}}{\'a}kov{\'a}, E.~Turkel, R.~Ranade, J.~Pathak, and G.~E. Karniadakis}, {\em Blending neural operators and relaxation methods in pde numerical solvers}, Nature Machine Intelligence, 6 (2024), pp.~1303--1313.

\bibitem{zhang2022hierarchical}
{\sc Z.~Zhang, Q.~Liu, Q.~Hu, and C.-K. Lee}, {\em Hierarchical graph transformer with adaptive node sampling}, Advances in Neural Information Processing Systems, 35 (2022), pp.~21171--21183.

\bibitem{zhu2022scaling}
{\sc W.~Zhu, Q.~Qiu, R.~Calderbank, G.~Sapiro, and X.~Cheng}, {\em Scaling-translation-equivariant networks with decomposed convolutional filters}, Journal of Machine Learning Research, 23 (2022), pp.~1--45.
\newblock \\ \url{http://jmlr.org/papers/v23/20-099.html}.

\bibitem{zoph2016neural}
{\sc B.~Zoph and Q.~Le}, {\em Neural architecture search with reinforcement learning}, in International Conference on Learning Representations, 2016.

\bibitem{zweig2021functional}
{\sc A.~Zweig and J.~Bruna}, {\em A functional perspective on learning symmetric functions with neural networks}, in International Conference on Machine Learning, PMLR, 2021, pp.~13023--13032.

\end{thebibliography}

\end{document}